\documentclass[authoryear,final,11pt,times]{elsarticle}

\usepackage{amssymb}
\usepackage{lineno}
\usepackage{multirow}
\usepackage{longtable}
\usepackage{hhline}
\usepackage{booktabs} 
\usepackage{array}
\usepackage{booktabs} 
\usepackage{listings}
\usepackage{amsmath}
\usepackage{graphicx}
\usepackage{rotating}
\usepackage{tikz}
\usepackage[colorlinks=true, linkcolor=blue, urlcolor=blue, citecolor=blue]{hyperref}

\usetikzlibrary{positioning, shapes, arrows}
\newtheorem{definition}{Definition}

\usepackage{url}
\usepackage{boxedminipage}
\usepackage{textpos}

\journal{}

\begin{document}

\begin{frontmatter}

\title{What is a Digital Twin Anyway? Deriving the Definition for the Built Environment from over 15,000 Scientific Publications}

\author[doa]{Mahmoud Abdelrahman}
\author[doa,upd]{Edgardo Macatulad}
\author[doa]{Binyu Lei}
\author[sec]{Matias Quintana}
\author[dbe]{Clayton Miller}
\author[doa,dre]{Filip Biljecki\corref{cor1}}

\affiliation[doa]{organization={Department of Architecture, National University of Singapore}, 
            addressline={4 Architecture Drive}, 
            city={Singapore},
            postcode={117566}, 
            country={Singapore}}
\affiliation[upd]{organization={Department of Geodetic Engineering, University of the Philippines}, 
            addressline={Diliman},
            city={Quezon City},
            postcode={1101},
            country={Philippines}}
\affiliation[sec]{organization={Future Cities Lab Global Programme, Singapore-ETH Centre},
            addressline={CREATE campus, \#06-01 CREATE Tower}, 
            city={Singapore},
            postcode={138602}, 
            country={Singapore}}
\affiliation[dbe]{organization={Department of the Built Environment, National University of Singapore}, 
            addressline={4 Architecture Drive}, 
            city={Singapore},
            postcode={117566}, 
            country={Singapore}}
\affiliation[dre]{organization={Department of Real Estate, National University of Singapore}, 
            addressline={15 Kent Ridge Dr}, 
            city={Singapore},
            postcode={119245}, 
            country={Singapore}}
\cortext[cor1]{Corresponding author}

\begin{abstract}
\begin{textblock*}{\textwidth}(0cm,-14.5cm)
\begin{center}
\begin{footnotesize}
\begin{boxedminipage}{1\textwidth}
This is the Accepted Manuscript version of an article published by Elsevier in the journal \emph{Building and Environment} in 2025, which is available at:\\ \url{https://doi.org/10.1016/j.buildenv.2025.112748}\\ Cite as:
Abdelrahman M, Macatulad E, Lei B, Quintana M, Miller C, Biljecki F (2025): What is a Digital Twin anyway? Deriving the definition for the built environment from over 15,000 scientific publications. \textit{Building and Environment}, 274: 112748.
\end{boxedminipage}
\end{footnotesize}
\end{center}
\end{textblock*}

\begin{textblock*}{1.5\textwidth}(-0.8cm,12.3cm)
{\tiny{\copyright{ }2025, Elsevier. Licensed under the Creative Commons Attribution-NonCommercial-NoDerivatives 4.0 International (\url{http://creativecommons.org/licenses/by-nc-nd/4.0/})}}
\end{textblock*}
The concept of Digital Twins (DT) has attracted significant attention across various domains, particularly within the built environment. However, there is a sheer volume of definitions and the terminological consensus remains out of reach. The lack of a universally accepted definition leads to ambiguities in their conceptualization and implementation, and may cause miscommunication for both researchers and practitioners. 

We employed Natural Language Processing (NLP) techniques to systematically extract and analyze definitions of DTs from a corpus of more than 15,000 full-text articles spanning diverse disciplines. The study compares these findings with insights from an expert survey that included 52 experts. The study identifies concurrence on the components that comprise a ``Digital Twin'' from a practical perspective across various domains, contrasting them with those that do not, to identify deviations. We investigate the evolution of digital twin definitions over time and across different scales, including manufacturing, building, and urban/geospatial perspectives. 
We extracted the main components of Digital Twins using Text Frequency Analysis and N-gram analysis. 
Subsequently, we identified components that appeared in the literature and conducted a Chi-square test to assess the significance of each component in different domains. 

Our analysis identified key components of digital twins and revealed significant variations in definitions based on application domains, such as manufacturing, building, and urban contexts. The analysis of DT components reveal two major groups of DT types: High-Performance Real-Time (HPRT) DTs, and Long-Term Decision Support (LTDS) DTs. Contrary to common assumptions, we found that components such as simulation, AI/ML, real-time capabilities, and bi-directional data flow are not yet fully mature in the digital twins of the built environment. We derived two definitions for the Building/Architecture DT and the City/Urban DTs. Both definitions have a must-have components (such as spatial and temporal data updates) and good-to-have components such as prediction, AI, bi-directional data flow, and Real-time data exchange. %

One of the key findings is that the definition of digital twins has not yet reached its equilibrium phase, highlighting the need for ongoing revisions as technologies emerge or existing ones become obsolete. To address this, we introduce a novel, reproducible methodology that enables researchers to refine and adapt the current definitions in response to technological advancements or deprecations.

\end{abstract}

\begin{keyword}
Digital Twin \sep Terminology \sep NLP \sep LLMs \sep Building Digital Twin \sep Urban Digital Twin \sep City Digital Twin 
\end{keyword}

\end{frontmatter}

\section{Introduction}
\label{sec:introduction}

The concept of a Digital Twin (DT) traces its origins to a book titled Mirrored Worlds published three decades ago \citep{gelernter1993mirror}. Various terms have been coined subsequently, e.g.\ Mirrored Space Model (MSM) and Information Mirroring Model (IMM), and later ``Conceptual Ideal for Product Lifecycle Management (PLM)'' by \citet{grieves2005product}, evolving the discourse with defining the main components of a DT.
Since then, the term has evolved to encompass a broad spectrum of concepts and levels of fidelity, ranging from static models to dynamic simulations with real-time data integration and even autonomous capabilities (from a data visualization tool to a fully autonomous decision making system such as autonomous cars).
For example, the term and concept of Digital Twin have been widely adopted in manufacturing industry \citep{leng2021digital,jiang2021industrial}, and transcended it, reaching the built environment and urban/geospatial domains \citep{lehner2020digital,dembski2020urban,juarez2021digital,Sommer2023,bartos2021pipedream,chomiak2023use,white2021digital,2022_3dgeoinfo_dt_use_cases,2023_scs_human_dt,Jeddoub_2023,Metcalfe_2024,Shirowzhan_2020,Yan_2019} including urban logistics and transportation \citep{Belfadel2023,kim2024digital},
urban-social interaction \citep{Wang2023}, participatory human-centric exploration \citep{lei2023humans,lei2024integrating}. conversion of Building Information Modeling (BIM), IoT, and GIS \citep{Shi2023, Xia2022}.
Furthermore, according to more recent literature, the DT scope appears to have broadened to represent the entire lifecycle of a system, not just a single lifecycle phase (e.g., conceptual, design and planning, construction/manufacturing, operation and maintenance, and decomposition) \citep{Boje2023} as well as a wide range of applications \citep{lei2023challenges}. 
At the time of writing this paper (mid-2024), a query for the term \texttt{`Digital Twin'} on ScienceDirect, a major database of peer-reviewed scholarly literature, yields 52,495 results. The majority of these results are dominated by Smart Manufacturing and Industrial Engineering (e.g., Industry 4.0). On a separate query, a total of 11,000 results are found to be related to the Built Environment (BE)\footnote{We obtained this estimation by using the following boolean search: \textit{ ``\texttt{(`Digital twin'  AND (`Building' OR `Built Environment' OR `City' OR `urban' OR `smart building' OR `BIM' OR `architecture')}}"}\footnote{In the context of this paper, we refer to the BE as buildings, urban spaces/places and cities. which constitute different scales within the Architecture, Engineering, and Construction (AEC) domain}. This multitude of literature demonstrates the intense and rapid interest in digital twin concept and technology and its applications in the built environment \citep{OPOKU2021102726,pereira_descriptive_2021,bolton2018gemini,Xia2022}. 
Despite the growing interest in DTs within the built environment, the concept has been reported to remain unclear for many practitioners and researchers \citep{OSADCHA2023106704,chong2022digital}, unlike in some other industries such as Manufacturing \citep{KRITZINGER20181016} and Aerospace \citep{9656111} where it appears to be converging towards a well-established consensus.

The rapid growth of interest in emergent technologies such as DTs can lead to fragmented knowledge about its core functionalities and key elements. This fragmentation often results in ambiguities across industry and research papers concerning its precise definition \citep{KRITZINGER20181016,Bartsch2021,CIMINO2019103130,yang2021developments,SteindlGeneric}, its essential components and minimum functionalities for a system to be considered a DT, and how it is applied specifically within the BE. For example, in discussions surrounding DTs, terms such as `data reference', `software representation', `living model', `virtual replica', and `visual asset management system', `3D city models' and `City Information Models (CIM)' are all used interchangeably. However, not all of them accurately capture the exact essence of a Digital Twin. Amplifying this confusion, researchers have defined DTs from one or more perspectives, often vaguely, leading to a range of interpretations (Table~\ref{table_definitions_review}).
Further, the ambiguity is exacerbated by its common use as a buzzword and the term is also affected by the expedient practice and trend of giving new names to existing concepts, simply to make them seem novel.

\begin{figure}[!h]
    \centering
    \includegraphics[width=\linewidth]{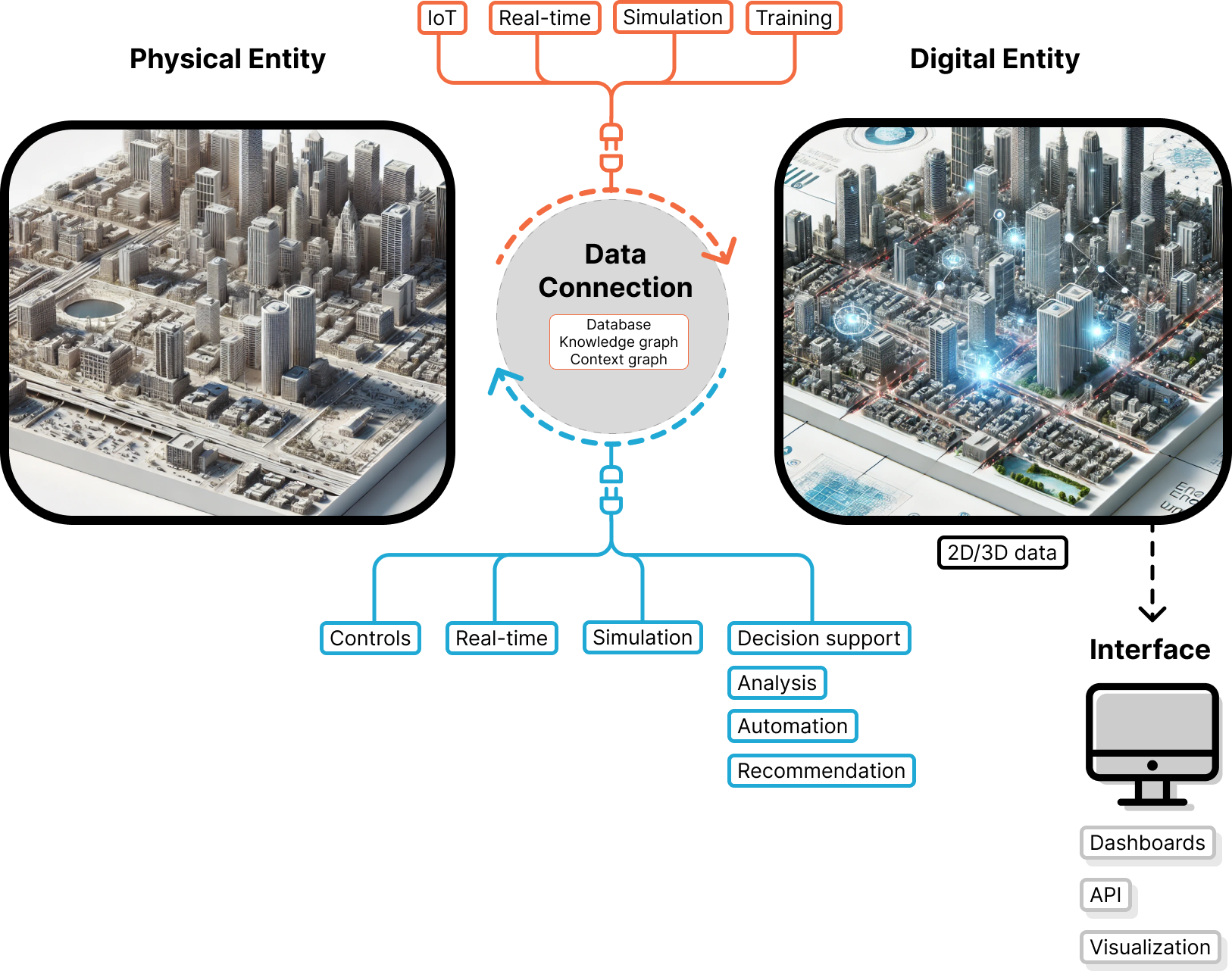}
    \caption{Illustration of commonly discussed Digital Twin components.}
    \label{fig:dt_components_illustration}
\end{figure}

On the other hand, many researchers have defined the lowest common denominator of the Digital Twin concept \citep{JONES202036,grieves2014digital,tao2018digitalindustry,wagg2020key}, which consist of 3 main components: 1) Physical entity, 2) Virtual entity, and 3) Data connection. Others have extended these minimum components to include 4) Real-time capabilities, 5) Real-time update of the digital system with the physical system, and 6) Decision support \citep{Boschert2016,deren2021smart,chaturvedi2016integrating} (Figure \ref{fig:dt_components_illustration}).
While these interpretations may not always provide a comprehensive understanding, and some may even collide, they can capture various aspects of digital twins relevant to the specific study where the definition is employed. 
By consolidating diverse perspectives on digital twins from a large number of sources, our goal is to formulate a data-driven, comprehensive, and standardized definition of digital twins that captures a unified understanding across the built environment, and bring us one step closer to a consensus.

Considering the inconsistent terminology and contrasting implementations, recent studies have introduced the concept of DT maturity stages \citep{10367836,HARAGUCHI2024123409,10.1007/978-3-031-39619-9_43,LIU2024102592}, which is used to assess how advanced a DT is in terms of technology and functionality. These may range from basic and static geometric models to highly integrated systems with advanced features such as real-time capabilities and AI/ML functionalities supporting simulations. 
For example, \cite{10367836} and \citet{evans2019digital} identified 5 maturity stages in City DTs: 1) Reality capture, 2) 2D maps or 3D models, 3) addition of real-time data, 4) two-way integration and interactions, and 5) autonomous operations and maintenance. \cite{HARAGUCHI2024123409} introduced the CITYSTEPS Maturity model which grades them according to eight maturity stages: 1) Preparation and planning stage; 2) 2D stage; 3) 3D static stage; 4) 3D dynamic stage; 5) Dynamically Integrated 3D Stage; 6) Real-time decision-making stage; 7) Autonomous decision-making stage; and 8) Real-time synchronization and autonomous implementation stage. \citet{doi:10.1080/20964471.2022.2160156} conducted a review on CDTs, highlighting their maturity levels and differentiating them from traditional 3D city models~\citep{2022_jag_3d_svi,2023_ijgis_3d_city_index,Pa_en_2024}, while identifying key research gaps, such as the need for advanced real-time data analytics and public participation in CDT development.
These recent developments signal that the community is increasingly acknowledging that DTs may come in different shapes and flavours and considering Digital Twin as a broader umbrella term for a range of concepts and implementations of virtual replicas of the built environment.
Nevertheless, an overarching definition and conceptualization of the term in the built environment remain missing \citep{delgado2021digital, DAVILADELGADO2021101332,JONES202036} despite the substantial efforts invested to address this gap \citep{ellul2022location, stoter2021digital, ellul2024towards, callcut2021digital, 2022_3dgeoinfo_dt_use_cases}.

In this paper, we conduct a large scale systematic review coupled with an expert survey to shed light on the definition of a digital twin in the built environment. A systematic review methodology was used because it helps reduce bias in the selection and analysis of existing research, and it follows initiatives in other fields such as circular economy that have also grappled with similar terminological issues~\citep{Kirchherr_2017,Reitsma_2013}. This review involves analyzing approximately one thousand Digital Twin mentions, interpretations, explanations, and definitions that have appeared in more than 15,000 publications from 2001 to January 2024. By systematically examining such a large sample size, we provided an empirical snapshot of current scholarly understandings of DTs, as well as the evolution of the definition over the years. We construct our approach on the premise that researchers are likely to choose definitions that align with their research focus or objectives, even if some of them do not align with the current predominant definitions. We then analyzed the definitions used in the Built Environment (Building, Urban domains), and we used Manufacturing domain as a benchmark for comparison to extract the DT components of each. The analysis aimed to understand the various perspectives on DTs, develop a definition based on the most common characteristics understood by researchers through practice and application, and ensure this definition was comprehensive enough to encompass a wide range of DT components, ultimately capturing the true essence of what constitutes a Digital Twin or in other words, of what BE researchers have been referring to as Digital Twins. We also consider the maturity levels of DT in different domains, recognizing that not all implementations require the same level of maturity.
Finally, by deriving a definition, in our paper we question some instances in practice that manipulate the definition of DT to align their concepts as DT but significantly depart from the actual concept.

The approach to obtain clarity on terminology from a vast body of knowledge that we introduce in this paper is novel and our development, and a contribution that can be applied also in other domains. The rest of the article is structured as follows. Section~\ref{sc:related} provides a review of the literature on Digital Twins, highlighting the various definitions that have been proposed in the literature. Section~\ref{sec:methodology} outlines the methodology used in this research, including data sources, data processing, and the methodology to derive a new definition of Digital Twin. Section~\ref{sec:results} presents the results of the study, including the analysis of the components and definitions extracted from the literature and the expert survey. Section~\ref{sec:discussion} discusses the findings, implications, and futures of the study, and Section~\ref{sec:conclusion} concludes the paper.

\section{Related work}\label{sc:related}
Numerous surveys and documents on DTs have been undertaken across various disciplines, with a notable emphasis on manufacturing and industrial sectors. 
In the context of manufacturing, ISO 23247-1:2021 defined DT as ``fit for purpose digital representation of an observable manufacturing element with a means to enable convergence between the element and its digital representation at an appropriate rate of synchronization'' \citep{iso2020automation}. 
\cite{glaessgen2012digital} at NASA defined DT as ``an integrated multi-physics, multi-scale, probabilistic simulation of an as-built vehicle or system that uses the best available physical models, sensor updates, fleet history, etc., to mirror the life of its corresponding flying twin''.

Table \ref{table_definitions_review} presents a list of review studies conducted on DTs and the number of definitions of DTs in each study. Some of these surveys do not primarily focus on defining DTs but rather their history and sometimes mention one or more definitions that are widely recognized or commonly used. Some papers have, however, conducted surveys that, for a large part, focus on characterizing and defining the DT concept~\citep{SEMERARO2021103469,NEGRI2017939,OPOKU2021102726,VANDERHORN2021113524,DALIBOR2022111361}, but these are not focused on the BE domains.

\begin{longtable}{p{4.0cm}p{5.0cm}r}
  \caption{Digital Twin related literature review papers and the number of definitions in each.}
    \label{table_definitions_review} \\
    \toprule
    \textbf{Category} & \textbf{Ref} & \textbf{No. of definitions} \\
    \midrule
    \endfirsthead
    
    \multicolumn{3}{c}%
    {{\tablename\ \thetable{} -- continued from previous page}} \\
    \midrule
    \textbf{Category} & \textbf{Ref} & \textbf{No. of definitions} \\
    \endhead
    \endlastfoot
    
    \multirow{10}{4cm}{Manufacturing} &
    \cite{JONES202036} & 5 \\ \cline{2-3}
    & \cite{SEMERARO2021103469} & 30 \\ \cline{2-3}
    & \cite{he2021digital} & 6 \\ \cline{2-3}
    & \cite{NEGRI2017939} & {16} \\ \cline{2-3}
    & \cite{Bartsch2021} & 6 \\ \cline{2-3}
    & \cite{doi:10.1080/00207543.2021.1898691} & 1 \\ \cline{2-3}
    & \cite{app12115727} & 1 \\ \cline{2-3}
    & \cite{ERRANDONEA2020103316} & 3 \\ \cline{2-3}
    & \cite{SOORI2023100017} & 1 \\ \cline{2-3}
    & \cite{CIMINO2019103130} & 1 \\ \hline
    
    \multirow{8}{4cm}{Smart Cities, Construction, and Built Environment} &
    \cite{OPOKU2021102726} & {25} \\ \cline{2-3}
    & \cite{app11156810} & 4 \\ \cline{2-3}
    & \cite{chong2022digital} & 1 \\ \cline{2-3}
    & \cite{su2023digital} & 8 \\ \cline{2-3}
    & \cite{OSADCHA2023106704} & 4 \\ \cline{2-3}
    & \cite{buildings12020113} & 1 \\ \cline{2-3}
    & \cite{su13063386} & 1 \\ \cline{2-3}

    & \cite{depretre2022exploring} & 1 \\ \cline{2-3}
    & \cite{Jeddoub_2023} & 1 \\ \cline{2-3}
    & \cite{ellul2022location} & 1 \\ \cline{2-3}

    & \cite{isprs-archives-XLVI-4-W3-2021-315-2022} & 3 \\ \hline
    
    \multirow{4}{4cm}{Energy} &
    \cite{GHENAI2022102837} & 4 \\ \cline{2-3}
    & \cite{SIFAT2023100213} & 1 \\ \cline{2-3}
    & \cite{XIA2023113322} & 1 \\ \hline
    
    \multirow{3}{4cm}{Transportation} &
    \cite{9628341} & 7 \\ \cline{2-3}
    & \cite{MAURO2023113479} & 10 \\ \cline{2-3}
    & \cite{en14164919} & 1 \\ \hline
    
    \multirow{1}{4cm}{Healthcare} &
    \cite{lin2022human} & 6 \\ \hline
    
    \multirow{4}{4cm}{Other Industries} &
    \cite{yang2021developments} & 7 \\ \cline{2-3}
    & \cite{NGUYEN2022108381} & 1 \\ \cline{2-3}
    & \cite{10055149} & 1 \\ \hline
    
    \multirow{9}{4cm}{General Topics} &
    \cite{9327756} & 1 \\ \cline{2-3}
    & \cite{VANDERHORN2021113524} & 47 \\ \cline{2-3}
    & \cite{app11062767} & 5 \\ \cline{2-3}
    & \cite{ADAMENKO202027} & 1 \\ \cline{2-3}
    & \cite{wang2022review} & 3 \\ \cline{2-3}
    & \cite{Hassani2022} & 1 \\ \cline{2-3}
    & \cite{Wu2023} & 3 \\ \cline{2-3}
    & \cite{9576739} & 4 \\
\bottomrule    
\end{longtable}

The different implementations across domains also reflect the varied definitions of DTs. Different DT architectures or frameworks for different use cases conceptualize and specify the respective DT components to be implemented. \cite{grieves2014digital}~introduced the DT concept model in manufacturing that is composed of 3 main parts: 1) physical products (i.e.,\ in real space), 2) virtual products (i.e.,\ in virtual space), and 3) connections that link the virtual and real. \cite{qi2021enabling}~expanded this concept model into a five-dimensional DT model, presented to be a reference to understand and implement the enabling technologies and tools for DTs. Their model is composed of 1) physical entities-- which may be any device, product, physical system, or process; 2) virtual models-- which are the replicas of physical entities (including their physical geometries, properties, behaviors, and rules); 3) DT data-- referring to the heterogeneous data that drives the DT (static attributes or and dynamic data obtained from physical entities, as well as data generated by virtual models e.g.,\ from simulation); 4) services, which capture the paradigm of Everything-as-a-Service (XaaS), directing the applications and use case of the DTs; and 5) connections, which specifies the links that enable the other four parts to collaborate.

In urban applications, DT is seen as one of the enabling factors for digital services (i.e., smart cities) and for modeling of urban processes~\citep{deng2021systematic, raes2021duet}. Such DT implementations enable simulations that can be used for analysis and prediction through scenario development~\citep{li2023novel}. This technical aspect of digital twin of cities is oftentimes the scope of DT review~\citep{abdeen2023citizen, ZHANG2023107859, qi2021enabling, OSADCHA2023106704}, however, this perspective sometimes lead to its biased understanding~\citep{liu2023recognition}. The Digital Urban European Twins (DUET) project\footnote{https://www.digitalurbantwins.com/} is an example of an established DT initiative for cities that promotes the adoption of local and urban DTs. The technical report on the architecture of the DUET system~\cite{DUET_2020} presented the overall architecture of their platform and the specifications of the DUET system in terms of the functionalities of the respective processes and the corresponding components that realize them. The DUET architecture refers to the system’s components as multiple distinct layers: 1) presentation layer-- contains all user interfaces and visualization modules; 2) access control layer-- contains all necessary \textit{Gateway} components that control data flow; 3) service layer-- exposes the functionalities of core components through a set of APIs; 4) business layer-- tackles user needs; 5) data layer-- consists of all the heterogeneous data repositories, including information about data sources, ontologies, users, and other metadata; 6) infrastructure layer-- consists of physical or virtual servers, and other relevant supporting tools; and, 7) security layer-- implements necessary security measures for all layers.

The concept of DTs continues to gain attention across different fields, particularly for its ability to continuously analyze incoming (i.e., real-time) data and perform data-driven simulations~\citep{hananto2024digital, lei2023challenges}. As different domains recognize the potential of DTs and move toward implementations for various applications, standards play an important role in the creation of DTs, especially concerning data from different sources for use in various applications~\cite{ellul2024towards}. For complex systems such as cities, the practical approach of digital twinning would be to begin with priority areas and scale over time which would require common frameworks for effective communication~\cite{raes2021duet}.

Defining standards---which may relate to one or several elements of a DT and of its construction--- is the scope of ongoing work~\citep{wang2022review}. Recently, the joint technical committee of the International Organization for Standardization (ISO) and International Electrotechnical Commission (IEC) published the standard \textit{ISO/IEC 30173:2023, Digital twin - Concepts and terminology} which presents general descriptions of DT concepts, terminology and application overviews and aims to provide a common basis for understanding its concept and composition and be applicable in any and across domains~\citep{ISO30173_2023}. In the standard, a DT system is described as a \textit{hybrid entity} or \textit{system of systems}, with the DT elements illustrated within this context, which include the Target entity, which refers to something (physical or non-physical) having a distinct existence and is the subject of digital representation; the Digital entity, which is the computational entity comprised of data and procedural elements; and the Data connection between the two, and corresponding models, data and interfaces (e.g., APIs) involved---mirroring the DT concept model of~\cite{grieves2014digital}. Beyond these three, the standard further describes the other elements of the DT system as the Services, which refer to the capabilities (e.g.\ analysis, simulation, visualization, optimization, etc.) of the DT, the Infrastructure, which include the supporting components to ensure the DT's operation (e.g., sensors, network, GIS, etc.) and the System aspects, which are concerned with the integration, governance, security, and privacy, and data ethics involved in the DT implementation. However, there has yet to be a DT reference model that unifies the \emph{ingredients} necessary to build a DT~\citep{raes2021duet}. Thus, alongside unifying definitions, it is equally important to identify and unify the enabling components of DT development and implementation, especially toward goals of interoperability and scalability. 

While several academic reviews have attempted to identify or define DTs, these definitions are often aligned with specific research domains or adapted to fit particular applications. This multiplicity has resulted in a variety of definitions, some of which accurately represent the concept of a DT, while others do not and diverge from the mainstream. There is no single definition that fits all domains, and most existing definitions are meant to be universally applied. However, our research aims to redefine DTs on the basis of the researchers' interpretation of the term within each specific domain, and we also challenge existing instances of (ostensible) DTs as we noticed papers that attempt to forcibly fit certain instances into the DT concept.
In doing so, we evaluate and establish definitions for each domain individually based on the essential components unique to each field and its users.

\section{Methodology}
\label{sec:methodology}
A coarse level of agreement exists on what qualifies as a Digital Twin, though the threshold between what a DT is or is not can be quite nebulous across different applications. To address this issue, we aim to accumulate as many insights as possible about interpretations of DTs from the current state-of-the-art. We then define the boundary between what is considered a DT and what is not, driven by data and statistical analysis. 

\begin{figure}[h]
    \centering
    \includegraphics[width=\linewidth]{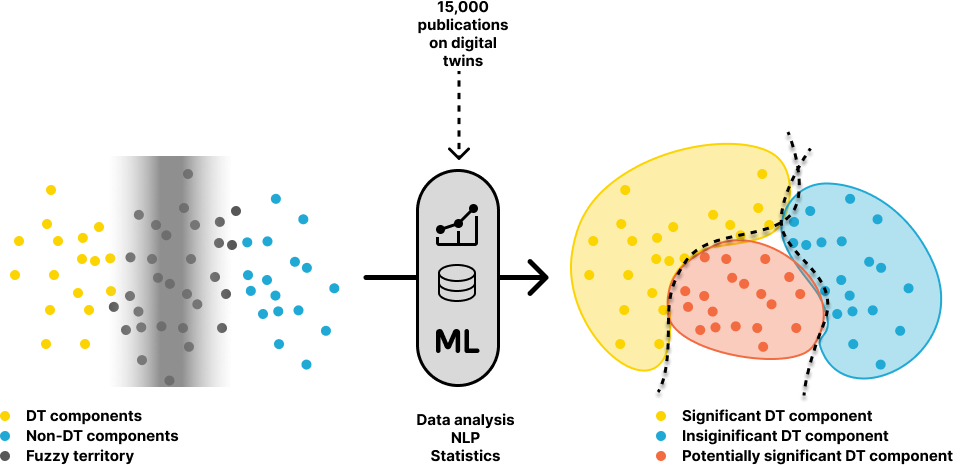}
    \caption{Our novel approach to infer a DT definition consists of deriving significance of each DT component from literature using Natural Language Processing (NLP) and statistical analysis.}
    \label{fig:novelty}
\end{figure}

Our approach leverages the credibility of existing literature, which has often undergone double-blind reviews by experts in various domains. This lends validity to several studies that qualify as a Digital Twin, even if they do not match standardized definitions. Accordingly, our methodology is grounded in both empirical observations and the extensive body of existing literature, combining theoretical insights with practical validation (Figure \ref{fig:novelty}). To validate our outcome, we conducted an expert survey from a panel of 52 experts in the BE DT domain. Additionally, we considered different maturity stages of DTs as defined by \cite{10367836} and \cite{HARAGUCHI2024123409}.

\begin{figure}[!h]
  \includegraphics[width=\linewidth]{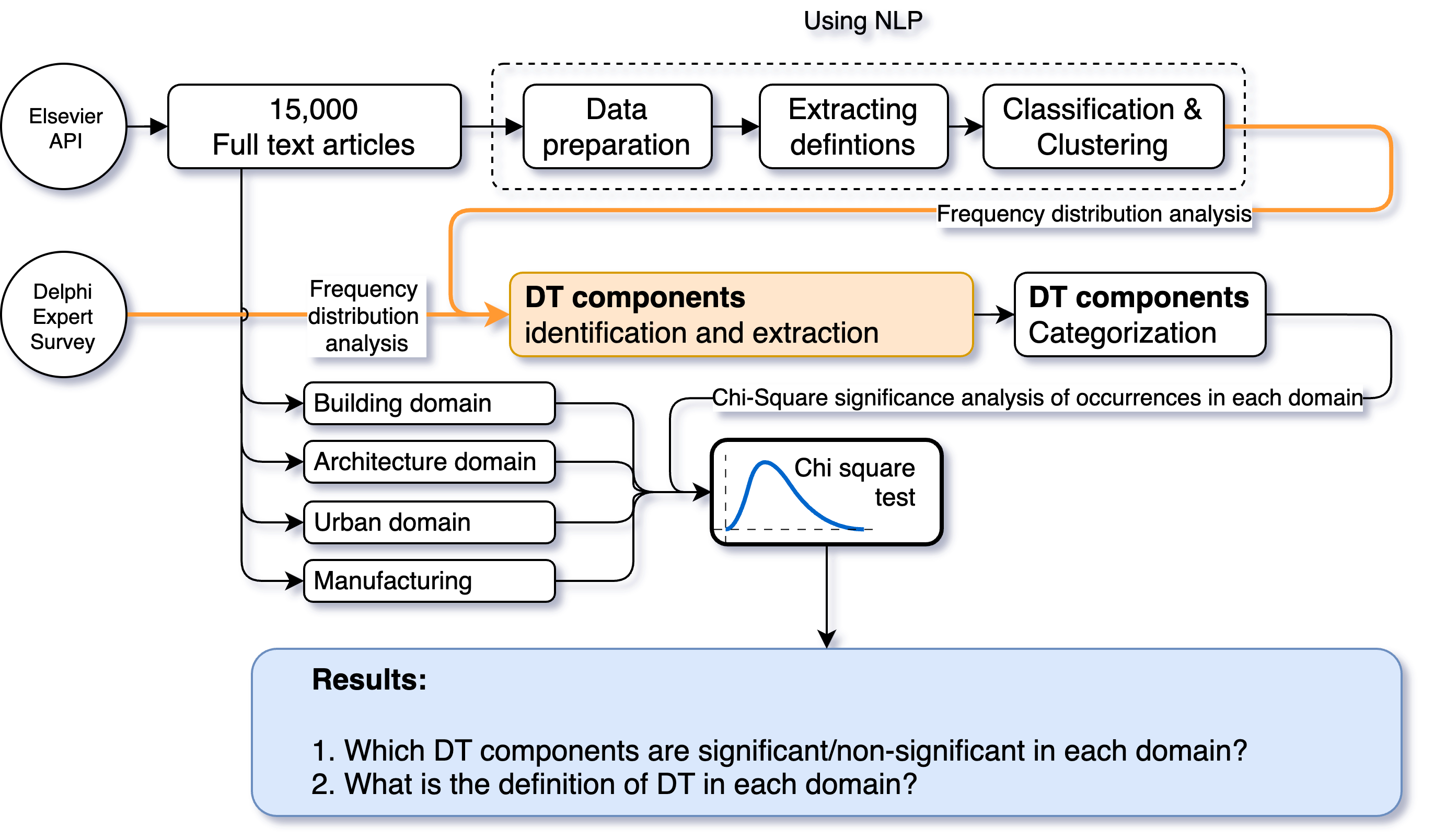}
  \caption{Overview of the methodology.}
  \label{fig:methodology}
\end{figure}

\begin{figure}[!h]
  \includegraphics[width=0.9\linewidth]{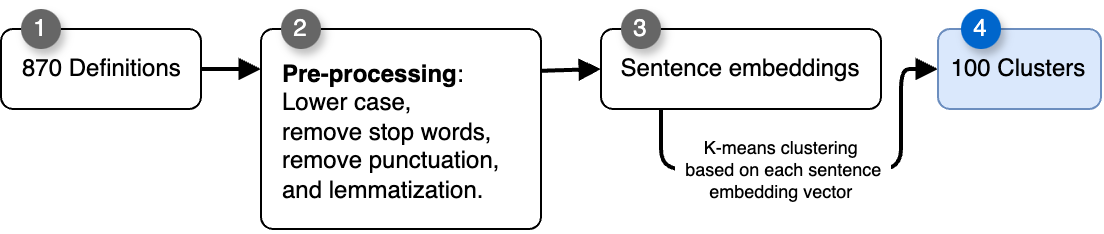}
  \caption{The flowchart represents the definition of extraction and clustering methods.}
  \label{fig:data_clustering}
\end{figure}

In this section, we first present our approach to identifying the sample from various resources, including full-text publications and the expert survey. Then, we outline the methods used to extract and process approximately 800 definitions and interpretations of DTs from this sample. Our procedure includes a variety of NLP methods such as Regular Expressions (Regex) \citep{li2008regular}, Sentence Embeddings \citep{le2014distributed}, Frequency Analysis and N-gram analysis \citep{cavnar1994n}, and Fuzzy Matching \citep{chaudhuri2003robust}. Additionally, we employ Machine Learning approaches to group and cluster semantically similar definitions using \textit{k}-means clustering and Large Language Models (LLMs). After that, we describe the methods used in extracting the DT components from different scales within the Architecture, Engineering, and Construction domain; namely, Building, Architecture, and Urban scales;  in addition to Manufacturing domain as a dominant domain in this area. We used word cloud analysis and LLMs for that purpose. Then we test the significance of each component using Chi-square analysis. We maintained our methodology as systematic as possible to ensure it is generalizable to other research domains, which may also benefit from standardizing and analyzing terminology. Thus, the pipeline and code have been openly released on GitHub\footnote{https://github.com/ualsg/dt-definitions}. Figures \ref{fig:methodology} and \ref{fig:data_clustering} summarize the methodology used in the current study.

\subsection{Sample Development}
\label{subsec:sampleselection}

This study is built on two main data sources. The first consists of \hyperref[ds1]{1) full-text articles}, while the second involves a \hyperref[ds2]{2) expert survey} which is used to validate the results from a Built Environment expert perspective. From these two datasets, we have derived: the \hyperref[ds3]{3) DT definitions dataset} and the \hyperref[ds4]{4) DT components dataset}. In this section, we overview our methodology of collecting the sample data and extracting the definitions' and components' datasets that are used for analysis. We keep our methodology as detailed as possible for the purpose of reproducibility and reusability across different domains and studies.

\subsubsection{Full-text articles dataset}
\label{ds1}

\begin{figure}[!h]
    \centering
    \includegraphics[width=\linewidth]{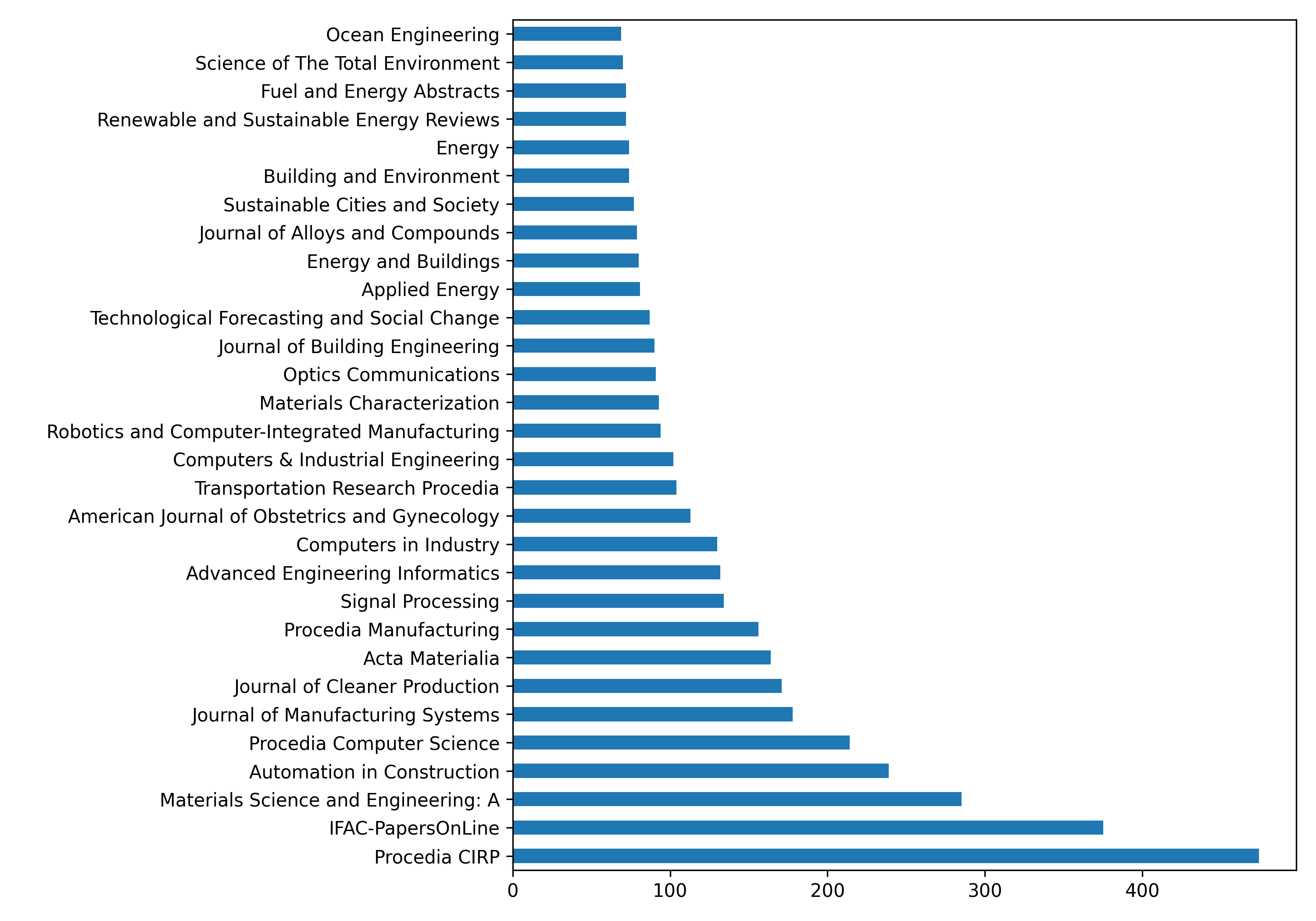}
    \caption{Most common publication sources in the dataset, spanning multiple communities, scales, and domains.}
    \label{fig:pub_names}
\end{figure}

\begin{figure}[!h]
    \centering
    \includegraphics[width=\linewidth]{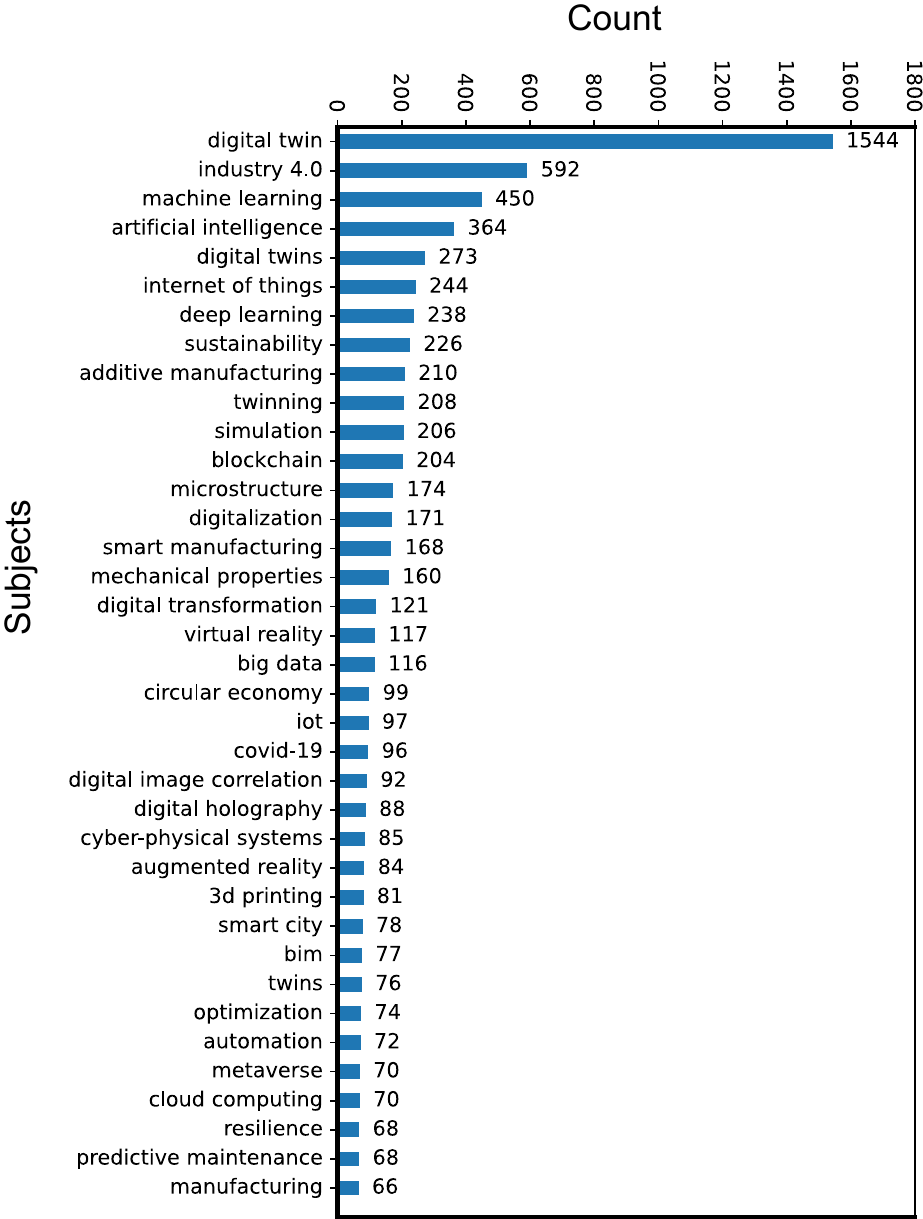}
    \caption{Distribution of subjects in the selected dataset. The total search returned 15,353 results; however, this figure represents a subset of articles with valid subject metadata relevant to the analysis after cleaning the data.}
    \label{fig:subject_dist}
\end{figure}

Full-text samples were acquired using an Application Programming Interface (API) provided by Elsevier\footnote{https://dev.elsevier.com/}, which provides interfaces to Scopus, Science Direct, and other Elsevier products. We chose to focus on the Elsevier API because it is one of the largest publishers of scientific literature and provides a comprehensive database of peer-reviewed literature across myriads of fields, thus well capturing the body of knowledge without bias. We used the API to search for articles containing the following terms: ``Digital Twin'', ``City Digital Twin'', ``Urban Digital Twin'', ``Smart City'', and ``Building Digital Twin'' in the title, abstract, or keywords. The search was conducted in English and covered articles published between 2000 and 2024. The search returned a total of 15,353(Figure \ref{fig:subject_dist}). The dataset was also collected from various sources (Figure \ref{fig:pub_names}), notably the top are from manufacturing (e.g, IFAC Journal of Systems and Control, Journal of Manufacturing Systems, Procedia Manufacturing); and the Built Environment (e.g, Automation in Construction, Transportation Research Procedia, Journal of Building Engineering, Energy and Buildings, and Building and Environment). %

\subsubsection{The expert survey dataset}
\label{ds2}
We compare the results of systematic review to a question --- `What is a digital twin?', collected in a Delphi survey distributed to experts. The definitions of DTs vary from different fields, making it challenging to achieve consensus in the built environment. 
Therefore, we built on top of a previously conducted Delphi survey by \citet{lei2023challenges} to gather insights from practitioners actively engaged in this line of research. In this work, the collected responses of DT definitions are used to understand different perspectives in practice, from industries to governments.
We incorporate the responses as a validation dataset, aiming to evaluate and compare with the developed corpus of research articles, thereby providing insights into the understandings of DTs. We deem that the results of this specific question can contribute to a practical perspective to supplement the analysis of the literature review and, meanwhile, detail differences between key components and focus.

The expert survey invited a structured panel of experts, who have domain expertise in DT related to the Built Environment, aiming to generate a consensus on the definition of DTs. A total of 52 international experts from different sectors, such as government, industry (e.g.\ companies, consultancies), and non-government organizations (NGOs), were selected to build the panel, contributing with diverse backgrounds and experiences in this topic. In particular, the panel of domain experts represents 23 countries, with the majority working in Europe. Of these participants, 14 were from industry, 12 were from government, and the remaining 23 were affiliated with universities or research institutes. Notably, 44\% of respondents reported over 20 years of experience in fields integral to digital twins. Through an online questionnaire, all participants were asked to identify a digital twin based on their understanding and practice.

\subsubsection{The definitions dataset}
\label{ds3}
We used Regular Expressions (regex) to extract the mentions of DTs from the dataset, based on the reasoning that a subset of instances will include definitions. Regex has the advantage of extracting patterns of sentences instead of searching for an exact word. For example, among other regex patterns, we used the following pattern to extract definitions: 

\begin{lstlisting}[language=Python, basicstyle=\small\ttfamily, linewidth=\textwidth, breaklines=true, frame=single]
r'\b(?:digital twin(?:s)?)\s+(?:is\s+)?(?:defined\s+as)\s+.+?(?=\.\s|\n|$)'
\end{lstlisting}

This pattern extracts all sentences that match ``digital twin/s (is/are/can be/could be) (defined as/described as/ characterized by) ...''. %
Many of the extracted definitions are valid, and some are not valid definitions such as: 
``Digital Twin is a revolutionizing technology in the industry 4.0 era'' and ``Digital Twin is defined as a new paradigm in simulation''. %
Therefore, we applied NLP and LLMs to filter the dataset, aiming to address two main issues: \emph{Incomplete} and \emph{Fuzzy Duplicated} definitions.

To improve the quality of the dataset, the filtering process was designed to address these two issues. \emph{Incomplete Definitions (ID)} refer to descriptions, explanations, or specifications that lack the necessary or essential details to fully define or explain the DT concept \citep{gantayat2014study}. Examples include: ``Digital twin is a revolution in the digital industry''~\citep{SINGH2023109711}, ``digital twin is defined as a virtual replica of a real system'' \citep{FARHANHUSSAIN20231}, ``digital twin is defined as a digital representation of assets, processes, or systems'' \citep{YEUNG2022103957}, and ``digital twin is a living model that drives a business outcome'' \citep{ELSADDIK20231}.

\emph{Fuzzy Duplicated Definitions (FDD)} refer to instances where there are multiple similar or nearly identical definitions or descriptions, but with slight variations or differences \citep{shahri2004flexible}. 

To perform the filter, we first excluded all the incomplete definitions using LLMs, namely, the Generative Pre-trained Transformer (GPT4) model. Then, we manually validated a subset of the results.

Secondly, to overcome the FDD issue, we first clustered the definitions using sentence embeddings \cite{reimers2019sentence} and \textit{k}-Means clustering into three iterations of clustering 400, 100, and 50 clusters. The Sentence Embeddings first convert all the definitions into a lower dimensional vector where the position of each definition in the vector space represents its semantic meaning. By doing this, the closer the definitions in the vector space, the more semantically similar these sentences are to each other. Each sentence is represented as a feature vector $\vec{X}_d=[x_0, x_1, x_2, \ldots, x_n]$ where $X_d$ is the definition sentence vector. We used the cosine similarity distance to measure the similarity between different definitions in the definition dataset. The cosine similarity distance between two vectors $\mathbf{X}$ and $\mathbf{Y}$ is given by:

\begin{equation}
\text{cosine\_similarity} (\mathbf{X_{d1}}, \mathbf{X_{d2}}) = \frac{\mathbf{X_{d1}} \cdot \mathbf{X_{d2}}}{\|\mathbf{X_{d1}}\| \cdot \|\mathbf{X_{d2}}\|}
\end{equation}

where:
\begin{itemize}
    \item $\cdot$ represents the dot product of two vectors.
    \item $\|\mathbf{X_{d1}}\|$ and $\|\mathbf{X_{d2}}\|$ represent the Euclidean norms (lengths) of the vectors $\mathbf{X_{d1}}$ and $\mathbf{X_{d2}}$ respectively.
\end{itemize}

After clustering the definitions into groups, we applied the Fuzzy Matching algorithm within each group to figure out how each pair of definitions matches. By doing so, we excluded semi-duplicated definitions. For example, the following two definitions are 92\% matching while the difference is only in the words in italics (``virtual representation'' and ``digital replica''):
\begin{itemize}
    \item ``Digital Twin is a \textit{virtual representation} of real-world entities and processes, synchronized at a specified frequency and fidelity'' which was introduced by in the Digital Twin Consortium in 2020 \citep{digitaltwin_con2020}.
    \item ``Digital Twin is a \textit{digital replica} of real-world entities and processes, synchronized at a specified frequency and fidelity'', which was found in the reference \citep{TRANTAS2023102357}.
\end{itemize}

The fuzzy matching algorithm between two definitions can be represented as follows:
\begin{equation}
\text{Fuzzy\_Match}(s_1, s_2) = 1 - \frac{\text{Levenshtein\_Distance}(s_1, s_2)}{\max(\text{length}(s_1), \text{length}(s_2))}
\end{equation}

where:
\begin{itemize}
    \item $s_1$ and $s_2$ are the strings being compared.
    \item $\text{Levenshtein\_Distance}(s_1, s_2)$ calculates the minimum number of single-character edits (insertions, deletions, or substitutions) required to change one string into the other.
    \item $\text{length}(s)$ represents the length of string $s$.
\end{itemize}

Table \ref{tab:definitions_example} shows a list of some definitions that were extracted from the dataset. The rest of definitions can be found on the GitHub repository\footnote{https://github.com/ualsg/dt-definitions}.

\begin{table}[!ht]
\caption{Ten examples of extracted DT definitions. The rest of definitions can be found on the GitHub repository}
\label{tab:definitions_example}
\centering
\begin{tabular}{p{10cm}p{3cm}}
\toprule
 \textbf{Definition} & \textbf{Ref.} \\
\toprule
Digital Twin is defined as an integrated multi-physics, multi-scale, probabilistic simulation of a vehicle or system that uses the best available physical models, sensor updates, fleet history, and so forth, to mirror the life of its physical twin. & \cite{shafto2010draft} \\
\hline
Digital Twin is defined as a software representation of a physical asset, system or process designed to detect, prevent, predict and optimize through real time analytics to deliver business value. & \cite{hribernik2013towards} \\
\hline
Digital Twin is a highly dynamic concept growing in complexity along the product life cycle, which is more than a pool of digital artifacts; Digital Twin has a structure consisting of connected elements and meta-information as well as semantics. & \cite{ROSEN2015567} \\
\hline
Digital Twin is a set of virtual information constructs that fully describes a potential or actual physical manufactured product from the micro atomic level to the macro geometrical level. & \cite{Grieves2017} \\
\hline
Digital Twin is a collection of all digital artifacts that accumulate during product development linked with all data that is generated during product use. & \citet{HAAG} \\
\hline
Digital twin is a living model of the physical asset or system, which continually adapts to operational changes based on the collected online data and information and can forecast the future of the corresponding physical counterpart. & \cite{gohari2019digital} \\
\hline
Digital twin is a digital representation of a physical asset reproducing its data model, behavior, and communication with other physical assets. & \cite{kamel2021digital} \\
\hline
Digital twin is a detailed simulation model of a system, for which application-specific data is stored to describe the real twin. & \cite{meier2021digital} \\
\hline
Digital twin is a dynamic representation of a physical system using interconnected data, models, and processes to enable access to knowledge of past, present, and future states to manage action on that system. & \cite{pinto2022reality} \\
\hline
Digital twin is a real-time, virtual replica of a physical object or system created using sensors and computational models. & \cite{VO2024107643} \\
\hline
\end{tabular}
\end{table}

\subsection{Digital Twin components identification}
\label{ds4}

As shown in the examples in Table \ref{tab:definitions_example}, each definition is comprised of a set of key components. For instance, the first one highlights components like ``multi-physics,'' ``multi-scale,'' ``probabilistic simulation,'' ``physical model,'' and ``sensors.'' Meanwhile, the next example introduces other elements such as ``software representation,'' ``predict and optimize,'' and ``real-time analytics''.

\begin{figure}[!ht]
    \centering
    \includegraphics[width=\linewidth]{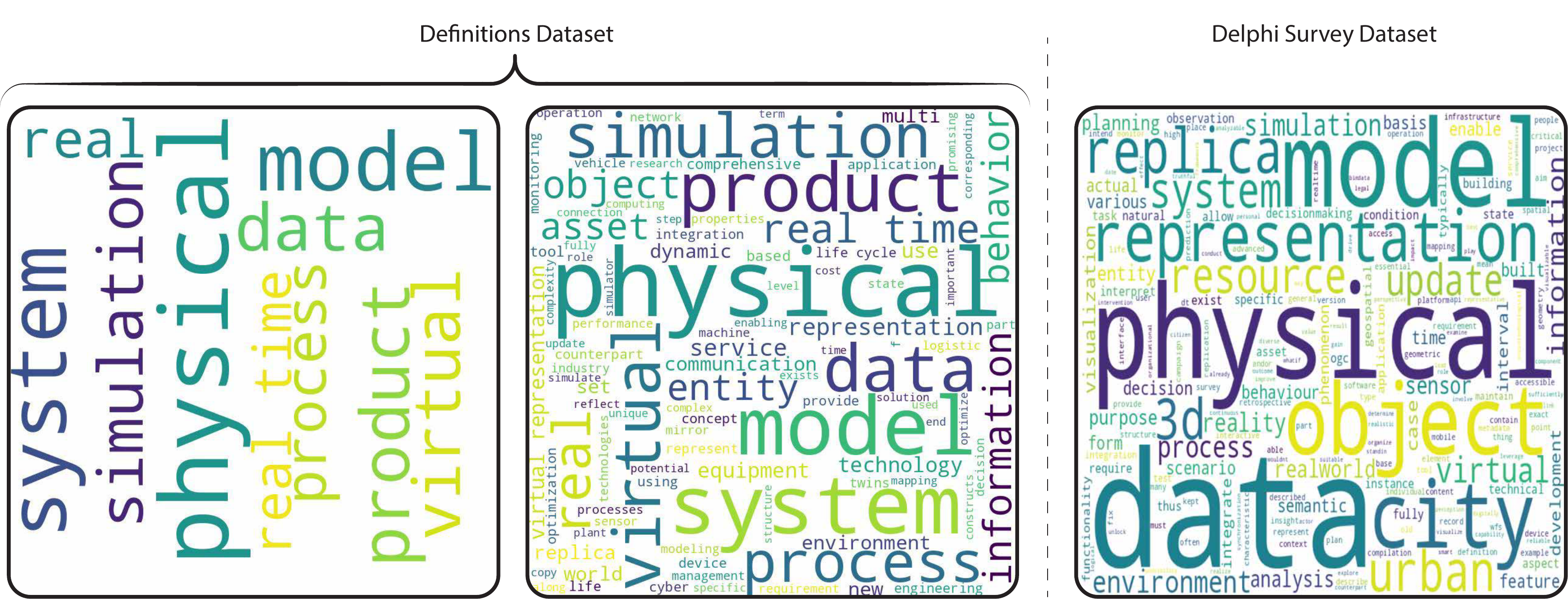}
    \caption{Word cloud visualization of word frequency analysis of the extracted definitions dataset (on the left) and the expert survey (on the right).}
    \label{fig:freq_wordcloud}
\end{figure}

To systematically identify the components of digital twins, we first conducted a Frequency Distributions Analysis from literature and experts, this process involves counting the occurrences of terms and phrases associated with digital twins in the extracted definitions dataset and within the survey dataset. The analysis provides numerical data on the most frequently referenced components and concepts, highlighting prevalent elements and revealing emerging trends and gaps in current digital twin architectures. Figure~\ref{fig:freq_wordcloud} illustrates the word cloud frequency analysis from the extracted definitions (on the left) and the expert survey (on the right).

We can observe that the top words identified in the frequency analysis in the definitions dataset closely correspond to the frequency analysis within the expert survey dataset, with a few differences, as seen in Figure~\ref{fig:freq_wordcloud}. For example, while `physical', `model', and `data' are frequently mentioned in both the research articles and the survey, experts put more focus on `3D' and `update' and researchers highlight the function of `simulation' in DTs. These differences reflect the influence of varying perspectives across domains. However, an equivalence can be seen, considering the ideas of a DT implementation, for example `update' from the expert survey could relate to the idea of `real-time', or `virtual' from the research articles may be synonymous to the `representation' that experts refer to. 

We can further note that the the top words identified have agreement and are closely captured within the ISO 30173 DT system elements descriptions and DUET architecture specifications described in Section~\ref{sc:related}. This correspondence of terms allows us to identify generalized DT components for conducting further analysis on the dataset. We refer to top words identified from the frequency analysis word cloud and unify with related components as described in the DUET system architecture, ISO 30173 and Five-dimensional reference model to create a list of representative DT components, grouped into 5 generalized categories as summarized in \autoref{tab:dt_compo_rev}, and described in more detail in the continuation.

\begin{table}[!h]
\centering
\caption{Generalized digital twin components.}
\label{tab:dt_compo_rev}
\begin{tabular}{@{}lll@{}}
\toprule
\textbf{Category} & \textbf{Description} & \textbf{Related   component} \\ \midrule
\multirow{3}{*}{Data} & \multirow{3}{*}{\begin{tabular}[c]{@{}l@{}}Data representations \\ of both the target entity and \\ the digital entity.\end{tabular}} & 2D and 3D data \\ \cmidrule(l){3-3} 
 &  & Real-time data \\ \cmidrule(l){3-3} 
 &  & Data modeling \\ \midrule
\multirow{4}{*}{Analysis and services} & \multirow{4}{*}{\begin{tabular}[c]{@{}l@{}}Setup for \\ modelling and analysis towards \\ specific services, application \\ use case and scenarios.\end{tabular}} & Simulation models \\ \cmidrule(l){3-3} 
 &  & Data analytics and AI/ML \\
 &  & models \\ \cmidrule(l){3-3} 
 &  & Data Catalogue \\ \midrule
\multirow{6}{*}{Infrastructure} & \multirow{6}{*}{\begin{tabular}[c]{@{}l@{}}Software and \\ hardware components, and \\ other relevant tools supporting \\ the operation of the digital \\ twin system.\end{tabular}} & Cloud platform and \\
 &  & architecture \\ \cmidrule(l){3-3} 
 &  & High-Performance \\
 &  & Computing (HPC) \\ \cmidrule(l){3-3} 
 &  & Internet of Things (IoT) and \\
 &  & sensor network \\ \midrule
\multirow{4}{*}{Interface} & \multirow{4}{*}{\begin{tabular}[c]{@{}l@{}}Requirements and \\ setup for connection and \\ interaction among digital twin \\ components, as well as with users.\end{tabular}} & Application Programming \\
 &  & Interface (API) \\ \cmidrule(l){3-3} 
 &  & Visualization \\ \cmidrule(l){3-3} 
 &  & Dashboards \\ \midrule
\multirow{5}{*}{System governance} & \multirow{5}{*}{\begin{tabular}[c]{@{}l@{}}System and design \\ characteristics of the digital \\ twin, relating to integration, \\ governance, security and \\ privacy, and ethics.\end{tabular}} & Data validation \\ \cmidrule(l){3-3} 
 &  & Security protocols \\ \cmidrule(l){3-3} 
 &  & Policy (and ethical protocols) \\ \cmidrule(l){3-3} 
 &  & User management and \\
 &  & administration \\ \bottomrule
\end{tabular}
\end{table}

\begin{enumerate}
    \item \textbf{Data}
    
    DTs are envisioned to deal with multi-temporal and -scale heterogeneous data, which could be from multiple sources, either obtained from physical entities (e.g., as attribute data or dynamic condition data from sensors) or generated by virtual models (e.g., results from simulations within the DT) in both directions (Bi-directional) either from the physical twin to the virtual twin or vice versa using actuators or using human in the loop. We generalize three related components under this category:
    
    \begin{itemize}
        \item \textit{2D and 3D data:} The data (e.g.,\ physical geometries, properties, behaviors, and rules) corresponds to either the target entity or digital entity.
        \item \textit{Real-time data:} Real-time data updates (i.e.,\ of the target entity as facilitated by sensors and Internet of Things) or dynamic data outputs from real-time simulation process. Also, this includes data that are transferred the other direction, i.e, from the virtual twin to the physical twin via actuators or human-informed-decision in the loop. 
        \item \textit{Data modeling:} Formalized and standardized digital representation of data, including definitions for data models, schema, ontologies, and digitization methods. 
    \end{itemize}

    \item \textbf{Analysis and services}
    
    In specific application domains, the DT system may be designed with capabilities to provide specific services (e.g.,\ integration, analysis, simulation, visualization, and optimization). We use the three general representative components for this category as follows:  
    
    \begin{itemize}
        \item \textit{Simulation models:} Implemented modeling of the specific processes (or scenarios) for representing the target entity or system.
        \item \textit{Data analytics and AI/ML models:} Functionalities of the DT for running data analytics, AI, and ML algorithms and models, i.e., implemented for specific use cases or services.
        \item \textit{Data Catalogue:} Relate to DT data management, including setup for data discovery and data access, e.g.,\ knowledge graphs and context graphs. 
    \end{itemize}

    \item \textbf{Infrastructure}
    
    Infrastructure is a major component supporting the DT to ensure its normal operation, which could include a slew of different hardware components as required for specific use cases. We include the three below for our analysis: 
    
    \begin{itemize}
        \item \textit{Cloud platform and architecture:} The technological components specifically set up for the DT in utilizing cloud environments for services.
        \item \textit{High-Performance Computing (HPC):} DT component for super-computing, i.e.,\ to handle huge volumes of data and complex models at high speeds.
        \item \textit{Internet of Things (IoT) and sensor network:} The network of physical devices interconnected and communicating with each other for the collection and sharing of data.
    \end{itemize}

    \item \textbf{Interface}
    
    Within the DT system, digital entities are connected dynamically with their physical counterparts. Data connections enable data exchange between entities and link the DT's models and services. Further, the DT is expected to have a platform for the user to get information and interact with the data and models. 
    
    \begin{itemize}
        \item \textit{Application Programming Interfaces (APIs):} Software implementation to facilitate the access and control of data flow between components (i.e., as data gateways), as well as between DT and user (i.e., through dashboards or other apps).
        \item \textit{Visualization:} The functionality of the DT for providing data presentation and other visual outputs.
        \item \textit{Dashboards:} The platform (and tools) for data and visualization display, and for DT user interface (i.e.,\ through web applications) that facilitate the users' access to the DT data, models and outputs. 
    \end{itemize}
    
    \item \textbf{System governance}
    
    Aspects of integration, governance, security and privacy, and data ethics are requirements for a DT system to coexist with other DT systems)~\citep{ISO30173_2023}, for example, matters related to data collection and access. Such system aspects could include (but are not limited to) the following:
    
    \begin{itemize}
        \item \textit{Data validation:} Refer to control measures for data checks to ensure the meeting of required data quality and accuracy.
        \item \textit{Security protocols:} Protocols specific to the prevention of risks (e.g.,\ hacks, data leaks, etc.) and securing of the DT's operation.
        \item \textit{Policy:} Includes policies in place to ensure ethical use and management of the DT data and products.
        \item \textit{User management and administration:} Protocols and provisions for implementing user authorization and access, consistent with security and policy. 
    \end{itemize}
    
\end{enumerate}

\subsection{Component Identification and Analysis}
\begin{figure}[!h]
    \centering
    \includegraphics[width=\linewidth]{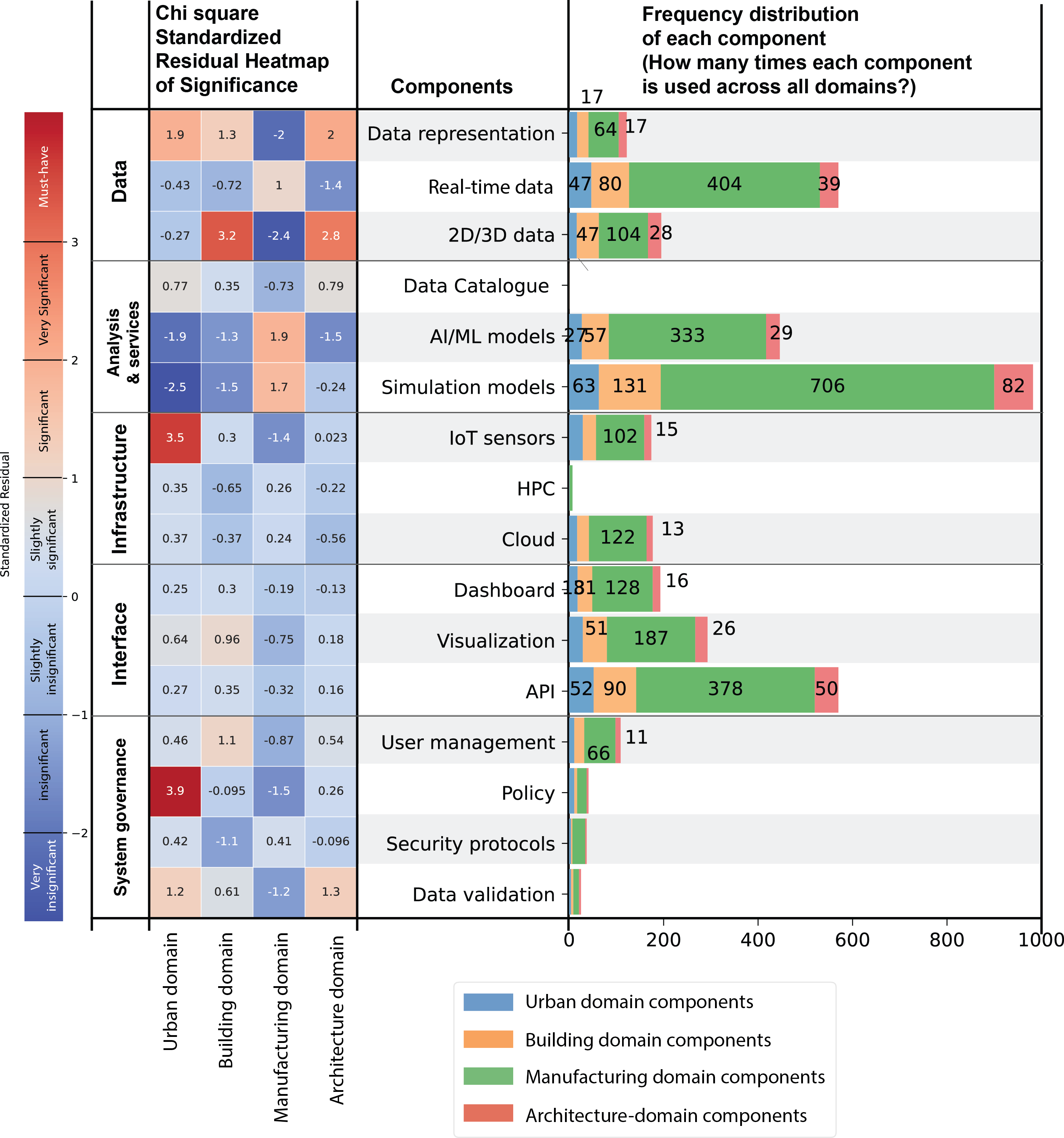}
    \caption{Chi-square frequency distribution heatmap across different domains and component frequency distribution across different domains.}
    \label{fig:component_frequency_analysis}
\end{figure}

\begin{figure}[!h]
    \centering
    \includegraphics[width=\linewidth]{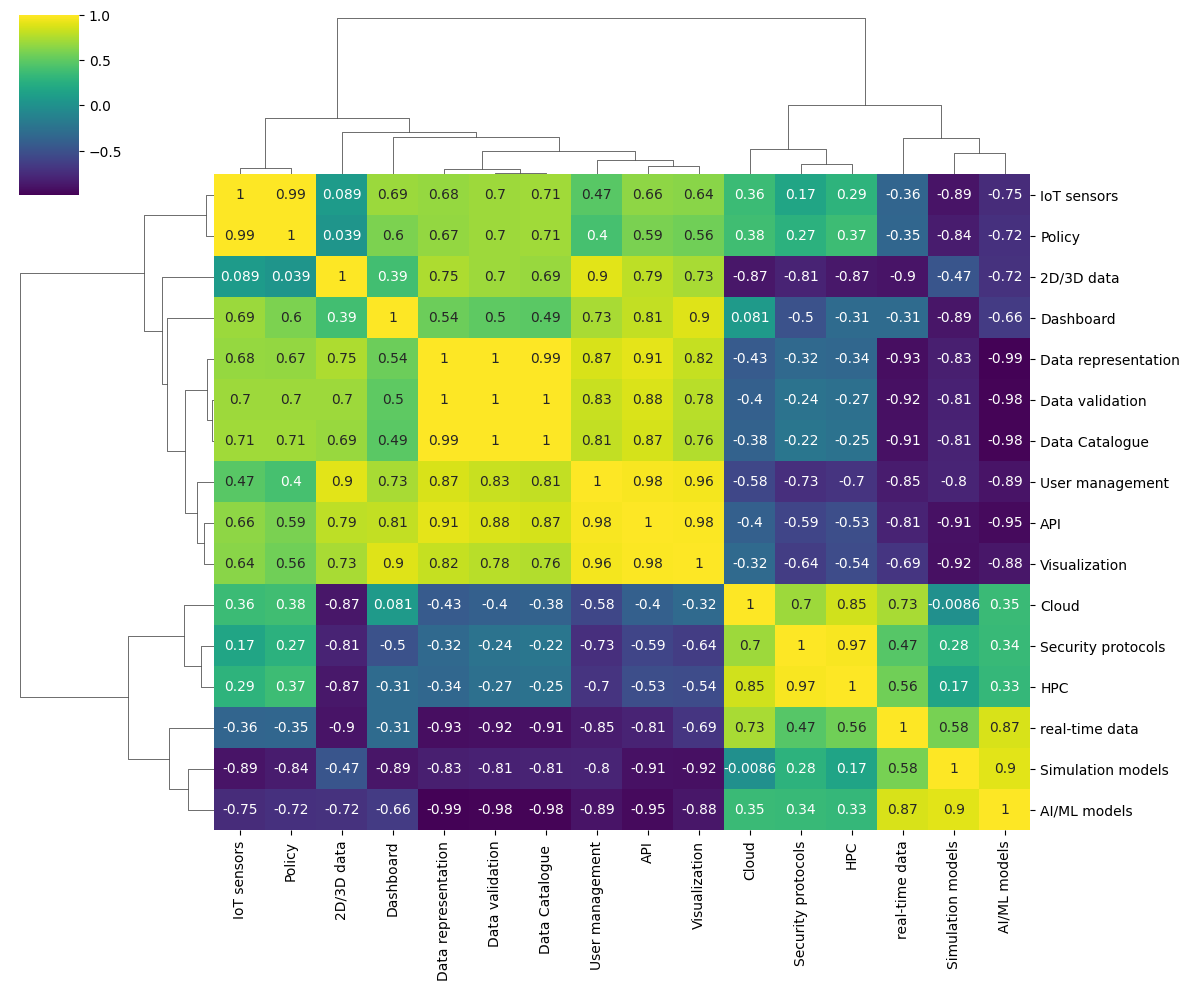}
    \caption{Clustered correlation between components residuals across all the domains. The higher correlation between two components means that these two components have either significant or non-significant residual.}
    \label{fig:clustered_correlation_all_domains}
\end{figure}

We conducted component frequency and Chi-square analysis to examine the association between DT components (Table \ref{tab:dt_compo_rev}) and various domains, including urban, building, manufacturing, and architecture. The use of Chi-square test is particularly useful in our study due its ability to handle categorical data. The significance was tested against the null hypothesis $H_0$ (No significant contribution) and the alternative hypothesis $H_1$ (There is a significant contribution) for each component. The chi-square statistic of 109.76 and a p-value of 2.48e-07 indicated a significant relationship between the components and their respective domains. We calculated the Standardized residuals to identify specific deviations, with significant residuals (absolute values greater than 2) highlighting components that were either over-represented or underrepresented. Notably, ``IoT sensors'' and ``Policy'' were significantly more frequent in the urban domain, while ``2D/3D data'' showed higher frequencies in the building and architecture domains but lower in manufacturing. This analysis, including adjustments for zero and low values, provided a detailed understanding of the distribution and significance of digital twin components across different fields, offering critical insights for the development and application of DT in various sectors (Figures \ref{fig:component_frequency_analysis}). We also performed a correlation and dendrogram clustering analysis between all component residuals to explore which components are joined in terms of significance and insignificance across all domains (Figure \ref{fig:clustered_correlation_all_domains}). The analysis clearly divides these components into two big clusters of correlations.

To extract the definitions across domains, we compared the definitions from the existing dataset (800 definitions) and the expert survey to the components analysis for each domain. LLM was used to extract the components from these definitions and compare them with the components identified through frequency analysis. 

\section{Results}\label{sec:results}

\subsection{Component Analysis Across Domains}
\label{subsec:component_identification}
Based on the Chi-square test and the correlation clustering of components, it is noticed that Cloud computing, HPC, Security protocols, Real time data, simulation models, and AI/ML models are highly correlated with each other and can be segregated into one group of components. On the other hand, Data representation, Data validation, Visualization, and Policy are highly correlated with each other and can be segregated into another group of components. 
We refer to the first group of components as HPRT (High-Performance Real-Time) Digital Twins, characterized by real-time, big-data-driven, dynamic, and adaptable components. The second group is referred to as LTDS (Long-Term Decision Support) Digital Twins, characterized by near-to-far real time data management and visualization components. The first group of components is used in computationally intensive applications, such as manufacturing, self-driving cars, and aerospace and rarely used in the BE and AEC domains. In contrast, the second group is suited for applications requiring long-term decision support, such as urban planning and building management.

\begin{figure}[!h]
    \centering
    \includegraphics[width=\linewidth]{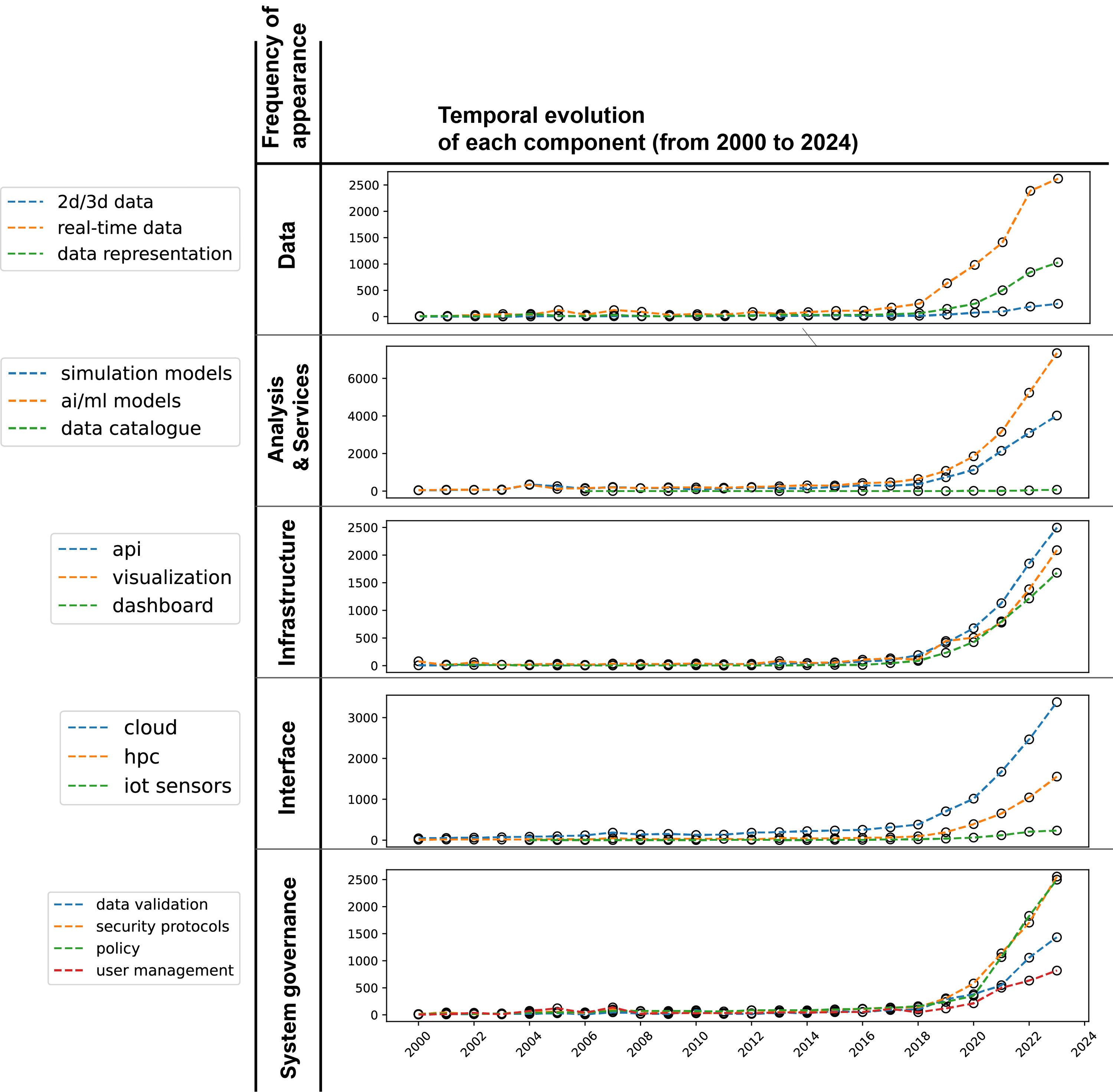}
    \caption{Temporal evolution of components from 2000 to 2024}
    \label{fig:temporal_evolution_of_componetns}
\end{figure}

The temporal analysis further highlights the evolving nature of these components (Figure \ref{fig:temporal_evolution_of_componetns}). While some components like AI/ML and real-time data capabilities have seen rapid growth in adoption due to advancements in technology and increasing data availability, others like policy and data representation have remained consistently significant due to their foundational role in ensuring the accuracy and applicability of digital twins in long-term planning and management contexts. 

The temporal evolution also shows some interesting insights about the adoption of some components; for example, Data Catalogues, Dashboards, and IoT sensors are lagging behind others, such as Data representations, API, Visualizations, and Systems as a result of the growing awareness of data issues such as data heterogeneity and quality \citep{10.1007/978-3-030-87101-7_15}. 

In the following sub-sections, we provide comprehensive analytical insight and key results focusing on each domain \footnote{Although Building, Architecture, and Urban terms can be identified as different scales of the BE or the AEC rather than domains perse, they are segregated into different domains in this study because it is noted that there are some differences in the requirements and understanding of DT of each scale.}.


\subsubsection*{Building domain}
\begin{figure}[!h]
        \centering
        \includegraphics[width=\linewidth]{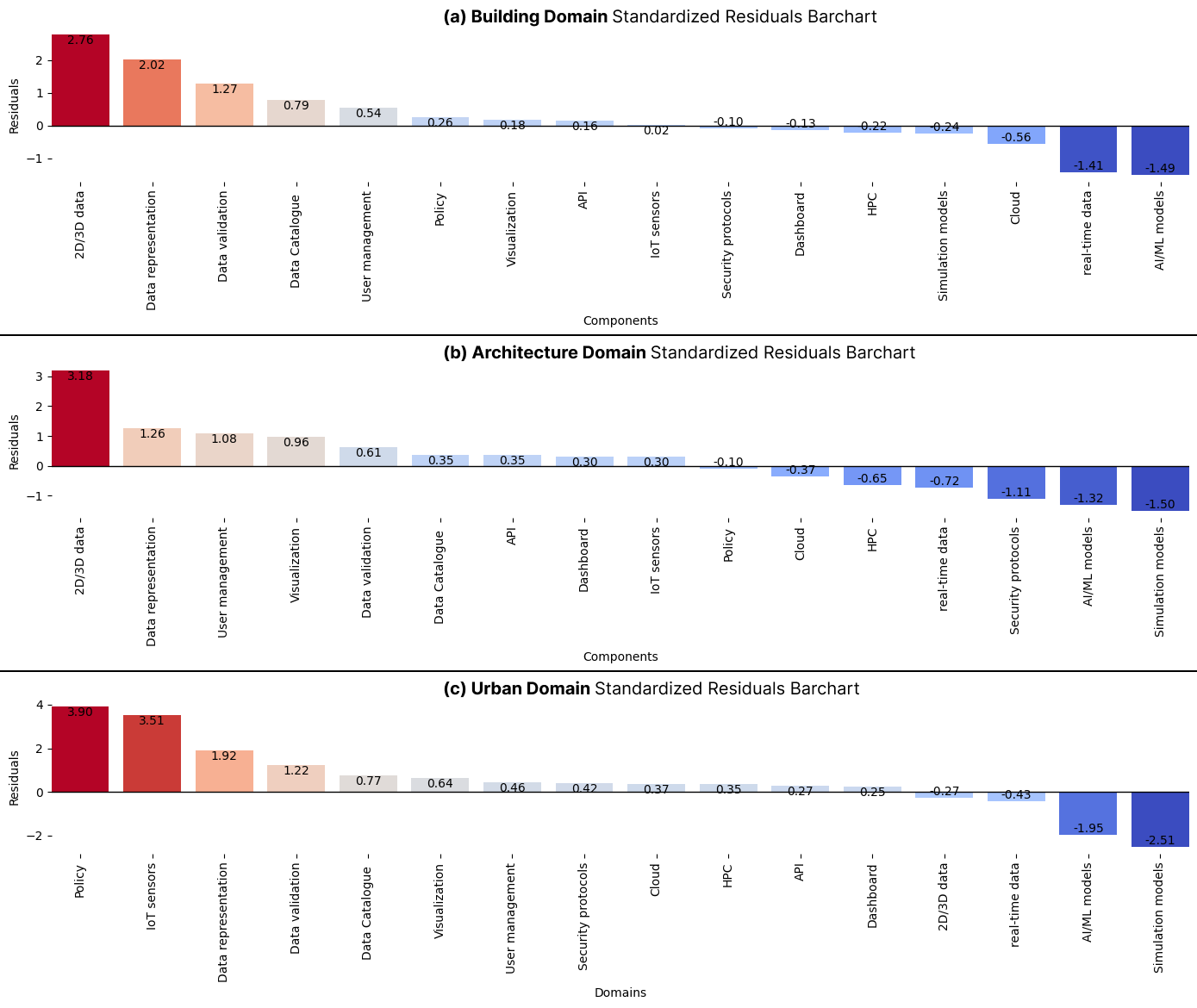}
        \caption{Components for the BE domains (a) Building, (b) Architecture, and (c) Urban sorted by significance. The manufacturing domain is excluded since its results is out of the scope of the research}
        \label{fig:significance_terms_sorted}
    \end{figure}

    Components like 2D/3D data (residual of 3.2), data representation (1.3), and visualization (0.96) are significantly more frequent than expected, underscoring their critical importance in accurately modeling and managing building structures in the form of Building Information Models (BIM) \citep{ABDELRAHMAN2022108532, ABDELRAHMAN2022109090, 10.1016/j.compag.2023.108441, 10.1016/j.seta.2021.101897} (Figure \ref{fig:significance_terms_sorted} - a).
  
    On the other hand, a negative residual indicates that simulation models are less frequent in the building domain than expected. This might suggest that while simulation models are important, they may not be as widely implemented in building digital twins as in other domains. Notably, real-time data, AI/ML, Security protocols, and HPC are not mature enough in the building digital twin domains in research and practice due to technical, organizational, and data-related aspects \citep{BILAL2016500, OSADCHA2023106704}. For instance, the integration of BIM and real-time data streams from IoT sensors is complex \citep{10.1016/j.autcon.2023.105109}, interoperability between different data schemas, sources, software and platforms is difficult, in addition to scalability issues \citep{doi:10.1061/(ASCE)ME.1943-5479.0000763, miller2021internet, ABDELRAHMAN2022109090}.

\subsubsection*{Architectural domain}
    Architecture terms show significance values closer to the building terms. Digital Twins are used in Architecture for various purposes, such as design and planning, construction management, and operation and maintenance. The architecture domain is characterized by the prominent use of 2D/3D data (Residual of 2.8), data representation (residual of 2), and validation (residual of 1.3) (Figure~\ref{fig:significance_terms_sorted} - b).

    The negative residual of simulation models indicates that simulation models are less frequent in the architecture domain of digital twins than expected. This finding might suggest that while simulation models are important, they may not be as widely implemented in architecture digital twins as in other domains. 

    AI/ML models (residual of -1.5) and Real-time data (residual of -1.4) are less significant in the architecture domain. This might be due to the lack of data and the complexity of the data in the architecture domain.

    Despite these challenges, the use of digital twins in architecture is growing. Architects are increasingly leveraging digital twins for improved project visualization, enhanced collaboration among stakeholders, better lifecycle management of buildings, facility management \citep{AGOSTINELLI2022149}, and construction monitoring \citep{10.1016/j.dibe.2023.100247, 10.1016/j.measurement.2023.112955}. For instance, digital twins can simulate the impact of design changes on building performance, enabling architects to make more informed decisions. Furthermore, the integration of augmented reality (AR) and virtual reality (VR) with digital twins is opening new avenues for immersive design experiences and client presentations. As technology continues to advance, it is expected that the adoption of AI/ML models and real-time data in architectural digital twins will increase, providing more sophisticated tools for predictive analytics and real-time monitoring of building systems.

\subsubsection*{Urban domain}
    Urban terms in digital twins (UDTs) reveal various insights and challenges. Our analysis shows that there are fewer significant components in UDT given the complexity of the domain and the requirements of massive computational and storage resources, among other challenges that hindered the adoption of some components, such as real-time data and AI/ML in the urban domain. In contrast, IoT sensors, data representation, data validation, and policy-driven decision-making show greater significance (Figure \ref{fig:significance_terms_sorted} - c).

    The dynamic and transient nature of cities requires computational resources and expertise across various domains. Acquiring big amounts of real-time data, primarily through sensors, is essential for enabling Urban Digital Twin (UDT). Despite the increasing number volume of data captured through IoT sensors (Residual of 3.5), many researchers argue that the sensor network in cities is still too sparse and heterogeneous \citep{KANDT2021102992}. This introduces significant challenges related to data acquisition, synchronization, and applications, which limited its use in UDTs. 

    The sparse and heterogeneous nature of IoT sensor networks affects AI/ML applications in urban digital twins. \textbf{AI/ML models (Residual of -1.9)} show less significance in the urban domain than expected, as these applications require massive computational power to handle big data from various sources. Several researchers argue that there is not enough data for proper AI/ML applications in urban digital twins. However, the trend (Figure \ref{fig:temporal_evolution_of_componetns}) indicates that the adoption of ML and AI in urban digital twins has been rapidly growing and gaining significant traction since 2020 \citep{SILVA2018697, LIM201886}. This growth is driven by advancements in computational technologies and the increasing availability of urban data from various resources, facilitating more sophisticated and scalable AI/ML applications in urban analytics and planning. A recent advancement in this field is GeoAI \citep{2022_jag_geoai,2024_epb_xai,2023_geoai_handbook_urban_sensing,gao2021geospatial}, which integrates AI/ML and data science with geoscience to enhance the capabilities of urban studies, including digital twins \citep{deren2021smart}.
    
    \textbf{Simulation models (Residual of -2.5)} indicate less frequent adoption of Simulation models in urban and city digital twin contexts. This is attributed to the dynamic and transient nature of cities, which requires adaptable models and hence substantial computational resources and expertise across various domains \citep{doi:10.1177/2399808318796416}. Many of these existing simulation models on the city scale are characterized by oversimplification \citep{WEIL2023104862}.

    Decision-making-enabled digital twins play a key role in the success of digital twin models, appearing significant in terms of \textbf{Policy within UDTs (Residual of 3.9)}. Urban policy typically focuses on the causal dynamics of data insights to address issues such as housing, transportation, public health, economic development, and social equity. Policies establish guidelines for the collection, management, and protection of big data and guide the strategic application of urban analytics to address long-term structural challenges facing cities \citep{KANDT2021102992}.

From the component analysis, we conclude that Building, Architecture, and Urban DTs can be categorized as long-term decision support rather than high-performance real-time systems. We acknowledge that there are several attempts to adopt real-time digital twin applications on city scales, however, there is no real and widely prominent HPRT scalable DT on city/urban scales to the best of the authors' knowledge. Thus, we categorize BDTs, ADTs, and UDTs as follows:  

\begin{itemize}
    \item \textbf{Building Digital Twins}: These digital twins emphasize accurate modeling, data representation, and visualization to support building management and operations. Key components include 2D/3D data, data representation, and visualization. Applications involve Building Information Models (BIM), building management systems, and lifecycle management of building structures. They help in accurate modeling, managing structures, and ensuring long-term operational efficiency.

       \item \textbf{Architectural Digital Twins}: Similar to building digital twins, these are focused on long-term planning and decision support, leveraging detailed data representation and validation. Key components include 2D/3D data, data representation, and validation. Applications include design and planning, construction management, operation, and maintenance. They are used to improve project visualization, enhance stakeholder collaboration, and manage building lifecycles.

    \item \textbf{Urban Digital Twins (UDTs)}: \textbf{LTDS} (Long Term Decision Support) Digital Twins. These digital twins focus on data representation, data validation, and policy-driven decision-making. Key components include IoT sensors, data representation, data validation, and policy. Applications involve urban planning, public health, transportation management, economic development, and social equity. These applications require long-term planning and policy formulation supported by data insights.

\end{itemize}

\subsection{Derived Definitions and Implications}
\label{subsec:definition_analysis}

\begin{figure}[!ht]
    \centering
    \includegraphics[width=\linewidth]{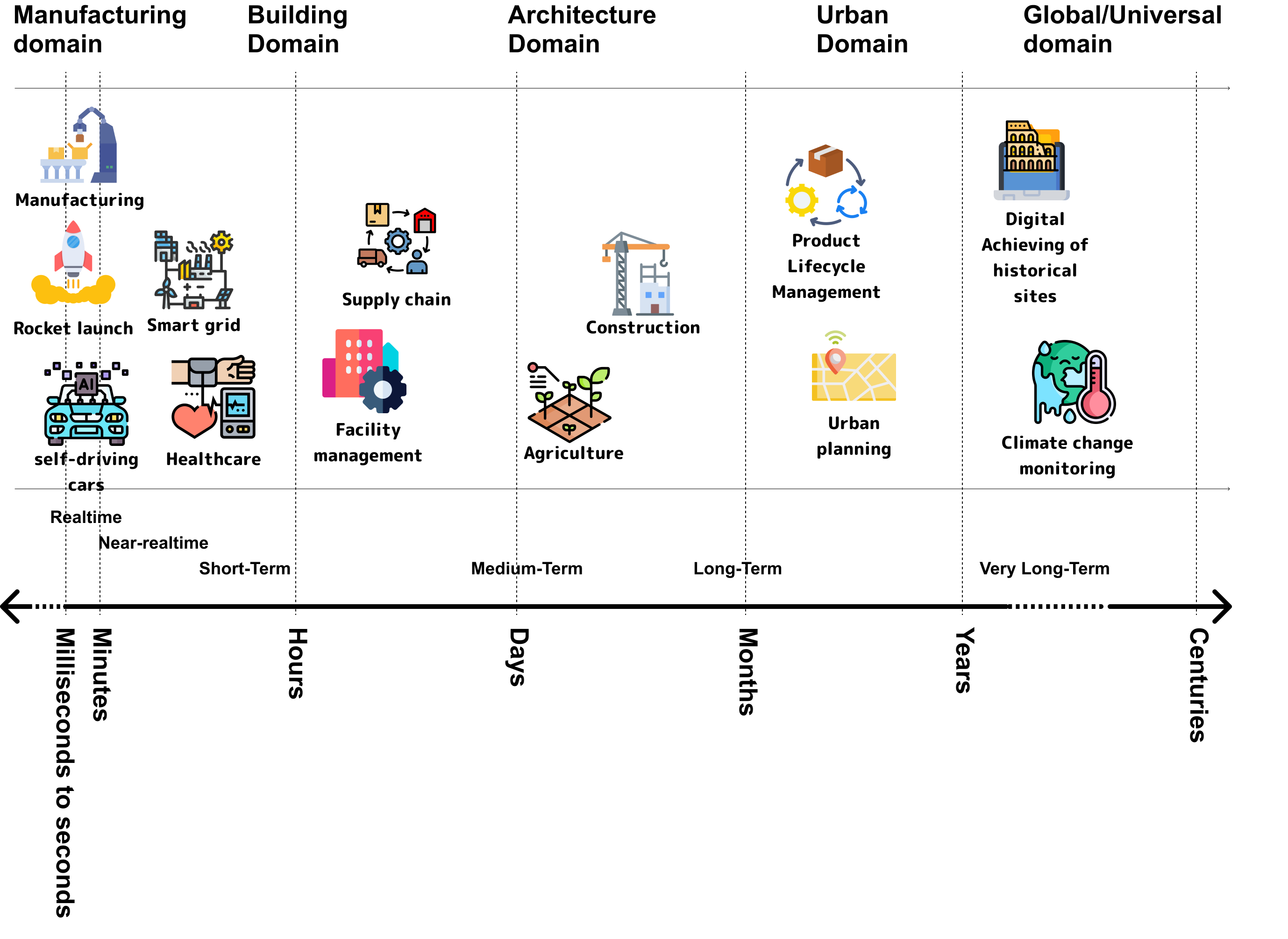}
    \caption{Applications of Digital Twins from real-time to very-long term temporal scales, suggesting that building, architecture, and urban DTs do not necessarily require real-time response as much as other domains such as manufacturing.}
    \label{fig:time_frames_dts}
\end{figure}

In this section, we show the results of the derived definitions based on the frequency analysis. The results reveal some divergences from the generalized definitions of digital twins. For example, when focusing on the urban/city domains, real-time data, and AI/ML models are not a significant portion of the definitions (Figure \ref{fig:time_frames_dts}). Also, given our analysis results, we identified minimum level of maturity requirements for each domain based on the CITYSTEPS framework by \cite{HARAGUCHI2024123409}.

The following subsections present the findings from the LLM analysis of the dataset, highlighting the top qualifying definitions that appeared in the literature for each domain and our derived definitions.

\begin{figure}[!h]
    \centering
    \includegraphics[width=\linewidth]{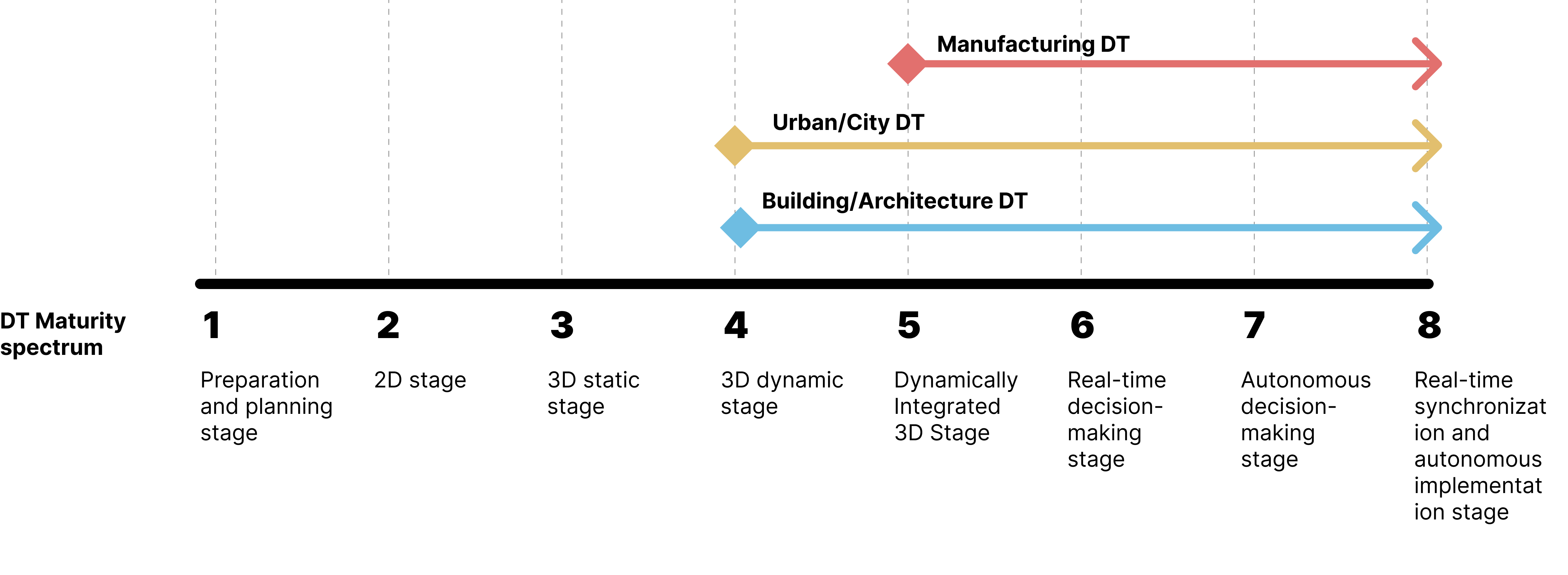}
    \caption{Approximate digital Twin maturity for each domain.}
    \label{fig:maturity}
\end{figure}
\subsubsection{Building and Architecture Digital Twin (BDT and ADT) definitions}

From the existing dataset, we extracted definitions that focus on comprehensive data representation, lifecycle management, and the ability to simulate and optimize operations without restrictions on data update frequencies. Such definitions are well-aligned with the significance test for building digital twins. These elements ensure that the digital twin can provide high-value services and support long-term asset management and operational efficiency: 

\textbf{ ``A digital twin is a virtual representation of a physical asset or system that securely holds all relevant static and dynamic information from concept to decommissioning, enabling high-value collaborative services''} by \cite{KOROTKOVA2023103787}.

This definition comprehensively emphasizes that a digital twin is a virtual representation that holds all relevant static and dynamic information, including 2D/3D data, data representation, data visualization, and data cataloging. Additionally, it mentions ``enabling high-value collaborative services'' which incorporates user management. The definition also does not restrict the frequency of data collection, processing, and action taking which aligns with the building DTs significance test. On the other hand, the definition focuses on the importance of the lifecycle of the asset/system, which spans over short to long-term frequencies and updates.

Another definition is \textbf{``Digital twin is a living model of the physical asset or system, which continually adapts to operational changes based on the collected online data and information and can forecast the future of the corresponding physical counterpart''} by \cite{MALEK20211047}. Although the definition mentions continual adaptation of changes, it does not restrict this adaptation to real-time. Furthermore, the reference to ``... can predict the future'' implies that ML/AI and simulation are considered ``good-to-have'' components rather than must-have components in a DT system.

A third definition that is also aligned with the significance test is the following: \textbf{``Digital twin is a virtual replica of a physical asset, such as a dam, that can be used to simulate and test various scenarios, monitor performance, and optimize operations''} as mentioned by \cite{HARIRIARDEBILI2023106813}. Although this definition gives an example of dams, it can also be applicable to building and construction. It doesn't have restrictions on the time-interval data updates between the physical and the virtual twins. Additionally, it highlights the capability to simulate and test various scenarios, which is crucial for optimizing performance and operations in building DTs. The flexibility in application and emphasis on optimization and monitoring are key aspects that resonate with the significance test.

A more generic building-oriented definition is: \textbf{``The digital twin is a digital representation of an asset or system and mimics its real-world behavior with the aim of managing, planning, predicting, and demonstrating current and future construction processes and models via synchronized representations''} by \cite{LEE2022101710}. The author also emphasized that ... ``The digital twin can capture overall changing site conditions in near real-time and creates a virtual simulation environment ...'' which indicates that the level of detail and time-frequency is subject to the nature of the asset, system, and the application of the digital twin. In this context, the authors refer to \textbf{``synchronized representations''}.

Another definition that leans more towards a generic digital twin but also aligns with the Building DT components: \textbf{``Digital Twin is defined as a software representation of a physical asset, system, or process designed to detect, prevent, predict, and optimize through analytics to deliver business value''} by \cite{LO2021101297}.

From the extracted definitions of the BDT, we have deduced the following definition: 

\begin{definition}
  \textbf{A Building/Architecture Digital Twin (BDT/ADT)} is a spatial-temporal virtual representation of a physical asset or system that manages all relevant static and dynamic information. It enables collaborative services, adapts to operational changes at suitable intervals, and can forecast future scenarios for optimized performance and decision-making.
\end{definition}

Gathering insights from the experts' responses, we highlighted ``accurate 3D geometry of buildings'' and ``rich semantic information'' as critical components in the BE DTs. Such findings align with the extract definitions from the developed corpus of research articles. According to the CITYSTEPS framework, for a BDT/ADT to be considered a true digital twin, it must reach at least maturity level 4 i.e., the 3D dynamic stage (Figure \ref{fig:maturity}). This dynamic aspect can range from real-time updates to long-term changes (Figure \ref{fig:time_frames_dts}). Therefore, a BIM model, whether hosted online or on local servers, that lacks regular updates reflecting changes in the real-world entity does not meet the criteria for a digital twin.

\subsubsection{Urban and City Digital Twin (UDTs and CDTs) definitions}
In the urban domain, there is no specific definition that accurately addresses all the components and their significance. However, some of these definitions are sufficiently generic to be applicable to urban/city/geospatial contexts. For example, one definition states \textbf{``Digital Twin is a dynamic representation of a real-life object that mirrors its states and behavior across its lifecycle and that can be used to monitor, analyze and simulate current and future states of and interventions on these objects, using data integration, artificial intelligence, and machine learning.''} \citep{verdouw2021digital, su2023digital}. This definition refers to mirroring the state and behavior of an object across its lifecycle. Although this could encompass that the mirror is in real-time. The authors mentioned that it is in near real-time. However, in the city and urban context, mirroring behavior does not necessarily need to be momentary; it can occur over various time spans (real-time, near real-time, short, mid, or long term intervals). Extensive data updates may lead to redundant and unnecessary information, computational and storage burdens, and, in many cases, may not inform decision-making effectively.

Another definition, the one by \cite{KONSTANTINIDOU20234026}, states that \textbf{``digital twin is a virtual model that replicates a real-world urban area/city, allowing the creator to determine the performance of its different systems''}. Although this definition is explicitly dedicated to the urban/city domain, it does not include information about the contribution of the DT in defining policies or informing decision making, which is a significant part of UDTs and CDTs. 

Evaluating the findings from our expert survey, we observe an alignment between research articles and experts' responses. For example, one expert defined an urban digital twin as ``a common and shareable (data) description of the current (physical) reality of the city''. However, several experts indicate critical components of digital twins in the survey which indeed contradict the analysis of systematic review. For example, one definition refers a digital twin to ``a general term for digital (virtual) replication of actual entities (object/processes), updated from real-time data, for supporting decision-making''. Further, some practitioners do not consider a 3D city model ``without automatic update procedures'' nor ``not reflecting real-time situation'' as a digital twin. Notably, the component analysis suggests that real-time data updates may not be necessary in the urban domain, given that urban development takes a relatively long time. Nevertheless, one expert addressed that `no comprehensive definition of the term urban digital twin has been provided yet'.

Based on the information extracted from the definitions, component analysis, and the expert survey, we derive the following definition for Urban and City Digital Twins (UDT and CDT). 

\begin{definition}
    \textbf{A City/Urban Digital Twin} is a spatial-temporal virtual representation of a real-world urban area or city, mirroring its states during its lifecycle through IoT sensors. It is used to monitor and analyze urban systems across different time spans to aid in decision-making and can be extended to simulate and predict various states and scenarios.
\end{definition}

In this definition, we emphasized that the scale of an UDT is a city or an urban area, and it must be a replica of an existing physical setting. IoT sensors is a key component in an urban digital twin due to the complexity and dynamics of cities. Data representation, monitoring, and analysis are a significant part of a CDT. Also, these data are used to make decisions in different time spans. On the other hand, a CDT can be extended to include real-time data integration, AI/ML, and simulation models. However, these components are not essential.  According to the CITYSTEPS framework (Figure \ref{fig:maturity}) for digital twin maturity, a CDT/UDT must reach at least level 4 (the 3D dynamic stage) to qualify as a digital twin. As with BDT/ADT, there is no strict requirement for the frequency of dynamic updates, which can range from real-time to long-term intervals or anything in between.

\section{Discussion and Future Directions}
\label{sec:discussion}

\subsection*{Comparison with traditional approaches}

Traditional approaches to analyzing definitions and concepts, such as manual literature reviews or expert-driven content analysis, often rely on smaller datasets due to the time-intensive nature of these methods. While these methods can provide nuanced insights, they are inherently limited in scalability and subject to potential bias introduced by individual reviewers.

Our approach leverages advanced NLP techniques, enabling us to process and analyze a large corpus of 15,000 articles efficiently. This methodology provides distinct advantages over traditional methods.

A key strength of this approach lies in its ability to derive definitions from a collective understanding of the concept. By analyzing a large and diverse set of papers discussing DTs across different disciplines, our methodology captures the broader agreement within the scientific community while identifying nuances and variations. This collective perspective enables the development of a definition that reflects a consensus-driven understanding, ensuring relevance and applicability across domains.

\subsection*{Current limitations in the BE DTs}
Our study has identified several limitations in the current interpretation and implementation of DTs within the BE. One significant limitation is the lack of real-time data integration and advanced AI/ML models despite being recognized by the majority of experts as essential. Our analysis and review showed that the adoption of such components is minimal. This gap is partly due to the complexity of integrating heterogeneous data sources and the high computational demands required for real-time processing on buildings and city scales. 

\subsection*{The future of ML/AI in BE DTs}
The integration of ML and AI in BE DTs is predicted to revolutionize the field by enabling predictive analytics, automated decision-making (Stage 7 on the CITYSTEPS model of maturity), and enhanced simulation capabilities. Future research should focus on developing scalable AI/ML models capable of handling the vast amounts of data generated in BE applications. This includes improving data quality and developing robust algorithms that can operate efficiently in real time. 

\subsection*{Multi-scale DTs (from human scale to the global scale)}
\textit{One size doesn't fit all} when it comes to a comprehensive definition of DTs. Although there are a few common denominator components of any DT, different scales (human, building, district, urban, city, global, and universal) DTs have different components. The organization and priority of these components differ significantly based on the use case. Future research should explore the development of multi-scale DT frameworks that seamlessly integrate data and insights across different scales, enabling a holistic approach to BE management.

\subsection*{Fully autonomous BE DTs}
By tracking the temporal trends and the evolution of DT components over the years, a clear trajectory toward fully autonomous DTs can be observed (Stage 8 of Maturity on the CITYSTEPS scale). The advancements in AI/ML, real-time data integration, and decision-making algorithms are key drivers in this direction. Autonomous DTs would possess the capability to continuously monitor and adapt to changes in the environment without human intervention, providing real-time responses to emerging issues and optimizing operations based on predictive insights.

Achieving full autonomy in BE DTs necessitates overcoming significant technical challenges, including the integration of advanced AI/ML models capable of handling vast and diverse datasets, ensuring robust data security and privacy, and developing seamless interoperability between different systems. 

\subsection*{Generative models for DTs}
Generative models hold great potential for enhancing the capabilities of DTs. These models can create realistic simulations of various scenarios, providing valuable insights for decision-making and planning. By leveraging generative models, DTs can simulate the impact of different design choices, environmental changes, and operational strategies. Future research should focus on integrating generative models with DT frameworks, enabling more accurate and comprehensive simulations.

\section{Conclusion}
\label{sec:conclusion}
Digital twin is now a common concept and technology in disciplines such as building science and urban planning. But its meaning is contested and often subject to specific interpretations, and there is a degree of gerrymandering as well.
In this research, we aspired to make sense of the terminology. We developed a novel method to do so, conducted an analysis of a massive set of descriptions of Digital Twin in scientific literature within the built environment using NLP, LLMs, and statistical approaches, capturing around one thousand definitions that appeared in the literature from 2000 to 2024. We extracted 15,000 full-text articles about digital twins and built on top of a pre-deployed Delphi expert survey~\citep{lei2023challenges} involving 52 experts in the related domains. Our goal was to identify the core components and characteristics of digital twins, focusing on the urban, architectural, and building domains. We depended on extracting DT components from a large corpus of text and then assessing these components against the standardized definitions and the existing definitions from other domains (e.g., Manufacturing). We conducted a frequency analysis and a Chi-Square statistical significance test to evaluate the prominence of various DT elements in specific domains.

The findings revealed significant variations in digital twin descriptions and paradigms across different domains, highlighting the complexity and evolving nature of this concept. We identified two primary categories of digital twins, Long Term Decision Support (LTDS) DTs and High-Performance Real Time (HPRT) DTs. We found that all the built-environment-related DTs fit on the first one.

Although this research addresses the current definition of Digital Twins (DTs), future trends indicate that some components are becoming increasingly prominent such as AI/ML, HPC, and Real-time data adoption. Thus, future research is needed to redefine DTs according to new trends, techniques, and tools. This reflects the fact that Building, Urban, and City DTs have not yet reached their state of equilibrium regarding a fully indicative definition due to their complexity and ever-changing landscapes.  This research offers a foundational framework for defining Digital Twins in the built environment and serves as a first step toward building consensus. The methodology and definitions presented here provide a sustainable approach that encourages researchers and practitioners to adopt these definitions in their implementations or refine them based on specific requirements. We encourage researchers to deploy the same methodology and revisit the definition periodically to track its evolution over time, reflecting advancements in technology and evolving understandings of the DT entity.

\section*{Declaration of Generative AI and AI-assisted technologies in the writing process}
During the preparation of this work, the authors used ChatGPT in order to format tables, proofread, and generate small elements in some of the illustrations (Figure \ref{fig:dt_components_illustration}). After using this tool/service, the authors reviewed and edited the content as needed and take full responsibility for the content of the published article.

\section*{Funding}
This research is part of the project Multi-scale Digital Twins for the Urban Environment: From Heartbeats to Cities, which is supported by the Singapore Ministry of Education Academic Research Fund Tier 1.
This research is part of the project Large-scale 3D Geospatial Data for Urban Analytics, which is supported by the National University of Singapore under the Start Up Grant R-295-000-171-133.
The research was partially conducted at the Future Cities Lab Global at the Singapore-ETH Centre, which was established collaboratively between ETH Z\"urich and the National Research Foundation Singapore (NRF) under its Campus for Research Excellence and Technological Enterprise (CREATE) programme. 
This research is part of the project From Models to Pavements: Advancing Urban Digital Twins with a Multi-Dimensional Human-Centric Approach for a Smart Walkability Analysis, which is supported by the Humanities Social Sciences Seed Fund at the National University of Singapore (A-8001957-00-00). 
EM's research work is supported through the Foreign PhD Scholarship Grant from the Department of Science and Technology - Engineering Research and Development for Technology (DOST-ERDT), Philippines.

\section*{Acknowledgements}
We thank the colleagues at the Urban Analytics Lab at the National University of Singapore (NUS) for the discussions. We extend our heartfelt thanks to the editor and the reviewers for their time, constructive feedback, and valuable suggestions, which enhanced the quality of the paper. We also sincerely thank YouPing Xie from the PwC China Data Management Office for providing feedback on the early version of the paper. Additional thanks go to the Delphi expert survey participants for their insights on Digital Twin definitions.


\begin{thebibliography}{155}
\expandafter\ifx\csname natexlab\endcsname\relax\def\natexlab#1{#1}\fi
\providecommand{\url}[1]{\texttt{#1}}
\providecommand{\href}[2]{#2}
\providecommand{\path}[1]{#1}
\providecommand{\DOIprefix}{doi:}
\providecommand{\ArXivprefix}{arXiv:}
\providecommand{\URLprefix}{URL: }
\providecommand{\Pubmedprefix}{pmid:}
\providecommand{\doi}[1]{\href{http://dx.doi.org/#1}{\path{#1}}}
\providecommand{\Pubmed}[1]{\href{pmid:#1}{\path{#1}}}
\providecommand{\bibinfo}[2]{#2}
\ifx\xfnm\relax \def\xfnm[#1]{\unskip,\space#1}\fi
\bibitem[{23247-1(2020)}]{iso2020automation}
\bibinfo{author}{23247-1, I.}, \bibinfo{year}{2020}.
\newblock \bibinfo{title}{Automation systems and integration—digital twin
  framework for manufacturing—part 1: overview and general principles}.
\bibitem[{Abdeen et~al.(2023)Abdeen, Shirowzhan and
  Sepasgozar}]{abdeen2023citizen}
\bibinfo{author}{Abdeen, F.N.}, \bibinfo{author}{Shirowzhan, S.},
  \bibinfo{author}{Sepasgozar, S.M.}, \bibinfo{year}{2023}.
\newblock \bibinfo{title}{Citizen-centric digital twin development with machine
  learning and interfaces for maintaining urban infrastructure}.
\newblock \bibinfo{journal}{Telematics and Informatics} ,
  \bibinfo{pages}{102032}.
\bibitem[{Abdelrahman et~al.(2022)Abdelrahman, Chong and
  Miller}]{ABDELRAHMAN2022108532}
\bibinfo{author}{Abdelrahman, M.M.}, \bibinfo{author}{Chong, A.},
  \bibinfo{author}{Miller, C.}, \bibinfo{year}{2022}.
\newblock \bibinfo{title}{Personal thermal comfort models using digital twins:
  Preference prediction with bim-extracted spatial–temporal proximity data
  from build2vec}.
\newblock \bibinfo{journal}{Building and Environment} \bibinfo{volume}{207},
  \bibinfo{pages}{108532}.
\newblock \URLprefix
  \url{https://www.sciencedirect.com/science/article/pii/S0360132321009240},
  \DOIprefix\doi{https://doi.org/10.1016/j.buildenv.2021.108532}.
\bibitem[{Abdelrahman and Miller(2022)}]{ABDELRAHMAN2022109090}
\bibinfo{author}{Abdelrahman, M.M.}, \bibinfo{author}{Miller, C.},
  \bibinfo{year}{2022}.
\newblock \bibinfo{title}{Targeting occupant feedback using digital twins:
  Adaptive spatial–temporal thermal preference sampling to optimize personal
  comfort models}.
\newblock \bibinfo{journal}{Building and Environment} \bibinfo{volume}{218},
  \bibinfo{pages}{109090}.
\newblock \URLprefix
  \url{https://www.sciencedirect.com/science/article/pii/S0360132322003274},
  \DOIprefix\doi{https://doi.org/10.1016/j.buildenv.2022.109090}.
\bibitem[{Adamenko et~al.(2020)Adamenko, Kunnen, Pluhnau, Loibl and
  Nagarajah}]{ADAMENKO202027}
\bibinfo{author}{Adamenko, D.}, \bibinfo{author}{Kunnen, S.},
  \bibinfo{author}{Pluhnau, R.}, \bibinfo{author}{Loibl, A.},
  \bibinfo{author}{Nagarajah, A.}, \bibinfo{year}{2020}.
\newblock \bibinfo{title}{Review and comparison of the methods of designing the
  digital twin}.
\newblock \bibinfo{journal}{Procedia CIRP} \bibinfo{volume}{91},
  \bibinfo{pages}{27--32}.
\newblock \URLprefix
  \url{https://www.sciencedirect.com/science/article/pii/S2212827120307800},
  \DOIprefix\doi{https://doi.org/10.1016/j.procir.2020.02.146}.
\bibitem[{Agnusdei et~al.(2021)Agnusdei, Elia and Gnoni}]{app11062767}
\bibinfo{author}{Agnusdei, G.P.}, \bibinfo{author}{Elia, V.},
  \bibinfo{author}{Gnoni, M.G.}, \bibinfo{year}{2021}.
\newblock \bibinfo{title}{Is digital twin technology supporting safety
  management? a bibliometric and systematic review}.
\newblock \bibinfo{journal}{Applied Sciences} \bibinfo{volume}{11}.
\newblock \URLprefix \url{https://www.mdpi.com/2076-3417/11/6/2767},
  \DOIprefix\doi{10.3390/app11062767}.
\bibitem[{Agostinelli and Heydari(2022)}]{AGOSTINELLI2022149}
\bibinfo{author}{Agostinelli, S.}, \bibinfo{author}{Heydari, A.},
  \bibinfo{year}{2022}.
\newblock \bibinfo{title}{Chapter six - digital twin predictive maintenance
  strategy based on machine learning improving facility management in built
  environment}, in: \bibinfo{editor}{Elsheikh, A.H.}, \bibinfo{editor}{{Abd
  Elaziz}, M.E.} (Eds.), \bibinfo{booktitle}{Artificial Neural Networks for
  Renewable Energy Systems and Real-World Applications}.
  \bibinfo{publisher}{Academic Press}, pp. \bibinfo{pages}{149--158}.
\newblock \URLprefix
  \url{https://www.sciencedirect.com/science/article/pii/B9780128207932000070},
  \DOIprefix\doi{https://doi.org/10.1016/B978-0-12-820793-2.00007-0}.
\bibitem[{Alva et~al.(2022)Alva, Biljecki and
  Stouffs}]{2022_3dgeoinfo_dt_use_cases}
\bibinfo{author}{Alva, P.}, \bibinfo{author}{Biljecki, F.},
  \bibinfo{author}{Stouffs, R.}, \bibinfo{year}{2022}.
\newblock \bibinfo{title}{Use cases for district-scale urban digital twins}.
\newblock \bibinfo{journal}{The International Archives of the Photogrammetry,
  Remote Sensing and Spatial Information Sciences}
  \bibinfo{volume}{XLVIII-4/W4-2022}, \bibinfo{pages}{5--12}.
\newblock \DOIprefix\doi{10.5194/isprs-archives-XLVIII-4-W4-2022-5-2022}.
\bibitem[{Bartos and Kerkez(2021)}]{bartos2021pipedream}
\bibinfo{author}{Bartos, M.}, \bibinfo{author}{Kerkez, B.},
  \bibinfo{year}{2021}.
\newblock \bibinfo{title}{Pipedream: An interactive digital twin model for
  natural and urban drainage systems}.
\newblock \bibinfo{journal}{Environmental Modelling \& Software}
  \bibinfo{volume}{144}, \bibinfo{pages}{105120}.
\bibitem[{Bartsch et~al.(2021)Bartsch, Pettke, H{\"u}bert, Lak{\"a}mper and
  Lange}]{Bartsch2021}
\bibinfo{author}{Bartsch, K.}, \bibinfo{author}{Pettke, A.},
  \bibinfo{author}{H{\"u}bert, A.}, \bibinfo{author}{Lak{\"a}mper, J.},
  \bibinfo{author}{Lange, F.}, \bibinfo{year}{2021}.
\newblock \bibinfo{title}{On the digital twin application and the role of
  artificial intelligence in additive manufacturing: a systematic review}.
\newblock \bibinfo{journal}{Journal of Physics: Materials} \bibinfo{volume}{4},
  \bibinfo{pages}{032005}.
\newblock \URLprefix \url{https://dx.doi.org/10.1088/2515-7639/abf3cf},
  \DOIprefix\doi{10.1088/2515-7639/abf3cf}.
\bibitem[{Batty(2018)}]{doi:10.1177/2399808318796416}
\bibinfo{author}{Batty, M.}, \bibinfo{year}{2018}.
\newblock \bibinfo{title}{Digital twins}.
\newblock \bibinfo{journal}{Environment and Planning B: Urban Analytics and
  City Science} \bibinfo{volume}{45}, \bibinfo{pages}{817--820}.
\newblock \DOIprefix\doi{10.1177/2399808318796416}.
\bibitem[{Belfadel et~al.(2023)Belfadel, Hörl, Tapia, Politaki, Kureshi,
  Tavasszy and Puchinger}]{Belfadel2023}
\bibinfo{author}{Belfadel, A.}, \bibinfo{author}{Hörl, S.},
  \bibinfo{author}{Tapia, R.J.}, \bibinfo{author}{Politaki, D.},
  \bibinfo{author}{Kureshi, I.}, \bibinfo{author}{Tavasszy, L.},
  \bibinfo{author}{Puchinger, J.}, \bibinfo{year}{2023}.
\newblock \bibinfo{title}{A conceptual digital twin framework for city
  logistics}.
\newblock \bibinfo{journal}{Computers, Environment and Urban Systems}
  \bibinfo{volume}{103}.
\newblock \URLprefix
  \url{https://www.sciencedirect.com/science/article/pii/S0198971523000522},
  \DOIprefix\doi{10.1016/j.compenvurbsys.2023.101989}.
\bibitem[{Bilal et~al.(2016)Bilal, Oyedele, Qadir, Munir, Ajayi, Akinade,
  Owolabi, Alaka and Pasha}]{BILAL2016500}
\bibinfo{author}{Bilal, M.}, \bibinfo{author}{Oyedele, L.O.},
  \bibinfo{author}{Qadir, J.}, \bibinfo{author}{Munir, K.},
  \bibinfo{author}{Ajayi, S.O.}, \bibinfo{author}{Akinade, O.O.},
  \bibinfo{author}{Owolabi, H.A.}, \bibinfo{author}{Alaka, H.A.},
  \bibinfo{author}{Pasha, M.}, \bibinfo{year}{2016}.
\newblock \bibinfo{title}{Big data in the construction industry: A review of
  present status, opportunities, and future trends}.
\newblock \bibinfo{journal}{Advanced Engineering Informatics}
  \bibinfo{volume}{30}, \bibinfo{pages}{500--521}.
\newblock \URLprefix
  \url{https://www.sciencedirect.com/science/article/pii/S1474034616301938},
  \DOIprefix\doi{https://doi.org/10.1016/j.aei.2016.07.001}.
\bibitem[{Biljecki(2023)}]{2023_geoai_handbook_urban_sensing}
\bibinfo{author}{Biljecki, F.}, \bibinfo{year}{2023}.
\newblock \bibinfo{title}{GeoAI for Urban Sensing}. \bibinfo{publisher}{CRC
  Press}. chapter~\bibinfo{chapter}{17}.
\newblock pp. \bibinfo{pages}{351--366}.
\newblock \DOIprefix\doi{10.1201/9781003308423-17}.
\bibitem[{Boje et~al.(2023)Boje, Álvaro José Hahn~Menacho, Marvuglia,
  Benetto, Kubicki, Schaubroeck and Gutiérrez}]{Boje2023}
\bibinfo{author}{Boje, C.}, \bibinfo{author}{Álvaro José Hahn~Menacho},
  \bibinfo{author}{Marvuglia, A.}, \bibinfo{author}{Benetto, E.},
  \bibinfo{author}{Kubicki, S.}, \bibinfo{author}{Schaubroeck, T.},
  \bibinfo{author}{Gutiérrez, T.N.}, \bibinfo{year}{2023}.
\newblock \bibinfo{title}{A framework using bim and digital twins in
  facilitating lcsa for buildings}.
\newblock \bibinfo{journal}{Journal of Building Engineering}
  \bibinfo{volume}{76}.
\newblock \URLprefix
  \url{https://www.sciencedirect.com/science/article/pii/S2352710223014122},
  \DOIprefix\doi{10.1016/j.jobe.2023.107232}.
\bibitem[{Bolton et~al.(2018)Bolton, Butler, Dabson, Enzer, Evans, Fenemore,
  Harradence, Keaney, Kemp, Luck et~al.}]{bolton2018gemini}
\bibinfo{author}{Bolton, A.}, \bibinfo{author}{Butler, L.},
  \bibinfo{author}{Dabson, I.}, \bibinfo{author}{Enzer, M.},
  \bibinfo{author}{Evans, M.}, \bibinfo{author}{Fenemore, T.},
  \bibinfo{author}{Harradence, F.}, \bibinfo{author}{Keaney, E.},
  \bibinfo{author}{Kemp, A.}, \bibinfo{author}{Luck, A.}, et~al.,
  \bibinfo{year}{2018}.
\newblock \bibinfo{title}{{Gemini principles}} \URLprefix
  \url{https://www.cdbb.cam.ac.uk/system/files/documents/TheGeminiPrinciples.pdf},
  \DOIprefix\doi{10.17863/CAM.32260}.
\bibitem[{Boschert and Rosen(2016)}]{Boschert2016}
\bibinfo{author}{Boschert, S.}, \bibinfo{author}{Rosen, R.},
  \bibinfo{year}{2016}.
\newblock \bibinfo{title}{Digital Twin---The Simulation Aspect}.
\newblock \bibinfo{publisher}{Springer International Publishing},
  \bibinfo{address}{Cham}.
\newblock \DOIprefix\doi{10.1007/978-3-319-32156-1\_5}.
\bibitem[{Callcut et~al.(2021)Callcut, Cerceau~Agliozzo, Varga and
  McMillan}]{callcut2021digital}
\bibinfo{author}{Callcut, M.}, \bibinfo{author}{Cerceau~Agliozzo, J.P.},
  \bibinfo{author}{Varga, L.}, \bibinfo{author}{McMillan, L.},
  \bibinfo{year}{2021}.
\newblock \bibinfo{title}{Digital twins in civil infrastructure systems}.
\newblock \bibinfo{journal}{Sustainability} \bibinfo{volume}{13},
  \bibinfo{pages}{11549}.
\bibitem[{Cavnar et~al.(1994)Cavnar, Trenkle et~al.}]{cavnar1994n}
\bibinfo{author}{Cavnar, W.B.}, \bibinfo{author}{Trenkle, J.M.}, et~al.,
  \bibinfo{year}{1994}.
\newblock \bibinfo{title}{N-gram-based text categorization}, in:
  \bibinfo{booktitle}{Proceedings of SDAIR-94, 3rd annual symposium on document
  analysis and information retrieval}, \bibinfo{organization}{Ann Arbor,
  Michigan}. p.~\bibinfo{pages}{14}.
\bibitem[{Chaturvedi and Kolbe(2016)}]{chaturvedi2016integrating}
\bibinfo{author}{Chaturvedi, K.}, \bibinfo{author}{Kolbe, T.H.},
  \bibinfo{year}{2016}.
\newblock \bibinfo{title}{{Integrating Dynamic Data and Sensors with Semantic
  3D City Models in the Context of Smart Cities}}.
\newblock \bibinfo{journal}{ISPRS Ann Photogramm Remote Sens Spatial Inf Sci}
  \bibinfo{volume}{4}, \bibinfo{pages}{31--38}.
\newblock \DOIprefix\doi{10.5194/isprs-annals-IV-2-W1-31-2016}.
\bibitem[{Chaudhuri et~al.(2003)Chaudhuri, Ganjam, Ganti and
  Motwani}]{chaudhuri2003robust}
\bibinfo{author}{Chaudhuri, S.}, \bibinfo{author}{Ganjam, K.},
  \bibinfo{author}{Ganti, V.}, \bibinfo{author}{Motwani, R.},
  \bibinfo{year}{2003}.
\newblock \bibinfo{title}{Robust and efficient fuzzy match for online data
  cleaning}, in: \bibinfo{booktitle}{Proceedings of the 2003 ACM SIGMOD
  international conference on Management of data}, pp.
  \bibinfo{pages}{313--324}.
\bibitem[{Cho et~al.(2023)Cho, Kim, Lim, Kim, Ji and
  Yeon}]{10.1016/j.compag.2023.108441}
\bibinfo{author}{Cho, J.}, \bibinfo{author}{Kim, C.}, \bibinfo{author}{Lim,
  K.J.}, \bibinfo{author}{Kim, J.}, \bibinfo{author}{Ji, B.},
  \bibinfo{author}{Yeon, J.}, \bibinfo{year}{2023}.
\newblock \bibinfo{title}{{Web-based agricultural infrastructure digital twin
  system integrated with GIS and BIM concepts}}.
\newblock \bibinfo{journal}{Computers and Electronics in Agriculture}
  \bibinfo{volume}{215}, \bibinfo{pages}{108441}.
\newblock \DOIprefix\doi{10.1016/j.compag.2023.108441}.
\bibitem[{Chomiak-Orsa et~al.(2023)Chomiak-Orsa, Hauke, Perechuda and
  Pondel}]{chomiak2023use}
\bibinfo{author}{Chomiak-Orsa, I.}, \bibinfo{author}{Hauke, K.},
  \bibinfo{author}{Perechuda, K.}, \bibinfo{author}{Pondel, M.},
  \bibinfo{year}{2023}.
\newblock \bibinfo{title}{The use of digital twin in the sustainable
  development of the city on the example of managing parking resources}.
\newblock \bibinfo{journal}{Procedia Computer Science} \bibinfo{volume}{225},
  \bibinfo{pages}{2183--2193}.
\bibitem[{Chong et~al.(2022)Chong, Hosamo and Hosamo}]{chong2022digital}
\bibinfo{author}{Chong, H.Y.}, \bibinfo{author}{Hosamo, H.H.},
  \bibinfo{author}{Hosamo, M.H.}, \bibinfo{year}{2022}.
\newblock \bibinfo{title}{Digital twin technology for bridge maintenance using
  3d laser scanning: A review}.
\newblock \bibinfo{journal}{Advances in Civil Engineering}
  \bibinfo{volume}{2022}, \bibinfo{pages}{2194949}.
\newblock \URLprefix \url{https://doi.org/10.1155/2022/2194949},
  \DOIprefix\doi{10.1155/2022/2194949}.
\bibitem[{Cimino et~al.(2019)Cimino, Negri and Fumagalli}]{CIMINO2019103130}
\bibinfo{author}{Cimino, C.}, \bibinfo{author}{Negri, E.},
  \bibinfo{author}{Fumagalli, L.}, \bibinfo{year}{2019}.
\newblock \bibinfo{title}{Review of digital twin applications in
  manufacturing}.
\newblock \bibinfo{journal}{Computers in Industry} \bibinfo{volume}{113},
  \bibinfo{pages}{103130}.
\newblock \DOIprefix\doi{https://doi.org/10.1016/j.compind.2019.103130}.
\bibitem[{Coupry et~al.(2021)Coupry, Noblecourt, Richard, Baudry and
  Bigaud}]{app11156810}
\bibinfo{author}{Coupry, C.}, \bibinfo{author}{Noblecourt, S.},
  \bibinfo{author}{Richard, P.}, \bibinfo{author}{Baudry, D.},
  \bibinfo{author}{Bigaud, D.}, \bibinfo{year}{2021}.
\newblock \bibinfo{title}{Bim-based digital twin and xr devices to improve
  maintenance procedures in smart buildings: A literature review}.
\newblock \bibinfo{journal}{Applied Sciences} \bibinfo{volume}{11}.
\newblock \URLprefix \url{https://www.mdpi.com/2076-3417/11/15/6810},
  \DOIprefix\doi{10.3390/app11156810}.
\bibitem[{Dalibor et~al.(2022)Dalibor, Jansen, Rumpe, Schmalzing, Wachtmeister,
  Wimmer and Wortmann}]{DALIBOR2022111361}
\bibinfo{author}{Dalibor, M.}, \bibinfo{author}{Jansen, N.},
  \bibinfo{author}{Rumpe, B.}, \bibinfo{author}{Schmalzing, D.},
  \bibinfo{author}{Wachtmeister, L.}, \bibinfo{author}{Wimmer, M.},
  \bibinfo{author}{Wortmann, A.}, \bibinfo{year}{2022}.
\newblock \bibinfo{title}{A cross-domain systematic mapping study on software
  engineering for digital twins}.
\newblock \bibinfo{journal}{Journal of Systems and Software}
  \bibinfo{volume}{193}, \bibinfo{pages}{111361}.
\newblock \URLprefix
  \url{https://www.sciencedirect.com/science/article/pii/S0164121222000917},
  \DOIprefix\doi{https://doi.org/10.1016/j.jss.2022.111361}.
\bibitem[{{Davila Delgado} and Oyedele(2021)}]{DAVILADELGADO2021101332}
\bibinfo{author}{{Davila Delgado}, J.M.}, \bibinfo{author}{Oyedele, L.},
  \bibinfo{year}{2021}.
\newblock \bibinfo{title}{Digital twins for the built environment: learning
  from conceptual and process models in manufacturing}.
\newblock \bibinfo{journal}{Advanced Engineering Informatics}
  \bibinfo{volume}{49}, \bibinfo{pages}{101332}.
\newblock \URLprefix
  \url{https://www.sciencedirect.com/science/article/pii/S1474034621000859},
  \DOIprefix\doi{https://doi.org/10.1016/j.aei.2021.101332}.
\bibitem[{Delgado and Oyedele(2021)}]{delgado2021digital}
\bibinfo{author}{Delgado, J.M.D.}, \bibinfo{author}{Oyedele, L.},
  \bibinfo{year}{2021}.
\newblock \bibinfo{title}{Digital twins for the built environment: learning
  from conceptual and process models in manufacturing}.
\newblock \bibinfo{journal}{Advanced Engineering Informatics}
  \bibinfo{volume}{49}, \bibinfo{pages}{101332}.
\bibitem[{Dembski et~al.(2020)Dembski, W{\"o}ssner, Letzgus, Ruddat and
  Yamu}]{dembski2020urban}
\bibinfo{author}{Dembski, F.}, \bibinfo{author}{W{\"o}ssner, U.},
  \bibinfo{author}{Letzgus, M.}, \bibinfo{author}{Ruddat, M.},
  \bibinfo{author}{Yamu, C.}, \bibinfo{year}{2020}.
\newblock \bibinfo{title}{{Urban digital twins for smart cities and citizens:
  The case study of Herrenberg, Germany}}.
\newblock \bibinfo{journal}{Sustainability} \bibinfo{volume}{12},
  \bibinfo{pages}{2307}.
\newblock \DOIprefix\doi{10.3390/su12062307}.
\bibitem[{Deng et~al.(2021)Deng, Zhang and Shen}]{deng2021systematic}
\bibinfo{author}{Deng, T.}, \bibinfo{author}{Zhang, K.}, \bibinfo{author}{Shen,
  Z.J.M.}, \bibinfo{year}{2021}.
\newblock \bibinfo{title}{A systematic review of a digital twin city: A new
  pattern of urban governance toward smart cities}.
\newblock \bibinfo{journal}{Journal of Management Science and Engineering}
  \bibinfo{volume}{6}, \bibinfo{pages}{125--134}.
\bibitem[{Depr{\^e}tre et~al.(2022)Depr{\^e}tre, Jacquinod and
  Mielniczek}]{depretre2022exploring}
\bibinfo{author}{Depr{\^e}tre, A.}, \bibinfo{author}{Jacquinod, F.},
  \bibinfo{author}{Mielniczek, A.}, \bibinfo{year}{2022}.
\newblock \bibinfo{title}{Exploring digital twin adaptation to the urban
  environment: comparison with cim to avoid silo-based approaches}.
\newblock \bibinfo{journal}{ISPRS Annals of the Photogrammetry, Remote Sensing
  and Spatial Information Sciences} \bibinfo{volume}{4},
  \bibinfo{pages}{337--344}.
\bibitem[{Deren et~al.(2021)Deren, Wenbo and Zhenfeng}]{deren2021smart}
\bibinfo{author}{Deren, L.}, \bibinfo{author}{Wenbo, Y.},
  \bibinfo{author}{Zhenfeng, S.}, \bibinfo{year}{2021}.
\newblock \bibinfo{title}{{Smart city based on digital twins}}.
\newblock \bibinfo{journal}{Computational Urban Science} \bibinfo{volume}{1},
  \bibinfo{pages}{1--11}.
\newblock \DOIprefix\doi{10.1007/s43762-021-00005-y}.
\bibitem[{{Digital Twin Consortium}(2020)}]{digitaltwin_con2020}
\bibinfo{author}{{Digital Twin Consortium}}, \bibinfo{year}{2020}.
\newblock \bibinfo{title}{Digital twin consortium defines digital twin}.
\newblock \URLprefix
  \url{https://www.digitaltwinconsortium.org/2020/12/digital-twin-consortium-defines-digital-twin/}.
  \bibinfo{note}{accessed: 2024-07-11}.
\bibitem[{{Digital Urban European Twins}(2020)}]{DUET_2020}
\bibinfo{author}{{Digital Urban European Twins}}, \bibinfo{year}{2020}.
\newblock \bibinfo{title}{{D5.1 System Architecture \& Implementation Plan}}.
\newblock \bibinfo{type}{Technical Report}. Digital Urban European Twins.
\newblock \URLprefix
  \url{https://www.digitalurbantwins.com/\_files/ugd/725ca8\_d088894bf9ee466b9934463840d79689.pdf}.
\bibitem[{Ehrlinger et~al.(2021)Ehrlinger, Schrott, Melichar, Kirchmayr and
  W{\"o}{\ss}}]{10.1007/978-3-030-87101-7_15}
\bibinfo{author}{Ehrlinger, L.}, \bibinfo{author}{Schrott, J.},
  \bibinfo{author}{Melichar, M.}, \bibinfo{author}{Kirchmayr, N.},
  \bibinfo{author}{W{\"o}{\ss}, W.}, \bibinfo{year}{2021}.
\newblock \bibinfo{title}{Data catalogs: A systematic literature review and
  guidelines to implementation}, in: \bibinfo{editor}{Kotsis, G.},
  \bibinfo{editor}{Tjoa, A.M.}, \bibinfo{editor}{Khalil, I.},
  \bibinfo{editor}{Moser, B.}, \bibinfo{editor}{Mashkoor, A.},
  \bibinfo{editor}{Sametinger, J.}, \bibinfo{editor}{Fensel, A.},
  \bibinfo{editor}{Martinez-Gil, J.}, \bibinfo{editor}{Fischer, L.},
  \bibinfo{editor}{Czech, G.}, \bibinfo{editor}{Sobieczky, F.},
  \bibinfo{editor}{Khan, S.} (Eds.), \bibinfo{booktitle}{Database and Expert
  Systems Applications - DEXA 2021 Workshops}, \bibinfo{publisher}{Springer
  International Publishing}, \bibinfo{address}{Cham}. pp.
  \bibinfo{pages}{148--158}.
\bibitem[{{El Saddik}(2023)}]{ELSADDIK20231}
\bibinfo{author}{{El Saddik}, A.}, \bibinfo{year}{2023}.
\newblock \bibinfo{title}{Chapter 1 - introduction}, in: \bibinfo{editor}{{El
  Saddik}, A.} (Ed.), \bibinfo{booktitle}{Digital Twin for Healthcare}.
  \bibinfo{publisher}{Academic Press}, pp. \bibinfo{pages}{1--13}.
\newblock \URLprefix
  \url{https://www.sciencedirect.com/science/article/pii/B9780323991636000068},
  \DOIprefix\doi{https://doi.org/10.1016/B978-0-32-399163-6.00006-8}.
\bibitem[{Ellul et~al.(2022)Ellul, Stoter and Bucher}]{ellul2022location}
\bibinfo{author}{Ellul, C.}, \bibinfo{author}{Stoter, J.},
  \bibinfo{author}{Bucher, B.}, \bibinfo{year}{2022}.
\newblock \bibinfo{title}{Location-enabled digital twins--understanding the
  role of nmcas in a european context}.
\newblock \bibinfo{journal}{ISPRS Annals of the Photogrammetry, Remote Sensing
  and Spatial Information Sciences} \bibinfo{volume}{10},
  \bibinfo{pages}{53--60}.
\bibitem[{Ellul et~al.(2024)Ellul, Stoter, Bucher, Olsson, Billen, DeLathouwer,
  Ellul, Stoter, Bucher, Olsson et~al.}]{ellul2024towards}
\bibinfo{author}{Ellul, C.}, \bibinfo{author}{Stoter, J.},
  \bibinfo{author}{Bucher, B.}, \bibinfo{author}{Olsson, P.O.},
  \bibinfo{author}{Billen, R.}, \bibinfo{author}{DeLathouwer, B.},
  \bibinfo{author}{Ellul, C.}, \bibinfo{author}{Stoter, J.},
  \bibinfo{author}{Bucher, B.}, \bibinfo{author}{Olsson, P.}, et~al.,
  \bibinfo{year}{2024}.
\newblock \bibinfo{title}{Towards national connected digital twins-a geospatial
  perspective}.
\newblock \bibinfo{journal}{ISPRS Annals of the Photogrammetry, Remote Sensing
  and Spatial Information Sciences} \bibinfo{volume}{10},
  \bibinfo{pages}{147--154}.
\bibitem[{Errandonea et~al.(2020)Errandonea, Beltr{\'a}n and
  Arrizabalaga}]{ERRANDONEA2020103316}
\bibinfo{author}{Errandonea, I.}, \bibinfo{author}{Beltr{\'a}n, S.},
  \bibinfo{author}{Arrizabalaga, S.}, \bibinfo{year}{2020}.
\newblock \bibinfo{title}{Digital twin for maintenance: A literature review}.
\newblock \bibinfo{journal}{Computers in Industry} \bibinfo{volume}{123},
  \bibinfo{pages}{103316}.
\newblock \URLprefix
  \url{https://www.sciencedirect.com/science/article/pii/S0166361520305509},
  \DOIprefix\doi{https://doi.org/10.1016/j.compind.2020.103316}.
\bibitem[{Evans et~al.(2019)Evans, Savian, Burns and Cooper}]{evans2019digital}
\bibinfo{author}{Evans, S.}, \bibinfo{author}{Savian, C.},
  \bibinfo{author}{Burns, A.}, \bibinfo{author}{Cooper, C.},
  \bibinfo{year}{2019}.
\newblock \bibinfo{title}{Digital twins for the built environment. an
  introduction to the opportunities, benefits, challenges and risks. white
  paper}.
\newblock \bibinfo{journal}{Built Environment Institution of Engineering and
  Technology} .
\bibitem[{{Farhan Hussain} et~al.(2023){Farhan Hussain}, Mokhtari, Ghalambor
  and {Amini Salehi}}]{FARHANHUSSAIN20231}
\bibinfo{author}{{Farhan Hussain}, R.}, \bibinfo{author}{Mokhtari, A.},
  \bibinfo{author}{Ghalambor, A.}, \bibinfo{author}{{Amini Salehi}, M.},
  \bibinfo{year}{2023}.
\newblock \bibinfo{title}{Chapter 1 - introduction to smart o\&g industry:
  Overview of smart o\&g industry}, in: \bibinfo{editor}{{Farhan Hussain}, R.},
  \bibinfo{editor}{Mokhtari, A.}, \bibinfo{editor}{Ghalambor, A.},
  \bibinfo{editor}{{Amini Salehi}, M.} (Eds.), \bibinfo{booktitle}{IoT for
  Smart Operations in the Oil and Gas Industry}. \bibinfo{publisher}{Gulf
  Professional Publishing}, pp. \bibinfo{pages}{1--17}.
\newblock \URLprefix
  \url{https://www.sciencedirect.com/science/article/pii/B9780323911511000101},
  \DOIprefix\doi{https://doi.org/10.1016/B978-0-32-391151-1.00010-1}.
\bibitem[{Gantayat et~al.(2014)Gantayat, Misra and Panda}]{gantayat2014study}
\bibinfo{author}{Gantayat, S.S.}, \bibinfo{author}{Misra, A.},
  \bibinfo{author}{Panda, B.}, \bibinfo{year}{2014}.
\newblock \bibinfo{title}{A study of incomplete data--a review}, in:
  \bibinfo{booktitle}{Proceedings of the International Conference on Frontiers
  of Intelligent Computing: Theory and Applications (FICTA) 2013},
  \bibinfo{organization}{Springer}. pp. \bibinfo{pages}{401--408}.
\bibitem[{Gao(2021)}]{gao2021geospatial}
\bibinfo{author}{Gao, S.}, \bibinfo{year}{2021}.
\newblock \bibinfo{title}{Geospatial artificial intelligence (GeoAI)}.
  volume~\bibinfo{volume}{10}.
\newblock \bibinfo{publisher}{Oxford University Press New York}.
\bibitem[{Gelernter(1993)}]{gelernter1993mirror}
\bibinfo{author}{Gelernter, D.}, \bibinfo{year}{1993}.
\newblock \bibinfo{title}{Mirror worlds: Or the day software puts the universe
  in a shoebox... How it will happen and what it will mean}.
\newblock \bibinfo{publisher}{Oxford University Press}.
\bibitem[{Ghenai et~al.(2022)Ghenai, Husein, {Al Nahlawi}, Hamid and
  Bettayeb}]{GHENAI2022102837}
\bibinfo{author}{Ghenai, C.}, \bibinfo{author}{Husein, L.A.},
  \bibinfo{author}{{Al Nahlawi}, M.}, \bibinfo{author}{Hamid, A.K.},
  \bibinfo{author}{Bettayeb, M.}, \bibinfo{year}{2022}.
\newblock \bibinfo{title}{Recent trends of digital twin technologies in the
  energy sector: A comprehensive review}.
\newblock \bibinfo{journal}{Sustainable Energy Technologies and Assessments}
  \bibinfo{volume}{54}, \bibinfo{pages}{102837}.
\newblock \URLprefix
  \url{https://www.sciencedirect.com/science/article/pii/S2213138822008852},
  \DOIprefix\doi{https://doi.org/10.1016/j.seta.2022.102837}.
\bibitem[{Glaessgen and Stargel(2012)}]{glaessgen2012digital}
\bibinfo{author}{Glaessgen, E.}, \bibinfo{author}{Stargel, D.},
  \bibinfo{year}{2012}.
\newblock \bibinfo{title}{The digital twin paradigm for future nasa and us air
  force vehicles}, in: \bibinfo{booktitle}{53rd AIAA/ASME/ASCE/AHS/ASC
  structures, structural dynamics and materials conference 20th AIAA/ASME/AHS
  adaptive structures conference 14th AIAA}, p. \bibinfo{pages}{1818}.
\bibitem[{Gohari et~al.(2019)Gohari, Berry and Barari}]{gohari2019digital}
\bibinfo{author}{Gohari, H.}, \bibinfo{author}{Berry, C.},
  \bibinfo{author}{Barari, A.}, \bibinfo{year}{2019}.
\newblock \bibinfo{title}{A digital twin for integrated inspection system in
  digital manufacturing}.
\newblock \bibinfo{journal}{IFAC-PapersOnLine} \bibinfo{volume}{52},
  \bibinfo{pages}{182--187}.
\bibitem[{Grieves(2014)}]{grieves2014digital}
\bibinfo{author}{Grieves, M.}, \bibinfo{year}{2014}.
\newblock \bibinfo{title}{Digital twin: manufacturing excellence through
  virtual factory replication}.
\newblock \bibinfo{journal}{White paper} \bibinfo{volume}{1},
  \bibinfo{pages}{1--7}.
\bibitem[{Grieves and Vickers(2017)}]{Grieves2017}
\bibinfo{author}{Grieves, M.}, \bibinfo{author}{Vickers, J.},
  \bibinfo{year}{2017}.
\newblock \bibinfo{title}{Digital twin: Mitigating unpredictable, undesirable
  emergent behavior in complex systems}, in:
  \bibinfo{booktitle}{Transdisciplinary Perspectives on Complex Systems: New
  Findings and Approaches}, \bibinfo{publisher}{Springer International
  Publishing}, \bibinfo{address}{Cham}. pp. \bibinfo{pages}{85--113}.
\bibitem[{Grieves(2005)}]{grieves2005product}
\bibinfo{author}{Grieves, M.W.}, \bibinfo{year}{2005}.
\newblock \bibinfo{title}{Product lifecycle management: the new paradigm for
  enterprises}.
\newblock \bibinfo{journal}{International Journal of Product Development}
  \bibinfo{volume}{2}, \bibinfo{pages}{71--84}.
\bibitem[{Haag and Anderl(2018)}]{HAAG}
\bibinfo{author}{Haag, S.}, \bibinfo{author}{Anderl, R.}, \bibinfo{year}{2018}.
\newblock \bibinfo{title}{Digital twin – proof of concept}.
\newblock \bibinfo{journal}{Manufacturing Letters} \bibinfo{volume}{15},
  \bibinfo{pages}{64--66}.
\newblock \URLprefix
  \url{https://www.sciencedirect.com/science/article/pii/S2213846318300208},
  \DOIprefix\doi{https://doi.org/10.1016/j.mfglet.2018.02.006}.
  \bibinfo{note}{industry 4.0 and Smart Manufacturing}.
\bibitem[{Hananto et~al.(2024)Hananto, Tirta, Herawan, Idris, Soudagar, Djamari
  and Veza}]{hananto2024digital}
\bibinfo{author}{Hananto, A.L.}, \bibinfo{author}{Tirta, A.},
  \bibinfo{author}{Herawan, S.G.}, \bibinfo{author}{Idris, M.},
  \bibinfo{author}{Soudagar, M.E.M.}, \bibinfo{author}{Djamari, D.W.},
  \bibinfo{author}{Veza, I.}, \bibinfo{year}{2024}.
\newblock \bibinfo{title}{Digital twin and 3d digital twin: Concepts,
  applications, and challenges in industry 4.0 for digital twin}.
\newblock \bibinfo{journal}{Computers} \bibinfo{volume}{13},
  \bibinfo{pages}{100}.
\bibitem[{Haraguchi et~al.(2024)Haraguchi, Funahashi and
  Biljecki}]{HARAGUCHI2024123409}
\bibinfo{author}{Haraguchi, M.}, \bibinfo{author}{Funahashi, T.},
  \bibinfo{author}{Biljecki, F.}, \bibinfo{year}{2024}.
\newblock \bibinfo{title}{Assessing governance implications of city digital
  twin technology: A maturity model approach}.
\newblock \bibinfo{journal}{Technological Forecasting and Social Change}
  \bibinfo{volume}{204}, \bibinfo{pages}{123409}.
\newblock \URLprefix
  \url{https://www.sciencedirect.com/science/article/pii/S0040162524002051},
  \DOIprefix\doi{https://doi.org/10.1016/j.techfore.2024.123409}.
\bibitem[{Hariri-Ardebili et~al.(2023)Hariri-Ardebili, Mahdavi, Nuss and
  Lall}]{HARIRIARDEBILI2023106813}
\bibinfo{author}{Hariri-Ardebili, M.A.}, \bibinfo{author}{Mahdavi, G.},
  \bibinfo{author}{Nuss, L.K.}, \bibinfo{author}{Lall, U.},
  \bibinfo{year}{2023}.
\newblock \bibinfo{title}{The role of artificial intelligence and digital
  technologies in dam engineering: Narrative review and outlook}.
\newblock \bibinfo{journal}{Engineering Applications of Artificial
  Intelligence} \bibinfo{volume}{126}, \bibinfo{pages}{106813}.
\newblock \URLprefix
  \url{https://www.sciencedirect.com/science/article/pii/S0952197623009971},
  \DOIprefix\doi{https://doi.org/10.1016/j.engappai.2023.106813}.
\bibitem[{Hassani et~al.(2022)Hassani, Huang and MacFeely}]{Hassani2022}
\bibinfo{author}{Hassani, H.}, \bibinfo{author}{Huang, X.},
  \bibinfo{author}{MacFeely, S.}, \bibinfo{year}{2022}.
\newblock \bibinfo{title}{Enabling digital twins to support the un sdgs}.
\newblock \bibinfo{journal}{Big Data and Cognitive Computing}
  \bibinfo{volume}{6}, \bibinfo{pages}{115}.
\newblock \URLprefix \url{http://dx.doi.org/10.3390/bdcc6040115},
  \DOIprefix\doi{10.3390/bdcc6040115}.
\bibitem[{He and Bai(2021)}]{he2021digital}
\bibinfo{author}{He, B.}, \bibinfo{author}{Bai, K.J.}, \bibinfo{year}{2021}.
\newblock \bibinfo{title}{Digital twin-based sustainable intelligent
  manufacturing: A review}.
\newblock \bibinfo{journal}{Advances in Manufacturing} \bibinfo{volume}{9},
  \bibinfo{pages}{1--21}.
\bibitem[{Homa~Masoumi and Pettit(2023)}]{doi:10.1080/20964471.2022.2160156}
\bibinfo{author}{Homa~Masoumi, Sara~Shirowzhan, P.E.}, \bibinfo{author}{Pettit,
  C.J.}, \bibinfo{year}{2023}.
\newblock \bibinfo{title}{City digital twins: their maturity level and
  differentiation from 3d city models}.
\newblock \bibinfo{journal}{Big Earth Data} \bibinfo{volume}{7},
  \bibinfo{pages}{1--36}.
\newblock \URLprefix \url{https://doi.org/10.1080/20964471.2022.2160156},
  \DOIprefix\doi{10.1080/20964471.2022.2160156},
  \href{http://arxiv.org/abs/https://doi.org/10.1080/20964471.2022.2160156}{{\tt
  arXiv:https://doi.org/10.1080/20964471.2022.2160156}}.
\bibitem[{Hribernik et~al.(2013)Hribernik, Wuest and
  Thoben}]{hribernik2013towards}
\bibinfo{author}{Hribernik, K.}, \bibinfo{author}{Wuest, T.},
  \bibinfo{author}{Thoben, K.D.}, \bibinfo{year}{2013}.
\newblock \bibinfo{title}{Towards product avatars representing middle-of-life
  information for improving design, development and manufacturing processes},
  in: \bibinfo{booktitle}{IFIP International Conference on Digital Product and
  Process Development Systems}, \bibinfo{organization}{Springer}. pp.
  \bibinfo{pages}{85--96}.
\bibitem[{{International Organization for
  Standardization}(2023)}]{ISO30173_2023}
\bibinfo{author}{{International Organization for Standardization}},
  \bibinfo{year}{2023}.
\newblock \bibinfo{title}{{ISO/IEC 30173:2023} Digital twin - Concepts and
  terminology}.
\newblock \bibinfo{type}{Technical Report}. International Organization for
  Standardization. \bibinfo{address}{Geneva, Switzerland}.
\newblock \URLprefix \url{hhttps://www.iso.org/standard/81442.html}.
\bibitem[{Jeddoub et~al.(2023)Jeddoub, Nys, Hajji and Billen}]{Jeddoub_2023}
\bibinfo{author}{Jeddoub, I.}, \bibinfo{author}{Nys, G.A.},
  \bibinfo{author}{Hajji, R.}, \bibinfo{author}{Billen, R.},
  \bibinfo{year}{2023}.
\newblock \bibinfo{title}{Digital twins for cities: Analyzing the gap between
  concepts and current implementations with a specific focus on data
  integration}.
\newblock \bibinfo{journal}{International Journal of Applied Earth Observation
  and Geoinformation} \bibinfo{volume}{122}, \bibinfo{pages}{103440}.
\newblock \URLprefix \url{http://dx.doi.org/10.1016/j.jag.2023.103440},
  \DOIprefix\doi{10.1016/j.jag.2023.103440}.
\bibitem[{Jiang et~al.(2021)Jiang, Yin, Li, Luo and
  Kaynak}]{jiang2021industrial}
\bibinfo{author}{Jiang, Y.}, \bibinfo{author}{Yin, S.}, \bibinfo{author}{Li,
  K.}, \bibinfo{author}{Luo, H.}, \bibinfo{author}{Kaynak, O.},
  \bibinfo{year}{2021}.
\newblock \bibinfo{title}{{Industrial applications of digital twins}}.
\newblock \bibinfo{journal}{Philosophical Transactions of the Royal Society A}
  \bibinfo{volume}{379}, \bibinfo{pages}{20200360}.
\newblock \DOIprefix\doi{10.1098/rsta.2020.0360}.
\bibitem[{Jones et~al.(2020)Jones, Snider, Nassehi, Yon and
  Hicks}]{JONES202036}
\bibinfo{author}{Jones, D.}, \bibinfo{author}{Snider, C.},
  \bibinfo{author}{Nassehi, A.}, \bibinfo{author}{Yon, J.},
  \bibinfo{author}{Hicks, B.}, \bibinfo{year}{2020}.
\newblock \bibinfo{title}{Characterising the digital twin: A systematic
  literature review}.
\newblock \bibinfo{journal}{CIRP Journal of Manufacturing Science and
  Technology} \bibinfo{volume}{29}, \bibinfo{pages}{36--52}.
\newblock \URLprefix
  \url{https://www.sciencedirect.com/science/article/pii/S1755581720300110},
  \DOIprefix\doi{https://doi.org/10.1016/j.cirpj.2020.02.002}.
\bibitem[{Juarez et~al.(2021)Juarez, Botti and Giret}]{juarez2021digital}
\bibinfo{author}{Juarez, M.G.}, \bibinfo{author}{Botti, V.J.},
  \bibinfo{author}{Giret, A.S.}, \bibinfo{year}{2021}.
\newblock \bibinfo{title}{{Digital Twins: Review and Challenges}}.
\newblock \bibinfo{journal}{Journal of Computing and Information Science in
  Engineering} \bibinfo{volume}{21}.
\newblock \DOIprefix\doi{10.1115/1.4050244}.
\bibitem[{Kamel~Boulos and Zhang(2021)}]{kamel2021digital}
\bibinfo{author}{Kamel~Boulos, M.N.}, \bibinfo{author}{Zhang, P.},
  \bibinfo{year}{2021}.
\newblock \bibinfo{title}{Digital twins: from personalised medicine to
  precision public health}.
\newblock \bibinfo{journal}{Journal of personalized medicine}
  \bibinfo{volume}{11}, \bibinfo{pages}{745}.
\bibitem[{Kandt and Batty(2021)}]{KANDT2021102992}
\bibinfo{author}{Kandt, J.}, \bibinfo{author}{Batty, M.}, \bibinfo{year}{2021}.
\newblock \bibinfo{title}{Smart cities, big data and urban policy: Towards
  urban analytics for the long run}.
\newblock \bibinfo{journal}{Cities} \bibinfo{volume}{109},
  \bibinfo{pages}{102992}.
\newblock \URLprefix
  \url{https://www.sciencedirect.com/science/article/pii/S0264275120313408},
  \DOIprefix\doi{https://doi.org/10.1016/j.cities.2020.102992}.
\bibitem[{Kim and Bartos(2024)}]{kim2024digital}
\bibinfo{author}{Kim, M.G.}, \bibinfo{author}{Bartos, M.},
  \bibinfo{year}{2024}.
\newblock \bibinfo{title}{A digital twin model for contaminant fate and
  transport in urban and natural drainage networks with online state
  estimation}.
\newblock \bibinfo{journal}{Environmental Modelling \& Software}
  \bibinfo{volume}{171}, \bibinfo{pages}{105868}.
\bibitem[{Kirchherr et~al.(2017)Kirchherr, Reike and Hekkert}]{Kirchherr_2017}
\bibinfo{author}{Kirchherr, J.}, \bibinfo{author}{Reike, D.},
  \bibinfo{author}{Hekkert, M.}, \bibinfo{year}{2017}.
\newblock \bibinfo{title}{Conceptualizing the circular economy: An analysis of
  114 definitions}.
\newblock \bibinfo{journal}{Resources, Conservation and Recycling}
  \bibinfo{volume}{127}, \bibinfo{pages}{221--232}.
\newblock \URLprefix \url{http://dx.doi.org/10.1016/j.resconrec.2017.09.005},
  \DOIprefix\doi{10.1016/j.resconrec.2017.09.005}.
\bibitem[{Klar et~al.(2024)Klar, Arvidsson and Angelakis}]{10367836}
\bibinfo{author}{Klar, R.}, \bibinfo{author}{Arvidsson, N.},
  \bibinfo{author}{Angelakis, V.}, \bibinfo{year}{2024}.
\newblock \bibinfo{title}{Digital twins' maturity: The need for
  interoperability}.
\newblock \bibinfo{journal}{IEEE Systems Journal} \bibinfo{volume}{18},
  \bibinfo{pages}{713--724}.
\newblock \DOIprefix\doi{10.1109/JSYST.2023.3340422}.
\bibitem[{Konstantinidou et~al.(2023)Konstantinidou, Grau, Sahala, Nesterova
  and {de Kruijf}}]{KONSTANTINIDOU20234026}
\bibinfo{author}{Konstantinidou, M.}, \bibinfo{author}{Grau, J.M.S.},
  \bibinfo{author}{Sahala, S.}, \bibinfo{author}{Nesterova, N.},
  \bibinfo{author}{{de Kruijf}, J.}, \bibinfo{year}{2023}.
\newblock \bibinfo{title}{A methodology for using dynamic visualizations to
  enhance citizens engagement}.
\newblock \bibinfo{journal}{Transportation Research Procedia}
  \bibinfo{volume}{72}, \bibinfo{pages}{4026--4032}.
\newblock \URLprefix
  \url{https://www.sciencedirect.com/science/article/pii/S2352146523006737},
  \DOIprefix\doi{https://doi.org/10.1016/j.trpro.2023.11.376}.
  \bibinfo{note}{tRA Lisbon 2022 Conference Proceedings Transport Research
  Arena (TRA Lisbon 2022),14th-17th November 2022, Lisboa, Portugal}.
\bibitem[{Korotkova et~al.(2023)Korotkova, Benders, Mikalef and
  Cameron}]{KOROTKOVA2023103787}
\bibinfo{author}{Korotkova, N.}, \bibinfo{author}{Benders, J.},
  \bibinfo{author}{Mikalef, P.}, \bibinfo{author}{Cameron, D.},
  \bibinfo{year}{2023}.
\newblock \bibinfo{title}{Maneuvering between skepticism and optimism about
  hyped technologies: Building trust in digital twins}.
\newblock \bibinfo{journal}{Information \& Management} \bibinfo{volume}{60},
  \bibinfo{pages}{103787}.
\newblock \URLprefix
  \url{https://www.sciencedirect.com/science/article/pii/S0378720623000356},
  \DOIprefix\doi{https://doi.org/10.1016/j.im.2023.103787}.
\bibitem[{Kosacka-Olejnik et~al.(2021)Kosacka-Olejnik, Kostrzewski, Marczewska,
  Mr{\'o}wczy{\'n}ska and Pawlewski}]{en14164919}
\bibinfo{author}{Kosacka-Olejnik, M.}, \bibinfo{author}{Kostrzewski, M.},
  \bibinfo{author}{Marczewska, M.}, \bibinfo{author}{Mr{\'o}wczy{\'n}ska, B.},
  \bibinfo{author}{Pawlewski, P.}, \bibinfo{year}{2021}.
\newblock \bibinfo{title}{How digital twin concept supports internal transport
  systems?---literature review}.
\newblock \bibinfo{journal}{Energies} \bibinfo{volume}{14}.
\newblock \URLprefix \url{https://www.mdpi.com/1996-1073/14/16/4919},
  \DOIprefix\doi{10.3390/en14164919}.
\bibitem[{Kritzinger et~al.(2018)Kritzinger, Karner, Traar, Henjes and
  Sihn}]{KRITZINGER20181016}
\bibinfo{author}{Kritzinger, W.}, \bibinfo{author}{Karner, M.},
  \bibinfo{author}{Traar, G.}, \bibinfo{author}{Henjes, J.},
  \bibinfo{author}{Sihn, W.}, \bibinfo{year}{2018}.
\newblock \bibinfo{title}{Digital twin in manufacturing: A categorical
  literature review and classification}.
\newblock \bibinfo{journal}{IFAC-PapersOnLine} \bibinfo{volume}{51},
  \bibinfo{pages}{1016--1022}.
\newblock \URLprefix
  \url{https://www.sciencedirect.com/science/article/pii/S2405896318316021},
  \DOIprefix\doi{https://doi.org/10.1016/j.ifacol.2018.08.474}.
  \bibinfo{note}{16th IFAC Symposium on Information Control Problems in
  Manufacturing INCOM 2018}.
\bibitem[{Le and Mikolov(2014)}]{le2014distributed}
\bibinfo{author}{Le, Q.}, \bibinfo{author}{Mikolov, T.}, \bibinfo{year}{2014}.
\newblock \bibinfo{title}{Distributed representations of sentences and
  documents}, in: \bibinfo{booktitle}{International conference on machine
  learning}, \bibinfo{organization}{PMLR}. pp. \bibinfo{pages}{1188--1196}.
\bibitem[{Lee et~al.(2022)Lee, Lee, Masoud, Krishnan and Li}]{LEE2022101710}
\bibinfo{author}{Lee, D.}, \bibinfo{author}{Lee, S.}, \bibinfo{author}{Masoud,
  N.}, \bibinfo{author}{Krishnan, M.}, \bibinfo{author}{Li, V.C.},
  \bibinfo{year}{2022}.
\newblock \bibinfo{title}{Digital twin-driven deep reinforcement learning for
  adaptive task allocation in robotic construction}.
\newblock \bibinfo{journal}{Advanced Engineering Informatics}
  \bibinfo{volume}{53}, \bibinfo{pages}{101710}.
\newblock \URLprefix
  \url{https://www.sciencedirect.com/science/article/pii/S1474034622001689},
  \DOIprefix\doi{https://doi.org/10.1016/j.aei.2022.101710}.
\bibitem[{Lehner and Dorffner(2020)}]{lehner2020digital}
\bibinfo{author}{Lehner, H.}, \bibinfo{author}{Dorffner, L.},
  \bibinfo{year}{2020}.
\newblock \bibinfo{title}{{Digital geoTwin Vienna: Towards a Digital Twin City
  as Geodata Hub}}.
\newblock \DOIprefix\doi{10.1007/s41064-020-00101-4}.
\bibitem[{Lei et~al.(2023a)Lei, Janssen, Stoter and
  Biljecki}]{lei2023challenges}
\bibinfo{author}{Lei, B.}, \bibinfo{author}{Janssen, P.},
  \bibinfo{author}{Stoter, J.}, \bibinfo{author}{Biljecki, F.},
  \bibinfo{year}{2023}a.
\newblock \bibinfo{title}{Challenges of urban digital twins: A systematic
  review and a delphi expert survey}.
\newblock \bibinfo{journal}{Automation in Construction} \bibinfo{volume}{147},
  \bibinfo{pages}{104716}.
\bibitem[{Lei et~al.(2024)Lei, Liang and Biljecki}]{lei2024integrating}
\bibinfo{author}{Lei, B.}, \bibinfo{author}{Liang, X.},
  \bibinfo{author}{Biljecki, F.}, \bibinfo{year}{2024}.
\newblock \bibinfo{title}{Integrating human perception in 3d city models and
  urban digital twins}.
\newblock \bibinfo{journal}{ISPRS Annals of the Photogrammetry, Remote Sensing
  and Spatial Information Sciences} \bibinfo{volume}{10},
  \bibinfo{pages}{211--218}.
\bibitem[{Lei et~al.(2023b)Lei, Stouffs and
  Biljecki}]{2023_ijgis_3d_city_index}
\bibinfo{author}{Lei, B.}, \bibinfo{author}{Stouffs, R.},
  \bibinfo{author}{Biljecki, F.}, \bibinfo{year}{2023}b.
\newblock \bibinfo{title}{{Assessing and benchmarking 3D city models}}.
\newblock \bibinfo{journal}{International Journal of Geographical Information
  Science} \bibinfo{volume}{37}, \bibinfo{pages}{788--809}.
\newblock \DOIprefix\doi{10.1080/13658816.2022.2140808}.
\bibitem[{Lei et~al.(2023c)Lei, Su and Biljecki}]{lei2023humans}
\bibinfo{author}{Lei, B.}, \bibinfo{author}{Su, Y.}, \bibinfo{author}{Biljecki,
  F.}, \bibinfo{year}{2023}c.
\newblock \bibinfo{title}{Humans as sensors in urban digital twins}, in:
  \bibinfo{booktitle}{International 3D GeoInfo Conference},
  \bibinfo{organization}{Springer}. pp. \bibinfo{pages}{693--706}.
\bibitem[{Leng et~al.(2021)Leng, Wang, Shen, Li, Liu and
  Chen}]{leng2021digital}
\bibinfo{author}{Leng, J.}, \bibinfo{author}{Wang, D.}, \bibinfo{author}{Shen,
  W.}, \bibinfo{author}{Li, X.}, \bibinfo{author}{Liu, Q.},
  \bibinfo{author}{Chen, X.}, \bibinfo{year}{2021}.
\newblock \bibinfo{title}{{Digital twins-based smart manufacturing system
  design in Industry 4.0: A review}}.
\newblock \bibinfo{journal}{Journal of manufacturing systems}
  \bibinfo{volume}{60}, \bibinfo{pages}{119--137}.
\newblock \DOIprefix\doi{10.1016/j.jmsy.2021.05.011}.
\bibitem[{Li and Tan(2023)}]{li2023novel}
\bibinfo{author}{Li, B.}, \bibinfo{author}{Tan, W.}, \bibinfo{year}{2023}.
\newblock \bibinfo{title}{A novel framework for integrating solar renewable
  source into smart cities through digital twin simulations}.
\newblock \bibinfo{journal}{Solar Energy} \bibinfo{volume}{262},
  \bibinfo{pages}{111869}.
\bibitem[{Li et~al.(2022)Li, Aslam, Wileman and Perinpanayagam}]{9656111}
\bibinfo{author}{Li, L.}, \bibinfo{author}{Aslam, S.},
  \bibinfo{author}{Wileman, A.}, \bibinfo{author}{Perinpanayagam, S.},
  \bibinfo{year}{2022}.
\newblock \bibinfo{title}{Digital twin in aerospace industry: A gentle
  introduction}.
\newblock \bibinfo{journal}{IEEE Access} \bibinfo{volume}{10},
  \bibinfo{pages}{9543--9562}.
\newblock \DOIprefix\doi{10.1109/ACCESS.2021.3136458}.
\bibitem[{Li et~al.(2008)Li, Krishnamurthy, Raghavan, Vaithyanathan and
  Jagadish}]{li2008regular}
\bibinfo{author}{Li, Y.}, \bibinfo{author}{Krishnamurthy, R.},
  \bibinfo{author}{Raghavan, S.}, \bibinfo{author}{Vaithyanathan, S.},
  \bibinfo{author}{Jagadish, H.}, \bibinfo{year}{2008}.
\newblock \bibinfo{title}{Regular expression learning for information
  extraction}, in: \bibinfo{booktitle}{Proceedings of the 2008 conference on
  empirical methods in natural language processing}, pp.
  \bibinfo{pages}{21--30}.
\bibitem[{Lim et~al.(2018)Lim, Kim and Maglio}]{LIM201886}
\bibinfo{author}{Lim, C.}, \bibinfo{author}{Kim, K.J.},
  \bibinfo{author}{Maglio, P.P.}, \bibinfo{year}{2018}.
\newblock \bibinfo{title}{Smart cities with big data: Reference models,
  challenges, and considerations}.
\newblock \bibinfo{journal}{Cities} \bibinfo{volume}{82},
  \bibinfo{pages}{86--99}.
\newblock \URLprefix
  \url{https://www.sciencedirect.com/science/article/pii/S0264275117308545},
  \DOIprefix\doi{https://doi.org/10.1016/j.cities.2018.04.011}.
\bibitem[{Lin et~al.(2022)Lin, Chen, Ali, Nugent, Ian, Li, ... and
  Ning}]{lin2022human}
\bibinfo{author}{Lin, Y.}, \bibinfo{author}{Chen, L.}, \bibinfo{author}{Ali,
  A.}, \bibinfo{author}{Nugent, C.}, \bibinfo{author}{Ian, C.},
  \bibinfo{author}{Li, R.}, \bibinfo{author}{...}, \bibinfo{author}{Ning, H.},
  \bibinfo{year}{2022}.
\newblock \bibinfo{title}{Human digital twin: A survey}.
\newblock \bibinfo{journal}{arXiv preprint arXiv:2212.05937}
  \href{http://arxiv.org/abs/2212.05937}{{\tt arXiv:2212.05937}}.
\bibitem[{Liu and Tian(2023)}]{liu2023recognition}
\bibinfo{author}{Liu, C.}, \bibinfo{author}{Tian, Y.}, \bibinfo{year}{2023}.
\newblock \bibinfo{title}{Recognition of digital twin city from the perspective
  of complex system theory: Lessons from chinese practice}.
\newblock \bibinfo{journal}{Journal of Urban Management} \bibinfo{volume}{12},
  \bibinfo{pages}{182--192}.
\bibitem[{Liu and Biljecki(2022)}]{2022_jag_geoai}
\bibinfo{author}{Liu, P.}, \bibinfo{author}{Biljecki, F.},
  \bibinfo{year}{2022}.
\newblock \bibinfo{title}{A review of spatially-explicit geoai applications in
  urban geography}.
\newblock \bibinfo{journal}{International Journal of Applied Earth Observation
  and Geoinformation} \bibinfo{volume}{112}, \bibinfo{pages}{102936}.
\newblock \DOIprefix\doi{10.1016/j.jag.2022.102936}.
\bibitem[{Liu et~al.(2024a)Liu, Yan and Biljecki}]{2024_epb_xai}
\bibinfo{author}{Liu, P.}, \bibinfo{author}{Yan, Z.},
  \bibinfo{author}{Biljecki, F.}, \bibinfo{year}{2024}a.
\newblock \bibinfo{title}{{Explainable spatially explicit geospatial artificial
  intelligence in urban analytics}}.
\newblock \bibinfo{journal}{Environment and Planning B: Urban Analytics and
  City Science} \bibinfo{volume}{51}, \bibinfo{pages}{1104--1123}.
\newblock \DOIprefix\doi{10.1177/23998083231204689}.
\bibitem[{Liu et~al.(2023)Liu, Zhao, Luo, Lei, Frei, Miller and
  Biljecki}]{2023_scs_human_dt}
\bibinfo{author}{Liu, P.}, \bibinfo{author}{Zhao, T.}, \bibinfo{author}{Luo,
  J.}, \bibinfo{author}{Lei, B.}, \bibinfo{author}{Frei, M.},
  \bibinfo{author}{Miller, C.}, \bibinfo{author}{Biljecki, F.},
  \bibinfo{year}{2023}.
\newblock \bibinfo{title}{Towards human-centric digital twins: Leveraging
  computer vision and graph models to predict outdoor comfort}.
\newblock \bibinfo{journal}{Sustainable Cities and Society}
  \bibinfo{volume}{93}, \bibinfo{pages}{104480}.
\newblock \DOIprefix\doi{10.1016/j.scs.2023.104480}.
\bibitem[{Liu et~al.(2024b)Liu, Feng, Lu and Zhou}]{LIU2024102592}
\bibinfo{author}{Liu, Y.}, \bibinfo{author}{Feng, J.}, \bibinfo{author}{Lu,
  J.}, \bibinfo{author}{Zhou, S.}, \bibinfo{year}{2024}b.
\newblock \bibinfo{title}{A review of digital twin capabilities, technologies,
  and applications based on the maturity model}.
\newblock \bibinfo{journal}{Advanced Engineering Informatics}
  \bibinfo{volume}{62}, \bibinfo{pages}{102592}.
\newblock \URLprefix
  \url{https://www.sciencedirect.com/science/article/pii/S1474034624002404},
  \DOIprefix\doi{https://doi.org/10.1016/j.aei.2024.102592}.
\bibitem[{Lo et~al.(2021)Lo, Chen and Zhong}]{LO2021101297}
\bibinfo{author}{Lo, C.}, \bibinfo{author}{Chen, C.}, \bibinfo{author}{Zhong,
  R.Y.}, \bibinfo{year}{2021}.
\newblock \bibinfo{title}{A review of digital twin in product design and
  development}.
\newblock \bibinfo{journal}{Advanced Engineering Informatics}
  \bibinfo{volume}{48}, \bibinfo{pages}{101297}.
\newblock \URLprefix
  \url{https://www.sciencedirect.com/science/article/pii/S1474034621000513},
  \DOIprefix\doi{https://doi.org/10.1016/j.aei.2021.101297}.
\bibitem[{Lu et~al.(2020)Lu, Parlikad, Woodall, Ranasinghe, Xie, Liang,
  Konstantinou, Heaton and Schooling}]{doi:10.1061/(ASCE)ME.1943-5479.0000763}
\bibinfo{author}{Lu, Q.}, \bibinfo{author}{Parlikad, A.K.},
  \bibinfo{author}{Woodall, P.}, \bibinfo{author}{Ranasinghe, G.D.},
  \bibinfo{author}{Xie, X.}, \bibinfo{author}{Liang, Z.},
  \bibinfo{author}{Konstantinou, E.}, \bibinfo{author}{Heaton, J.},
  \bibinfo{author}{Schooling, J.}, \bibinfo{year}{2020}.
\newblock \bibinfo{title}{Developing a digital twin at building and city
  levels: Case study of west cambridge campus}.
\newblock \bibinfo{journal}{Journal of Management in Engineering}
  \bibinfo{volume}{36}, \bibinfo{pages}{05020004}.
\newblock \DOIprefix\doi{10.1061/(ASCE)ME.1943-5479.0000763}.
\bibitem[{Malek et~al.(2021)Malek, Tayefeh, Bender and Barari}]{MALEK20211047}
\bibinfo{author}{Malek, N.G.}, \bibinfo{author}{Tayefeh, M.},
  \bibinfo{author}{Bender, D.}, \bibinfo{author}{Barari, A.},
  \bibinfo{year}{2021}.
\newblock \bibinfo{title}{Live digital twin for smart maintenance in structural
  systems}.
\newblock \bibinfo{journal}{IFAC-PapersOnLine} \bibinfo{volume}{54},
  \bibinfo{pages}{1047--1052}.
\newblock \URLprefix
  \url{https://www.sciencedirect.com/science/article/pii/S2405896321008818},
  \DOIprefix\doi{https://doi.org/10.1016/j.ifacol.2021.08.124}.
  \bibinfo{note}{17th IFAC Symposium on Information Control Problems in
  Manufacturing INCOM 2021}.
\bibitem[{Mauro and Kana(2023)}]{MAURO2023113479}
\bibinfo{author}{Mauro, F.}, \bibinfo{author}{Kana, A.}, \bibinfo{year}{2023}.
\newblock \bibinfo{title}{Digital twin for ship life-cycle: A critical
  systematic review}.
\newblock \bibinfo{journal}{Ocean Engineering} \bibinfo{volume}{269},
  \bibinfo{pages}{113479}.
\newblock \URLprefix
  \url{https://www.sciencedirect.com/science/article/pii/S0029801822027627},
  \DOIprefix\doi{https://doi.org/10.1016/j.oceaneng.2022.113479}.
\bibitem[{Meier et~al.(2021)Meier, M{\"u}ller-Polyzou, Brach and
  Georgiadis}]{meier2021digital}
\bibinfo{author}{Meier, N.}, \bibinfo{author}{M{\"u}ller-Polyzou, R.},
  \bibinfo{author}{Brach, L.}, \bibinfo{author}{Georgiadis, A.},
  \bibinfo{year}{2021}.
\newblock \bibinfo{title}{Digital twin support for laser-based assembly
  assistance}.
\newblock \bibinfo{journal}{Procedia CIRP} \bibinfo{volume}{99},
  \bibinfo{pages}{460--465}.
\bibitem[{Metcalfe et~al.(2024)Metcalfe, Ellul, Morley and
  Stoter}]{Metcalfe_2024}
\bibinfo{author}{Metcalfe, J.}, \bibinfo{author}{Ellul, C.},
  \bibinfo{author}{Morley, J.}, \bibinfo{author}{Stoter, J.},
  \bibinfo{year}{2024}.
\newblock \bibinfo{title}{Characterizing the role of geospatial science in
  digital twins}.
\newblock \bibinfo{journal}{ISPRS International Journal of Geo-Information}
  \bibinfo{volume}{13}, \bibinfo{pages}{320}.
\newblock \URLprefix \url{http://dx.doi.org/10.3390/ijgi13090320},
  \DOIprefix\doi{10.3390/ijgi13090320}.
\bibitem[{Miller et~al.(2021)Miller, Abdelrahman, Chong, Biljecki, Quintana,
  Frei, Chew and Wong}]{miller2021internet}
\bibinfo{author}{Miller, C.}, \bibinfo{author}{Abdelrahman, M.},
  \bibinfo{author}{Chong, A.}, \bibinfo{author}{Biljecki, F.},
  \bibinfo{author}{Quintana, M.}, \bibinfo{author}{Frei, M.},
  \bibinfo{author}{Chew, M.}, \bibinfo{author}{Wong, D.}, \bibinfo{year}{2021}.
\newblock \bibinfo{title}{The internet-of-buildings (iob)—digital twin
  convergence of wearable and iot data with gis/bim}, in:
  \bibinfo{booktitle}{Journal of Physics: Conference Series},
  \bibinfo{organization}{IOP Publishing}. p. \bibinfo{pages}{012041}.
\bibitem[{Mylonas et~al.(2021)Mylonas, Kalogeras, Kalogeras, Anagnostopoulos,
  Alexakos and Mu{\~n}oz}]{9576739}
\bibinfo{author}{Mylonas, G.}, \bibinfo{author}{Kalogeras, A.},
  \bibinfo{author}{Kalogeras, G.}, \bibinfo{author}{Anagnostopoulos, C.},
  \bibinfo{author}{Alexakos, C.}, \bibinfo{author}{Mu{\~n}oz, L.},
  \bibinfo{year}{2021}.
\newblock \bibinfo{title}{Digital twins from smart manufacturing to smart
  cities: A survey}.
\newblock \bibinfo{journal}{IEEE Access} \bibinfo{volume}{9},
  \bibinfo{pages}{143222--143249}.
\newblock \DOIprefix\doi{10.1109/ACCESS.2021.3120843}.
\bibitem[{Negri et~al.(2017)Negri, Fumagalli and Macchi}]{NEGRI2017939}
\bibinfo{author}{Negri, E.}, \bibinfo{author}{Fumagalli, L.},
  \bibinfo{author}{Macchi, M.}, \bibinfo{year}{2017}.
\newblock \bibinfo{title}{A review of the roles of digital twin in cps-based
  production systems}.
\newblock \bibinfo{journal}{Procedia Manufacturing} \bibinfo{volume}{11},
  \bibinfo{pages}{939--948}.
\newblock \URLprefix
  \url{https://www.sciencedirect.com/science/article/pii/S2351978917304067},
  \DOIprefix\doi{https://doi.org/10.1016/j.promfg.2017.07.198}.
  \bibinfo{note}{27th International Conference on Flexible Automation and
  Intelligent Manufacturing, FAIM2017, 27-30 June 2017, Modena, Italy}.
\bibitem[{Nguyen et~al.(2022)Nguyen, Duong, Nguyen, Zhu and
  Zhou}]{NGUYEN2022108381}
\bibinfo{author}{Nguyen, T.}, \bibinfo{author}{Duong, Q.H.},
  \bibinfo{author}{Nguyen, T.V.}, \bibinfo{author}{Zhu, Y.},
  \bibinfo{author}{Zhou, L.}, \bibinfo{year}{2022}.
\newblock \bibinfo{title}{Knowledge mapping of digital twin and physical
  internet in supply chain management: A systematic literature review}.
\newblock \bibinfo{journal}{International Journal of Production Economics}
  \bibinfo{volume}{244}, \bibinfo{pages}{108381}.
\newblock \URLprefix
  \url{https://www.sciencedirect.com/science/article/pii/S0925527321003571},
  \DOIprefix\doi{https://doi.org/10.1016/j.ijpe.2021.108381}.
\bibitem[{Niaz et~al.(2021)Niaz, Shoukat, Jia, Khan, Niaz and Raza}]{9628341}
\bibinfo{author}{Niaz, A.}, \bibinfo{author}{Shoukat, M.U.},
  \bibinfo{author}{Jia, Y.}, \bibinfo{author}{Khan, S.}, \bibinfo{author}{Niaz,
  F.}, \bibinfo{author}{Raza, M.U.}, \bibinfo{year}{2021}.
\newblock \bibinfo{title}{Autonomous driving test method based on digital twin:
  A survey}, in: \bibinfo{booktitle}{2021 International Conference on
  Computing, Electronic and Electrical Engineering (ICE Cube)}, pp.
  \bibinfo{pages}{1--7}.
\newblock \DOIprefix\doi{10.1109/ICECube53880.2021.9628341}.
\bibitem[{Opoku et~al.(2021)Opoku, Perera, Osei-Kyei and
  Rashidi}]{OPOKU2021102726}
\bibinfo{author}{Opoku, D.G.J.}, \bibinfo{author}{Perera, S.},
  \bibinfo{author}{Osei-Kyei, R.}, \bibinfo{author}{Rashidi, M.},
  \bibinfo{year}{2021}.
\newblock \bibinfo{title}{Digital twin application in the construction
  industry: A literature review}.
\newblock \bibinfo{journal}{Journal of Building Engineering}
  \bibinfo{volume}{40}, \bibinfo{pages}{102726}.
\newblock \URLprefix
  \url{https://www.sciencedirect.com/science/article/pii/S2352710221005842},
  \DOIprefix\doi{https://doi.org/10.1016/j.jobe.2021.102726}.
\bibitem[{Opoku et~al.(2022)Opoku, Perera, Osei-Kyei, Rashidi, Famakinwa and
  Bamdad}]{buildings12020113}
\bibinfo{author}{Opoku, D.G.J.}, \bibinfo{author}{Perera, S.},
  \bibinfo{author}{Osei-Kyei, R.}, \bibinfo{author}{Rashidi, M.},
  \bibinfo{author}{Famakinwa, T.}, \bibinfo{author}{Bamdad, K.},
  \bibinfo{year}{2022}.
\newblock \bibinfo{title}{Drivers for digital twin adoption in the construction
  industry: A systematic literature review}.
\newblock \bibinfo{journal}{Buildings} \bibinfo{volume}{12}.
\newblock \URLprefix \url{https://www.mdpi.com/2075-5309/12/2/113},
  \DOIprefix\doi{10.3390/buildings12020113}.
\bibitem[{Osadcha et~al.(2023)Osadcha, Jurelionis and
  Fokaides}]{OSADCHA2023106704}
\bibinfo{author}{Osadcha, I.}, \bibinfo{author}{Jurelionis, A.},
  \bibinfo{author}{Fokaides, P.}, \bibinfo{year}{2023}.
\newblock \bibinfo{title}{Geometric parameter updating in digital twin of built
  assets: A systematic literature review}.
\newblock \bibinfo{journal}{Journal of Building Engineering}
  \bibinfo{volume}{73}, \bibinfo{pages}{106704}.
\newblock \URLprefix
  \url{https://www.sciencedirect.com/science/article/pii/S2352710223008835},
  \DOIprefix\doi{https://doi.org/10.1016/j.jobe.2023.106704}.
\bibitem[{Paden et~al.(2024)Paden, Peters, Garcia-Sanchez and
  Ledoux}]{Pa_en_2024}
\bibinfo{author}{Paden, I.}, \bibinfo{author}{Peters, R.},
  \bibinfo{author}{Garcia-Sanchez, C.}, \bibinfo{author}{Ledoux, H.},
  \bibinfo{year}{2024}.
\newblock \bibinfo{title}{Automatic high-detailed building reconstruction
  workflow for urban microscale simulations}.
\newblock \bibinfo{journal}{Building and Environment} \bibinfo{volume}{265},
  \bibinfo{pages}{111978}.
\newblock \URLprefix \url{http://dx.doi.org/10.1016/j.buildenv.2024.111978},
  \DOIprefix\doi{10.1016/j.buildenv.2024.111978}.
\bibitem[{Pal et~al.(2023)Pal, Lin, Hsieh and
  Golparvar-Fard}]{10.1016/j.dibe.2023.100247}
\bibinfo{author}{Pal, A.}, \bibinfo{author}{Lin, J.J.}, \bibinfo{author}{Hsieh,
  S.H.}, \bibinfo{author}{Golparvar-Fard, M.}, \bibinfo{year}{2023}.
\newblock \bibinfo{title}{{Automated vision-based construction progress
  monitoring in built environment through digital twin}}.
\newblock \bibinfo{journal}{Developments in the Built Environment}
  \bibinfo{volume}{16}, \bibinfo{pages}{100247}.
\newblock \DOIprefix\doi{10.1016/j.dibe.2023.100247}.
\bibitem[{Pang and Biljecki(2022)}]{2022_jag_3d_svi}
\bibinfo{author}{Pang, H.E.}, \bibinfo{author}{Biljecki, F.},
  \bibinfo{year}{2022}.
\newblock \bibinfo{title}{{3D building reconstruction from single street view
  images using deep learning}}.
\newblock \bibinfo{journal}{International Journal of Applied Earth Observation
  and Geoinformation} \bibinfo{volume}{112}, \bibinfo{pages}{102859}.
\newblock \DOIprefix\doi{10.1016/j.jag.2022.102859}.
\bibitem[{Pereira et~al.(2021)Pereira, Buzzo, Zimermann, Neto and
  Malgarezi}]{pereira_descriptive_2021}
\bibinfo{author}{Pereira, A.P.}, \bibinfo{author}{Buzzo, M.},
  \bibinfo{author}{Zimermann, I.}, \bibinfo{author}{Neto, F.H.},
  \bibinfo{author}{Malgarezi, H.}, \bibinfo{year}{2021}.
\newblock \bibinfo{title}{A descriptive {3D} city information model built from
  infrastructure {BIM}: {Capacity} building as a strategy for implementation}.
\newblock \bibinfo{journal}{International Journal of E-Planning Research}
  \bibinfo{volume}{10}.
\newblock \DOIprefix\doi{10.4018/IJEPR.20211001.oa9}.
\bibitem[{Pinto(2022)}]{pinto2022reality}
\bibinfo{author}{Pinto, S.C.D.}, \bibinfo{year}{2022}.
\newblock \bibinfo{title}{Reality Anchor Methodology: Designing a digital twin
  supporting situation awareness}.
\newblock Ph.D. thesis. universit{\'e} Paris-Saclay.
\bibitem[{Qi et~al.(2021)Qi, Tao, Hu, Anwer, Liu, Wei, Wang and
  Nee}]{qi2021enabling}
\bibinfo{author}{Qi, Q.}, \bibinfo{author}{Tao, F.}, \bibinfo{author}{Hu, T.},
  \bibinfo{author}{Anwer, N.}, \bibinfo{author}{Liu, A.}, \bibinfo{author}{Wei,
  Y.}, \bibinfo{author}{Wang, L.}, \bibinfo{author}{Nee, A.Y.},
  \bibinfo{year}{2021}.
\newblock \bibinfo{title}{Enabling technologies and tools for digital twin}.
\newblock \bibinfo{journal}{Journal of Manufacturing Systems}
  \bibinfo{volume}{58}, \bibinfo{pages}{3--21}.
\bibitem[{Raes et~al.(2021)Raes, Michiels, Adolphi, Tampere, Dalianis, McAleer
  and Kogut}]{raes2021duet}
\bibinfo{author}{Raes, L.}, \bibinfo{author}{Michiels, P.},
  \bibinfo{author}{Adolphi, T.}, \bibinfo{author}{Tampere, C.},
  \bibinfo{author}{Dalianis, A.}, \bibinfo{author}{McAleer, S.},
  \bibinfo{author}{Kogut, P.}, \bibinfo{year}{2021}.
\newblock \bibinfo{title}{Duet: A framework for building interoperable and
  trusted digital twins of smart cities}.
\newblock \bibinfo{journal}{{IEEE} Internet Computing} \bibinfo{volume}{26},
  \bibinfo{pages}{43--50}.
\bibitem[{Ramonell et~al.(2023)Ramonell, Chacón and
  Posada}]{10.1016/j.autcon.2023.105109}
\bibinfo{author}{Ramonell, C.}, \bibinfo{author}{Chacón, R.},
  \bibinfo{author}{Posada, H.}, \bibinfo{year}{2023}.
\newblock \bibinfo{title}{{Knowledge graph-based data integration system for
  digital twins of built assets}}.
\newblock \bibinfo{journal}{Automation in Construction} \bibinfo{volume}{156},
  \bibinfo{pages}{105109}.
\newblock \DOIprefix\doi{10.1016/j.autcon.2023.105109}.
\bibitem[{Reimers and Gurevych(2019)}]{reimers2019sentence}
\bibinfo{author}{Reimers, N.}, \bibinfo{author}{Gurevych, I.},
  \bibinfo{year}{2019}.
\newblock \bibinfo{title}{Sentence-bert: Sentence embeddings using siamese
  bert-networks}.
\newblock \bibinfo{journal}{arXiv preprint arXiv:1908.10084} \URLprefix
  \url{https://arxiv.org/abs/1908.10084}. \bibinfo{note}{presented at EMNLP
  2019}.
\bibitem[{Reitsma(2013)}]{Reitsma_2013}
\bibinfo{author}{Reitsma, F.}, \bibinfo{year}{2013}.
\newblock \bibinfo{title}{Revisiting the 'is giscience a science?' debate (or
  quite possibly scientific gerrymandering)}.
\newblock \bibinfo{journal}{International Journal of Geographical Information
  Science} \bibinfo{volume}{27}, \bibinfo{pages}{211--221}.
\newblock \URLprefix \url{http://dx.doi.org/10.1080/13658816.2012.674529},
  \DOIprefix\doi{10.1080/13658816.2012.674529}.
\bibitem[{Rosen et~al.(2015)Rosen, {von Wichert}, Lo and
  Bettenhausen}]{ROSEN2015567}
\bibinfo{author}{Rosen, R.}, \bibinfo{author}{{von Wichert}, G.},
  \bibinfo{author}{Lo, G.}, \bibinfo{author}{Bettenhausen, K.D.},
  \bibinfo{year}{2015}.
\newblock \bibinfo{title}{About the importance of autonomy and digital twins
  for the future of manufacturing}.
\newblock \bibinfo{journal}{IFAC-PapersOnLine} \bibinfo{volume}{48},
  \bibinfo{pages}{567--572}.
\newblock \URLprefix
  \url{https://www.sciencedirect.com/science/article/pii/S2405896315003808},
  \DOIprefix\doi{https://doi.org/10.1016/j.ifacol.2015.06.141}.
  \bibinfo{note}{15th IFAC Symposium onInformation Control Problems
  inManufacturing}.
\bibitem[{dos Santos et~al.(2022)dos Santos, Montevechi, de~Queiroz,
  de~Carvalho~Miranda and Leal}]{doi:10.1080/00207543.2021.1898691}
\bibinfo{author}{dos Santos, C.H.}, \bibinfo{author}{Montevechi, J.A.B.},
  \bibinfo{author}{de~Queiroz, J.A.}, \bibinfo{author}{de~Carvalho~Miranda,
  R.}, \bibinfo{author}{Leal, F.}, \bibinfo{year}{2022}.
\newblock \bibinfo{title}{Decision support in productive processes through des
  and abs in the digital twin era: a systematic literature review}.
\newblock \bibinfo{journal}{International Journal of Production Research}
  \bibinfo{volume}{60}, \bibinfo{pages}{2662--2681}.
\newblock \URLprefix \url{https://doi.org/10.1080/00207543.2021.1898691},
  \DOIprefix\doi{10.1080/00207543.2021.1898691},
  \href{http://arxiv.org/abs/https://doi.org/10.1080/00207543.2021.1898691}{{\tt
  arXiv:https://doi.org/10.1080/00207543.2021.1898691}}.
\bibitem[{Semeraro et~al.(2021)Semeraro, Lezoche, Panetto and
  Dassisti}]{SEMERARO2021103469}
\bibinfo{author}{Semeraro, C.}, \bibinfo{author}{Lezoche, M.},
  \bibinfo{author}{Panetto, H.}, \bibinfo{author}{Dassisti, M.},
  \bibinfo{year}{2021}.
\newblock \bibinfo{title}{Digital twin paradigm: A systematic literature
  review}.
\newblock \bibinfo{journal}{Computers in Industry} \bibinfo{volume}{130},
  \bibinfo{pages}{103469}.
\newblock \URLprefix
  \url{https://www.sciencedirect.com/science/article/pii/S0166361521000762},
  \DOIprefix\doi{https://doi.org/10.1016/j.compind.2021.103469}.
\bibitem[{Shafto et~al.(2010)Shafto, Conroy, Doyle, Glaessgen, Kemp, LeMoigne
  and Wang}]{shafto2010draft}
\bibinfo{author}{Shafto, M.}, \bibinfo{author}{Conroy, M.},
  \bibinfo{author}{Doyle, R.}, \bibinfo{author}{Glaessgen, E.},
  \bibinfo{author}{Kemp, C.}, \bibinfo{author}{LeMoigne, J.},
  \bibinfo{author}{Wang, L.}, \bibinfo{year}{2010}.
\newblock \bibinfo{title}{Draft modeling, simulation, information technology \&
  processing roadmap}.
\newblock \bibinfo{journal}{Technology area} \bibinfo{volume}{11},
  \bibinfo{pages}{1--32}.
\bibitem[{Shaharuddin et~al.(2022)Shaharuddin, Abdul~Maulud, Syed Abdul~Rahman
  and Che~Ani}]{isprs-archives-XLVI-4-W3-2021-315-2022}
\bibinfo{author}{Shaharuddin, S.}, \bibinfo{author}{Abdul~Maulud, K.N.},
  \bibinfo{author}{Syed Abdul~Rahman, S.A.F.}, \bibinfo{author}{Che~Ani, A.I.},
  \bibinfo{year}{2022}.
\newblock \bibinfo{title}{Digital twin for indoor disaster in smart city: A
  systematic review}.
\newblock \bibinfo{journal}{The International Archives of the Photogrammetry,
  Remote Sensing and Spatial Information Sciences}
  \bibinfo{volume}{XLVI-4/W3-2021}, \bibinfo{pages}{315--322}.
\newblock \URLprefix
  \url{https://isprs-archives.copernicus.org/articles/XLVI-4-W3-2021/315/2022/},
  \DOIprefix\doi{10.5194/isprs-archives-XLVI-4-W3-2021-315-2022}.
\bibitem[{Shahat et~al.(2021)Shahat, Hyun and Yeom}]{su13063386}
\bibinfo{author}{Shahat, E.}, \bibinfo{author}{Hyun, C.T.},
  \bibinfo{author}{Yeom, C.}, \bibinfo{year}{2021}.
\newblock \bibinfo{title}{City digital twin potentials: A review and research
  agenda}.
\newblock \bibinfo{journal}{Sustainability} \bibinfo{volume}{13}.
\newblock \URLprefix \url{https://www.mdpi.com/2071-1050/13/6/3386},
  \DOIprefix\doi{10.3390/su13063386}.
\bibitem[{Shahri and Barforush(2004)}]{shahri2004flexible}
\bibinfo{author}{Shahri, H.H.}, \bibinfo{author}{Barforush, A.A.},
  \bibinfo{year}{2004}.
\newblock \bibinfo{title}{A flexible fuzzy expert system for fuzzy duplicate
  elimination in data cleaning}, in: \bibinfo{booktitle}{International
  Conference on Database and Expert Systems Applications},
  \bibinfo{organization}{Springer}. pp. \bibinfo{pages}{161--170}.
\bibitem[{Sharef et~al.(2022)Sharef, Nasharuddin, Mohamed, Zamani, Osman and
  Yaakob}]{10055149}
\bibinfo{author}{Sharef, N.M.}, \bibinfo{author}{Nasharuddin, N.A.},
  \bibinfo{author}{Mohamed, R.}, \bibinfo{author}{Zamani, N.W.},
  \bibinfo{author}{Osman, M.H.}, \bibinfo{author}{Yaakob, R.},
  \bibinfo{year}{2022}.
\newblock \bibinfo{title}{Applications of data analytics and machine learning
  for digital twin-based precision biodiversity: A review}, in:
  \bibinfo{booktitle}{2022 International Conference on Advanced Creative
  Networks and Intelligent Systems (ICACNIS)}, pp. \bibinfo{pages}{1--7}.
\newblock \DOIprefix\doi{10.1109/ICACNIS57039.2022.10055149}.
\bibitem[{Shi et~al.(2023)Shi, Pan, Jiang and Zhai}]{Shi2023}
\bibinfo{author}{Shi, J.}, \bibinfo{author}{Pan, Z.}, \bibinfo{author}{Jiang,
  L.}, \bibinfo{author}{Zhai, X.}, \bibinfo{year}{2023}.
\newblock \bibinfo{title}{An ontology-based methodology to establish city
  information model of digital twin city by merging bim, gis and iot}.
\newblock \bibinfo{journal}{Advanced Engineering Informatics}
  \bibinfo{volume}{57}.
\newblock \URLprefix
  \url{https://www.sciencedirect.com/science/article/pii/S1474034623002422},
  \DOIprefix\doi{10.1016/j.aei.2023.102114}.
\bibitem[{Shirowzhan et~al.(2020)Shirowzhan, Tan and
  Sepasgozar}]{Shirowzhan_2020}
\bibinfo{author}{Shirowzhan, S.}, \bibinfo{author}{Tan, W.},
  \bibinfo{author}{Sepasgozar, S.M.E.}, \bibinfo{year}{2020}.
\newblock \bibinfo{title}{Digital twin and cybergis for improving connectivity
  and measuring the impact of infrastructure construction planning in smart
  cities}.
\newblock \bibinfo{journal}{ISPRS International Journal of Geo-Information}
  \bibinfo{volume}{9}, \bibinfo{pages}{240}.
\newblock \URLprefix \url{http://dx.doi.org/10.3390/ijgi9040240},
  \DOIprefix\doi{10.3390/ijgi9040240}.
\bibitem[{Sifat et~al.(2023)Sifat, Choudhury, Das, Ahamed, Muyeen, Hasan, Ali,
  Tasneem, Islam, Islam, Badal, Abhi, Sarker and Das}]{SIFAT2023100213}
\bibinfo{author}{Sifat, M.M.H.}, \bibinfo{author}{Choudhury, S.M.},
  \bibinfo{author}{Das, S.K.}, \bibinfo{author}{Ahamed, M.H.},
  \bibinfo{author}{Muyeen, S.}, \bibinfo{author}{Hasan, M.M.},
  \bibinfo{author}{Ali, M.F.}, \bibinfo{author}{Tasneem, Z.},
  \bibinfo{author}{Islam, M.M.}, \bibinfo{author}{Islam, M.R.},
  \bibinfo{author}{Badal, M.F.R.}, \bibinfo{author}{Abhi, S.H.},
  \bibinfo{author}{Sarker, S.K.}, \bibinfo{author}{Das, P.},
  \bibinfo{year}{2023}.
\newblock \bibinfo{title}{Towards electric digital twin grid: Technology and
  framework review}.
\newblock \bibinfo{journal}{Energy and AI} \bibinfo{volume}{11},
  \bibinfo{pages}{100213}.
\newblock \URLprefix
  \url{https://www.sciencedirect.com/science/article/pii/S2666546822000593},
  \DOIprefix\doi{https://doi.org/10.1016/j.egyai.2022.100213}.
\bibitem[{Silva et~al.(2018)Silva, Khan and Han}]{SILVA2018697}
\bibinfo{author}{Silva, B.N.}, \bibinfo{author}{Khan, M.},
  \bibinfo{author}{Han, K.}, \bibinfo{year}{2018}.
\newblock \bibinfo{title}{Towards sustainable smart cities: A review of trends,
  architectures, components, and open challenges in smart cities}.
\newblock \bibinfo{journal}{Sustainable Cities and Society}
  \bibinfo{volume}{38}, \bibinfo{pages}{697--713}.
\newblock \URLprefix
  \url{https://www.sciencedirect.com/science/article/pii/S2210670717311125},
  \DOIprefix\doi{https://doi.org/10.1016/j.scs.2018.01.053}.
\bibitem[{Singh et~al.(2023)Singh, Singh, Daultani and
  Chouhan}]{SINGH2023109711}
\bibinfo{author}{Singh, G.}, \bibinfo{author}{Singh, S.},
  \bibinfo{author}{Daultani, Y.}, \bibinfo{author}{Chouhan, M.},
  \bibinfo{year}{2023}.
\newblock \bibinfo{title}{Measuring the influence of digital twins on the
  sustainability of manufacturing supply chain: A mediating role of supply
  chain resilience and performance}.
\newblock \bibinfo{journal}{Computers \& Industrial Engineering}
  \bibinfo{volume}{186}, \bibinfo{pages}{109711}.
\newblock \URLprefix
  \url{https://www.sciencedirect.com/science/article/pii/S0360835223007350},
  \DOIprefix\doi{https://doi.org/10.1016/j.cie.2023.109711}.
\bibitem[{Singh et~al.(2022)Singh, Srivastava, Fuenmayor, Kuts, Qiao, Murray
  and Devine}]{app12115727}
\bibinfo{author}{Singh, M.}, \bibinfo{author}{Srivastava, R.},
  \bibinfo{author}{Fuenmayor, E.}, \bibinfo{author}{Kuts, V.},
  \bibinfo{author}{Qiao, Y.}, \bibinfo{author}{Murray, N.},
  \bibinfo{author}{Devine, D.}, \bibinfo{year}{2022}.
\newblock \bibinfo{title}{Applications of digital twin across industries: A
  review}.
\newblock \bibinfo{journal}{Applied Sciences} \bibinfo{volume}{12}.
\newblock \URLprefix \url{https://www.mdpi.com/2076-3417/12/11/5727},
  \DOIprefix\doi{10.3390/app12115727}.
\bibitem[{Sommer et~al.(2023)Sommer, Stjepandić, Stobrawa and von
  Soden}]{Sommer2023}
\bibinfo{author}{Sommer, M.}, \bibinfo{author}{Stjepandić, J.},
  \bibinfo{author}{Stobrawa, S.}, \bibinfo{author}{von Soden, M.},
  \bibinfo{year}{2023}.
\newblock \bibinfo{title}{Automated generation of digital twin for a built
  environment using scan and object detection as input for production
  planning}.
\newblock \bibinfo{journal}{Journal of Industrial Information Integration}
  \bibinfo{volume}{33}, \bibinfo{pages}{100462}.
\newblock \URLprefix
  \url{https://www.sciencedirect.com/science/article/pii/S2452414X23000353},
  \DOIprefix\doi{https://doi.org/10.1016/j.jii.2023.100462}.
\bibitem[{Soori et~al.(2023)Soori, Arezoo and Dastres}]{SOORI2023100017}
\bibinfo{author}{Soori, M.}, \bibinfo{author}{Arezoo, B.},
  \bibinfo{author}{Dastres, R.}, \bibinfo{year}{2023}.
\newblock \bibinfo{title}{Digital twin for smart manufacturing, a review}.
\newblock \bibinfo{journal}{Sustainable Manufacturing and Service Economics}
  \bibinfo{volume}{2}, \bibinfo{pages}{100017}.
\newblock \URLprefix
  \url{https://www.sciencedirect.com/science/article/pii/S2667344423000099},
  \DOIprefix\doi{https://doi.org/10.1016/j.smse.2023.100017}.
\bibitem[{Steindl et~al.(2020)Steindl, Stagl, Kasper, Kastner and
  Hofmann}]{SteindlGeneric}
\bibinfo{author}{Steindl, G.}, \bibinfo{author}{Stagl, M.},
  \bibinfo{author}{Kasper, L.}, \bibinfo{author}{Kastner, W.},
  \bibinfo{author}{Hofmann, R.}, \bibinfo{year}{2020}.
\newblock \bibinfo{title}{Generic digital twin architecture for industrial
  energy systems}.
\newblock \bibinfo{journal}{Applied Sciences} \bibinfo{volume}{10}.
\newblock \URLprefix \url{https://doi.org/10.3390/app10248903},
  \DOIprefix\doi{10.3390/app10248903}.
\bibitem[{Stoter et~al.(2021)Stoter, Ohori and Noardo}]{stoter2021digital}
\bibinfo{author}{Stoter, J.}, \bibinfo{author}{Ohori, K.A.},
  \bibinfo{author}{Noardo, F.}, \bibinfo{year}{2021}.
\newblock \bibinfo{title}{Digital twins: A comprehensive solution or hopeful
  vision?}
\newblock \bibinfo{journal}{GIM international: the worldwide magazine for
  geomatics} .
\bibitem[{Su et~al.(2023)Su, Zhong and Jiang}]{su2023digital}
\bibinfo{author}{Su, S.}, \bibinfo{author}{Zhong, R.}, \bibinfo{author}{Jiang,
  Y.}, \bibinfo{year}{2023}.
\newblock \bibinfo{title}{Digital twin and its applications in the construction
  industry: A state-of-art systematic review}.
\newblock \bibinfo{journal}{Digital Twin} \bibinfo{volume}{2}.
\newblock \URLprefix \url{https://doi.org/10.12688/digitaltwin.17664.2},
  \DOIprefix\doi{10.12688/digitaltwin.17664.2}. \bibinfo{note}{version 2; peer
  review: 2 approved with reservations}.
\bibitem[{Tao et~al.(2018)Tao, Zhang, Liu and Nee}]{tao2018digitalindustry}
\bibinfo{author}{Tao, F.}, \bibinfo{author}{Zhang, H.}, \bibinfo{author}{Liu,
  A.}, \bibinfo{author}{Nee, A.Y.}, \bibinfo{year}{2018}.
\newblock \bibinfo{title}{{Digital twin in industry: State-of-the-art}}.
\newblock \bibinfo{journal}{IEEE Transactions on Industrial Informatics}
  \bibinfo{volume}{15}, \bibinfo{pages}{2405--2415}.
\newblock \DOIprefix\doi{10.1109/TII.2018.2873186}.
\bibitem[{Thaduri(2024)}]{10.1007/978-3-031-39619-9_43}
\bibinfo{author}{Thaduri, A.}, \bibinfo{year}{2024}.
\newblock \bibinfo{title}{Digital twin: Definitions, classification, and
  maturity}, in: \bibinfo{editor}{Kumar, U.}, \bibinfo{editor}{Karim, R.},
  \bibinfo{editor}{Galar, D.}, \bibinfo{editor}{Kour, R.} (Eds.),
  \bibinfo{booktitle}{International Congress and Workshop on Industrial AI and
  eMaintenance 2023}, \bibinfo{publisher}{Springer Nature Switzerland},
  \bibinfo{address}{Cham}. pp. \bibinfo{pages}{585--599}.
\bibitem[{Trantas et~al.(2023)Trantas, Plug, Pileggi and
  Lazovik}]{TRANTAS2023102357}
\bibinfo{author}{Trantas, A.}, \bibinfo{author}{Plug, R.},
  \bibinfo{author}{Pileggi, P.}, \bibinfo{author}{Lazovik, E.},
  \bibinfo{year}{2023}.
\newblock \bibinfo{title}{Digital twin challenges in biodiversity modelling}.
\newblock \bibinfo{journal}{Ecological Informatics} \bibinfo{volume}{78},
  \bibinfo{pages}{102357}.
\newblock \URLprefix
  \url{https://www.sciencedirect.com/science/article/pii/S1574954123003862},
  \DOIprefix\doi{https://doi.org/10.1016/j.ecoinf.2023.102357}.
\bibitem[{VanDerHorn and Mahadevan(2021)}]{VANDERHORN2021113524}
\bibinfo{author}{VanDerHorn, E.}, \bibinfo{author}{Mahadevan, S.},
  \bibinfo{year}{2021}.
\newblock \bibinfo{title}{Digital twin: Generalization, characterization and
  implementation}.
\newblock \bibinfo{journal}{Decision Support Systems} \bibinfo{volume}{145},
  \bibinfo{pages}{113524}.
\newblock \URLprefix
  \url{https://www.sciencedirect.com/science/article/pii/S0167923621000348},
  \DOIprefix\doi{https://doi.org/10.1016/j.dss.2021.113524}.
\bibitem[{Verdouw et~al.(2021)Verdouw, Tekinerdogan, Beulens and
  Wolfert}]{verdouw2021digital}
\bibinfo{author}{Verdouw, C.}, \bibinfo{author}{Tekinerdogan, B.},
  \bibinfo{author}{Beulens, A.}, \bibinfo{author}{Wolfert, S.},
  \bibinfo{year}{2021}.
\newblock \bibinfo{title}{Digital twins in smart farming}.
\newblock \bibinfo{journal}{Agricultural Systems} \bibinfo{volume}{189},
  \bibinfo{pages}{103046}.
\newblock \URLprefix
  \url{https://www.sciencedirect.com/science/article/pii/S0308521X20309070},
  \DOIprefix\doi{https://doi.org/10.1016/j.agsy.2020.103046}.
\bibitem[{Vo et~al.(2024)Vo, Liu and Tran}]{VO2024107643}
\bibinfo{author}{Vo, T.T.}, \bibinfo{author}{Liu, M.K.}, \bibinfo{author}{Tran,
  M.Q.}, \bibinfo{year}{2024}.
\newblock \bibinfo{title}{Harnessing attention mechanisms in a comprehensive
  deep learning approach for induction motor fault diagnosis using raw
  electrical signals}.
\newblock \bibinfo{journal}{Engineering Applications of Artificial
  Intelligence} \bibinfo{volume}{129}, \bibinfo{pages}{107643}.
\newblock \URLprefix
  \url{https://www.sciencedirect.com/science/article/pii/S0952197623018274},
  \DOIprefix\doi{https://doi.org/10.1016/j.engappai.2023.107643}.
\bibitem[{Wagg et~al.(2020)Wagg, Gardner, Barthorpe and Worden}]{wagg2020key}
\bibinfo{author}{Wagg, D.J.}, \bibinfo{author}{Gardner, P.},
  \bibinfo{author}{Barthorpe, R.J.}, \bibinfo{author}{Worden, K.},
  \bibinfo{year}{2020}.
\newblock \bibinfo{title}{{On Key Technologies for Realising Digital Twins for
  Structural Dynamics Applications}}, in: \bibinfo{booktitle}{Model Validation
  and Uncertainty Quantification, Volume 3}. \bibinfo{publisher}{Springer}, pp.
  \bibinfo{pages}{267--272}.
\newblock \DOIprefix\doi{10.1007/978-3-030-12075-7\_30}.
\bibitem[{Wang et~al.(2022a)Wang, Wang, Li and et~al.}]{wang2022review}
\bibinfo{author}{Wang, K.}, \bibinfo{author}{Wang, Y.}, \bibinfo{author}{Li,
  Y.}, \bibinfo{author}{et~al.}, \bibinfo{year}{2022}a.
\newblock \bibinfo{title}{A review of the technology standards for enabling
  digital twin}.
\newblock \bibinfo{journal}{Digital Twin} \bibinfo{volume}{2},
  \bibinfo{pages}{4}.
\newblock \URLprefix \url{https://doi.org/10.12688/digitaltwin.17549.2},
  \DOIprefix\doi{10.12688/digitaltwin.17549.2}.
\bibitem[{Wang and Vu(2023)}]{Wang2023}
\bibinfo{author}{Wang, S.}, \bibinfo{author}{Vu, L.H.}, \bibinfo{year}{2023}.
\newblock \bibinfo{title}{The integration of digital twin and serious game
  framework for new normal virtual urban exploration and social interaction}.
\newblock \bibinfo{journal}{Journal of Urban Management} \bibinfo{volume}{12}.
\newblock \URLprefix
  \url{https://www.sciencedirect.com/science/article/pii/S2226585623000225},
  \DOIprefix\doi{10.1016/j.jum.2023.03.001}.
\bibitem[{Wang et~al.(2022b)Wang, Guo, Li, Tang, Xia and
  Lv}]{10.1016/j.seta.2021.101897}
\bibinfo{author}{Wang, W.}, \bibinfo{author}{Guo, H.}, \bibinfo{author}{Li,
  X.}, \bibinfo{author}{Tang, S.}, \bibinfo{author}{Xia, J.},
  \bibinfo{author}{Lv, Z.}, \bibinfo{year}{2022}b.
\newblock \bibinfo{title}{{Deep learning for assessment of environmental
  satisfaction using BIM big data in energy efficient building digital twins}}.
\newblock \bibinfo{journal}{Sustainable Energy Technologies and Assessments}
  \bibinfo{volume}{50}, \bibinfo{pages}{101897}.
\newblock \DOIprefix\doi{10.1016/j.seta.2021.101897}.
\bibitem[{Weil et~al.(2023)Weil, Bibri, Longchamp, Golay and
  Alahi}]{WEIL2023104862}
\bibinfo{author}{Weil, C.}, \bibinfo{author}{Bibri, S.E.},
  \bibinfo{author}{Longchamp, R.}, \bibinfo{author}{Golay, F.},
  \bibinfo{author}{Alahi, A.}, \bibinfo{year}{2023}.
\newblock \bibinfo{title}{Urban digital twin challenges: A systematic review
  and perspectives for sustainable smart cities}.
\newblock \bibinfo{journal}{Sustainable Cities and Society}
  \bibinfo{volume}{99}, \bibinfo{pages}{104862}.
\newblock \URLprefix
  \url{https://www.sciencedirect.com/science/article/pii/S2210670723004730},
  \DOIprefix\doi{https://doi.org/10.1016/j.scs.2023.104862}.
\bibitem[{White et~al.(2021)White, Zink, Codec{\'a} and
  Clarke}]{white2021digital}
\bibinfo{author}{White, G.}, \bibinfo{author}{Zink, A.},
  \bibinfo{author}{Codec{\'a}, L.}, \bibinfo{author}{Clarke, S.},
  \bibinfo{year}{2021}.
\newblock \bibinfo{title}{A digital twin smart city for citizen feedback}.
\newblock \bibinfo{journal}{Cities} \bibinfo{volume}{110},
  \bibinfo{pages}{103064}.
\bibitem[{Wu et~al.(2023)Wu, Ji, Ma and Xing}]{Wu2023}
\bibinfo{author}{Wu, H.}, \bibinfo{author}{Ji, P.}, \bibinfo{author}{Ma, H.},
  \bibinfo{author}{Xing, L.}, \bibinfo{year}{2023}.
\newblock \bibinfo{title}{A comprehensive review of digital twin from the
  perspective of total process: Data, models, networks and applications}.
\newblock \bibinfo{journal}{Sensors} \bibinfo{volume}{23},
  \bibinfo{pages}{8306}.
\newblock \URLprefix \url{http://dx.doi.org/10.3390/s23198306},
  \DOIprefix\doi{10.3390/s23198306}.
\bibitem[{Wu et~al.(2020)Wu, Yang, Cheng, Zuo and Cheng}]{9327756}
\bibinfo{author}{Wu, J.}, \bibinfo{author}{Yang, Y.}, \bibinfo{author}{Cheng,
  X.}, \bibinfo{author}{Zuo, H.}, \bibinfo{author}{Cheng, Z.},
  \bibinfo{year}{2020}.
\newblock \bibinfo{title}{The development of digital twin technology review},
  in: \bibinfo{booktitle}{2020 Chinese Automation Congress (CAC)}, pp.
  \bibinfo{pages}{4901--4906}.
\newblock \DOIprefix\doi{10.1109/CAC51589.2020.9327756}.
\bibitem[{Xia et~al.(2022)Xia, Liu, Efremochkina, Liu and Lin}]{Xia2022}
\bibinfo{author}{Xia, H.}, \bibinfo{author}{Liu, Z.},
  \bibinfo{author}{Efremochkina, M.}, \bibinfo{author}{Liu, X.},
  \bibinfo{author}{Lin, C.}, \bibinfo{year}{2022}.
\newblock \bibinfo{title}{Study on city digital twin technologies for
  sustainable smart city design: A review and bibliometric analysis of
  geographic information system and building information modeling integration}.
\newblock \bibinfo{journal}{Sustainable Cities and Society}
  \bibinfo{volume}{84}, \bibinfo{pages}{104009}.
\newblock \URLprefix
  \url{https://www.sciencedirect.com/science/article/pii/S2210670722003298},
  \DOIprefix\doi{10.1016/j.scs.2022.104009}.
\bibitem[{Xia and Zou(2023)}]{XIA2023113322}
\bibinfo{author}{Xia, J.}, \bibinfo{author}{Zou, G.}, \bibinfo{year}{2023}.
\newblock \bibinfo{title}{Operation and maintenance optimization of offshore
  wind farms based on digital twin: A review}.
\newblock \bibinfo{journal}{Ocean Engineering} \bibinfo{volume}{268},
  \bibinfo{pages}{113322}.
\newblock \URLprefix
  \url{https://www.sciencedirect.com/science/article/pii/S0029801822026051},
  \DOIprefix\doi{https://doi.org/10.1016/j.oceaneng.2022.113322}.
\bibitem[{Xu et~al.(2023)Xu, Shu, Qiao, Li and
  Xu}]{10.1016/j.measurement.2023.112955}
\bibinfo{author}{Xu, J.}, \bibinfo{author}{Shu, X.}, \bibinfo{author}{Qiao,
  P.}, \bibinfo{author}{Li, S.}, \bibinfo{author}{Xu, J.},
  \bibinfo{year}{2023}.
\newblock \bibinfo{title}{{Developing a digital twin model for monitoring
  building structural health by combining a building information model and a
  real-scene 3D model}}.
\newblock \bibinfo{journal}{Measurement} \bibinfo{volume}{217},
  \bibinfo{pages}{112955}.
\newblock \DOIprefix\doi{10.1016/j.measurement.2023.112955}.
\bibitem[{Yan et~al.(2019)Yan, Zlatanova, Aleksandrov, Diakite and
  Pettit}]{Yan_2019}
\bibinfo{author}{Yan, J.}, \bibinfo{author}{Zlatanova, S.},
  \bibinfo{author}{Aleksandrov, M.}, \bibinfo{author}{Diakite, A.A.},
  \bibinfo{author}{Pettit, C.}, \bibinfo{year}{2019}.
\newblock \bibinfo{title}{Integration of 3d objects and terrain for 3d
  modelling supporting the digital twin}.
\newblock \bibinfo{journal}{ISPRS Annals of the Photogrammetry, Remote Sensing
  and Spatial Information Sciences} \bibinfo{volume}{IV-4/W8},
  \bibinfo{pages}{147--154}.
\newblock \URLprefix
  \url{http://dx.doi.org/10.5194/isprs-annals-IV-4-W8-147-2019},
  \DOIprefix\doi{10.5194/isprs-annals-iv-4-w8-147-2019}.
\bibitem[{Yang et~al.(2021)Yang, Karimi, Kaynak and Yin}]{yang2021developments}
\bibinfo{author}{Yang, D.}, \bibinfo{author}{Karimi, H.},
  \bibinfo{author}{Kaynak, O.}, \bibinfo{author}{Yin, S.},
  \bibinfo{year}{2021}.
\newblock \bibinfo{title}{Developments of digital twin technologies in
  industrial, smart city and healthcare sectors: A survey}.
\newblock \bibinfo{journal}{Complex Engineering Systems} \bibinfo{volume}{1},
  \bibinfo{pages}{3}.
\bibitem[{Yeung et~al.(2022)Yeung, Kim, Donmez and Neira}]{YEUNG2022103957}
\bibinfo{author}{Yeung, H.}, \bibinfo{author}{Kim, F.},
  \bibinfo{author}{Donmez, M.}, \bibinfo{author}{Neira, J.},
  \bibinfo{year}{2022}.
\newblock \bibinfo{title}{Keyhole pores reduction in laser powder bed fusion
  additive manufacturing of nickel alloy 625}.
\newblock \bibinfo{journal}{International Journal of Machine Tools and
  Manufacture} \bibinfo{volume}{183}, \bibinfo{pages}{103957}.
\newblock \URLprefix
  \url{https://www.sciencedirect.com/science/article/pii/S0890695522001080},
  \DOIprefix\doi{https://doi.org/10.1016/j.ijmachtools.2022.103957}.
\bibitem[{Zhang et~al.(2023)Zhang, Yang and Wang}]{ZHANG2023107859}
\bibinfo{author}{Zhang, A.}, \bibinfo{author}{Yang, J.}, \bibinfo{author}{Wang,
  F.}, \bibinfo{year}{2023}.
\newblock \bibinfo{title}{Application and enabling digital twin technologies in
  the operation and maintenance stage of the aec industry: A literature
  review}.
\newblock \bibinfo{journal}{Journal of Building Engineering}
  \bibinfo{volume}{80}, \bibinfo{pages}{107859}.
\newblock \URLprefix
  \url{https://www.sciencedirect.com/science/article/pii/S2352710223020399},
  \DOIprefix\doi{https://doi.org/10.1016/j.jobe.2023.107859}.

\end{thebibliography}
\end{document}